\documentclass[twoside,11pt]{article}

\usepackage[preprint]{jmlr2e}

\newcommand{\nb}{\text{Nb}}

\usepackage{hyperref}       %
\usepackage{url}            %
\usepackage{booktabs}       %
\usepackage{amsfonts}       %
\usepackage{amssymb}        %
\usepackage{bbm}
\usepackage{nicefrac}       %
\usepackage{microtype}      %
\usepackage{tikz}           %
\usepackage[T1]{fontenc}

\usepackage[textsize=tiny]{todonotes}

\usepackage[normalem]{ulem}
\newcommand{\stkout}[1]{\ifmmode\text{\sout{\ensuremath{#1}}}\else\sout{#1}\fi}
\newcommand{\rev}[1]{\textcolor{black}{#1}}

\usepackage[ruled,vlined,linesnumbered]{algorithm2e}
\usepackage{setspace}
\usepackage{amsmath}
\usepackage{amssymb}
\usepackage{bm}
\usepackage{graphicx}
\graphicspath{{figs/}}
\usepackage{mathtools}
\usepackage{subcaption}
\newcommand\ci{\perp\!\!\!\perp}
\DeclareMathOperator*{\argmin}{argmin}
\DeclareMathOperator*{\argmax}{argmax}

\newcommand{\comment}[1]

\usetikzlibrary{shapes, arrows}
\usetikzlibrary{calc}
\usetikzlibrary{matrix}
\tikzset{
   entriesMatrix/.style={rectangle, draw=black, fill=blue!20, inner sep=0pt, minimum size=3mm},
   entriesMatrixZero/.style={rectangle, draw=black, fill=white, inner sep=0pt, minimum size=3mm},
   entriesMatrixZeroDashed/.style={rectangle, draw=black, fill=white, dashed, inner sep=0pt, minimum size=3mm}
}

\jmlrheading{1}{2020}{1-48}{4/00}{10/00}{meila00a}{Ricardo Baptista, Rebecca E. Morrison, Olivier Zahm and Youssef Marzouk}

\ShortHeadings{Learning non-Gaussian graphical models}{Baptista, Marzouk, Morrison, and Zahm}
\firstpageno{1}

\begin{document}

\title{Learning non-Gaussian graphical models\\ via Hessian scores and triangular transport}

\author{\name Ricardo Baptista\thanks{Authors contributed equally.}%
\email rsb@mit.edu\\
       \name Youssef Marzouk \email ymarz@mit.edu \\
       \addr Center for Computational Science and Engineering \\ %
       Massachusetts Institute of Technology\\
       Cambridge, MA 02139-4301, USA
       \AND 
       \name Rebecca E.~Morrison\footnotemark[1] \email rebeccam@colorado.edu \\
       \addr Department of Computer Science\\
       University of Colorado Boulder\\
       Boulder, CO 80309-0430, USA
       \AND
       \name Olivier Zahm \email olivier.zahm@inria.fr \\
       \addr Universit\'e Grenoble Alpes\\
       Inria, CNRS, Grenoble INP, LJK \\
       38000 Grenoble, France
       }

\editor{}%

\maketitle

\begin{abstract}%
    Undirected probabilistic graphical models %
    represent the %
    conditional
    dependencies, or Markov properties, of a collection of random variables. %
    Knowing the sparsity of such a graphical model is
    valuable for modeling multivariate distributions and for efficiently performing inference. %
    While the problem of learning graph structure from data 
    has been studied
    extensively for certain parametric families of distributions, 
    most existing methods fail to consistently recover
    the graph structure for non-Gaussian data. Here we propose an algorithm for learning the
    Markov structure of continuous and non-Gaussian distributions. To characterize conditional
    independence, we introduce a score  based on integrated Hessian information from the joint log-density, and we
    prove that this score upper bounds the conditional mutual information for a general class of distributions. To compute the score, our algorithm \textsc{sing} estimates the density using a deterministic
    coupling, induced by a triangular transport map, and iteratively exploits sparse structure in the map to reveal 
    sparsity in the graph. For certain
    non-Gaussian datasets, we show that our algorithm recovers the graph structure even with
    a biased approximation to the density. Among other examples, %
    we apply \textsc{sing} to learn the dependencies between the states of a %
    chaotic dynamical system with local interactions.
\end{abstract}

\medskip


\section{Introduction} \label{sec:int}
Consider a collection of random variables $\bm{Z} = (Z_1, \dots, Z_d)$ with \rev{probability measure} $\bm{\nu}_\pi$ and
Lebesgue density $\pi$. An undirected graphical model is a representation of the conditional independence structure, or Markov properties, satisfied by $\bm{Z}$. 
In particular, it is a graph $\mathcal{G} = (V,E)$ where the set of vertices $V=\{1,\hdots,d\}$ contains the indices of the random variables and where the set of edges $E$ encodes the statistical dependence structure of $\bm{Z}$ in the following way: for any disjoint subsets $A$, $B$, and $C$ of $V$, $\bm{Z}_A $ and $\bm{Z}_B$ are conditionally independent given $\bm{Z}_C$ if $C$ \emph{separates} $A$ and $B$ in the graph $\mathcal{G}$. That is, if after removing nodes $C$ from $\mathcal{G}$, there is no path between sets $A$ and $B$ in the resulting graph, then the conditional independence property
\begin{equation}\label{eq:GlobalMarkov}
    \bm{Z}_A \ci \bm{Z}_B \,  | \, \bm{Z}_C
\end{equation}
holds. In this case, we say that $\mathcal{G}$ is an independence map (I-map), or Markov network, for $\bm{\nu}_\pi$.
In this work, we focus on minimal I-maps, which are the sparsest I-maps for $\bm{\nu}_\pi$. 
Our objective is to identify the minimal I-map of $\bm{\nu}_\pi$ using samples from $\bm{\nu}_\pi$

Learning the Markov properties of a distribution, given a set of data drawn from it,
is useful for various reasons. Undirected graphs can reveal hierarchical and cyclic interactions between the variables~\citep{epskamp2018gaussian, shen2009detect}. These graph structures have been used to interpret datasets
from many application areas such as handwriting recognition~\citep{fotak2011handwritten,
fischer2013fast}, natural language processing~\citep{biemann2006chinese, mihalcea2011graph}, protein folding~\citep{thomas2008protein}, and
many more~\citep{bloom1977applications, saaty1965finite}. The graphical model also defines efficient factorizations of high-dimensional distributions by representing the density as a product of potential functions that each depend on low-dimensional subsets of variables~\citep{wainwright2006high}. In some fields, such as image processing
and spatial statistics, the Markov structure of the problem may be immediately recognizable~\citep{koller2009probabilistic}, and exploiting the structure of the undirected graphical model greatly simplifies algorithms for inference and
prediction.

Previous work in learning probabilistic graphical models has mainly focused on certain parametric families of distributions~\citep{drton2017structure}. For Gaussian random variables, the conditional independence properties are
encoded by the sparsity of the inverse covariance, or \emph{precision}, matrix. That is, the $(i,j)$ entry of the precision
matrix is zero if and only if variables $Z_i$ and $Z_j$ are conditionally independent given the
rest. Thus, learning the graph reduces to identifying  the support of the non-zero entries in the precision matrix. An active area of research considers how to learn the graph with (very) sparse data relative to the dimension $d$ of the variables. One of the best-known methods %
to tackle this problem 
is the graphical lasso (\textsc{glasso}), introduced by~\citet{banerjee2008model, yuan2007model}, %
which solves an $\ell_1$-penalized maximum likelihood estimation problem for the precision matrix. %
One popular method for finding the \textsc{glasso} estimator is the coordinate descent algorithm of~\citet{friedman2008sparse}. In the case of discrete random variables, approaches for finding sparse graphs include $\ell_1$-penalized logistic regression~\citep{wainwright2006high}, and the estimation of a generalized covariance matrix whose inverse (via its support) encodes conditional independence properties~\citep{loh2012structure}.

In the case of continuous and non-Gaussian data, the connection between the inverse covariance matrix and conditional
independence is lost. Outside of the Gaussian setting, a regularized score matching method was proposed to learn sparse graphs for distributions within the exponential family~\citep{lin2016estimation}. Recently, a large class of multivariate graphical models was considered by combining node-wise conditional distributions in the exponential family~\citep{yang2015graphical, suggala2017expxorcist}. 
Another area of research
has proposed semiparametric methods based on Gaussian copulas to model non-Gaussian data~\citep{liu2009nonparanormal}. %
In this case, the observations are assumed to come from marginal nonlinear transformations of a multivariate Gaussian random vector with known Markov properties. The marginal transformations %
yield potentially non-Gaussian marginal distributions while at the same time preserving the I-map of
the original multivariate Gaussian distribution. Per Sklar's theorem, any multivariate distribution can be written
in terms of a copula and the univariate marginals~\citep{nelsen2007introduction}. Recently, these copulas were extended to fit models based on elliptical distributions~\citep{liu2012transelliptical}. 
However, the specific families of copulas that
can be easily or explicitly described remain rather limited \citep{asmussen2007stochastic}.
Moreover, data generated via Gaussian copula transformations may fail to fully test a structure learning algorithm's
ability to handle non-Gaussianity. %

So far, there has been relatively little work for general non-Gaussian distributions outside of the exponential family.
The current paper is concerned with providing a mathematical and algorithmic
framework to describe and identify the Markov properties of a continuous and non-Gaussian distribution. As an expansion of our NeurIPS paper, \cite{morrison2017beyond}, the main contributions are as follows.

First, we establish a framework which allows for the description and computation of conditional independence properties in the non-Gaussian setting. This is represented by a new \emph{conditional independence score} matrix $\Omega$: each entry $\Omega_{ij}$ is a score for the independence of $Z_i$ and $Z_j$ conditioned on the remaining variables. Each score $\Omega_{ij}$ is defined by the expected magnitude of certain mixed derivatives---in other words, integrated \emph{Hessian} information---from the joint density $\pi$. \rev{For strictly positive and continuously differentiable $\pi$, this score is zero if and only if $Z_i$ and $Z_j$ are conditionally independent.} 
We also show that, under certain assumptions, the score provides an upper bound for the \emph{conditional mutual information}, a widely used measure of conditional dependence.  To compute $\Omega$ given only samples from $\pi$, an estimate of the joint density is needed. This is achieved via a transport map---a transformation that deterministically couples one probability measure to another.

Second, %
we expand the use and analysis of an algorithm called \textsc{sing} (Sparsity Identification in Non-Gaussian distributions). A key element of the algorithm is a thresholding scheme, and thus we propose a class of threshold estimators that are \emph{consistent} for graph recovery. The success of the algorithm also depends on a strong and explicit connection between the sparsity of the graph and the sparsity of triangular transport maps associated with $\pi$. %
 This connection is exploited to iteratively reveal sparsity in the graph. We show numerically that this iterative algorithm provides an improved estimator for the conditional independence score, compared to a non-iterative approach that does not account for the structure in the map. %

Third, we explore the relationship between the distribution in question and how to learn its
corresponding graph. 
Because there exist an infinite number of probability distributions with the same minimal I-map, intuitively it should be easier to identify the graph than to estimate the entire joint distribution. 
We refer to this notion as the \emph{information gap}. Through several empirical and analytical studies, we explore the extent to which this notion is in fact true and discuss its consequences for graph identification. Moreover, we see how this notion can be directly exploited by the algorithm. That is, in some cases, we still recover the correct graph even with a biased approximation to the density resulting from a constrained parameterization for the transport map.
On the other hand, it is also possible that constraining the form of the map yields an incorrect graph. We show examples of both cases in Section~\ref{sec:ex}.

The remainder of the paper is organized as follows. Section~\ref{sec:gp_bis} introduces the
conditional independence score and its connections to conditional mutual information.
Section~\ref{sec:tm} describes a sample-based estimator for the score based on a
transport map approximation to the density. Section~\ref{sec:msing} introduces our first
algorithm for learning the Markov structure, and proposes a class of threshold estimators that are consistent for this structure. To take advantage of the connection between the sparsity of the graph and the transport map, Section~\ref{sec:sing} presents the iterative algorithm 
\textsc{sing}. Section~\ref{sec:ex} demonstrates the performance of the algorithm on a variety of numerical examples and explores the
notion of the information gap.  Section~\ref{sec:con} provides a discussion and outlook on future research directions. Finally, Appendix~\ref{app:CMI_bound} contains the proof of our main result on the
conditional independence score, Appendix~\ref{app:other_proofs} collects the proofs of
results related to transport maps, and Appendix~\ref{app:consistency} contains the proof of consistency for the graph estimator. %

\section{Measures of conditional  independence}\label{sec:gp_bis}
An alternative to the global Markov properties \eqref{eq:GlobalMarkov} for characterizing conditional independence are the pairwise Markov properties. A distribution for $\bm{Z}=(Z_1,\hdots,Z_d)$ satisfies a pairwise Markov property between two variables when the associated nodes are not connected in the graph $\mathcal{G} = (V,E)$.
That is, given two variables $Z_i$ and $Z_j$ for $i\neq j$, the lack of an edge between nodes $i$ and $j$ means the two variables are conditionally independent given the remaining variables: \[ (i,j) \notin E
\iff
Z_i \ci Z_j  \, | \, \bm{Z}_{-ij}.\] 
Here $\bm{Z}_{-ij}$ denotes the random vector obtained by removing the $i$th and $j$th components from $\bm{Z}$.
The pairwise Markov property is, in general, weaker than the global, but when the density is
strictly positive, i.e., $\pi(\bm{z}) > 0\,\, \forall \bm{z} \in \mathbb{R}^{d}$, the global and the pairwise Markov properties are equivalent~\citep[see][]{lauritzen1996graphical}.  In this paper, we restrict our attention to this setting. %

By definition, $Z_i \ci Z_j | \bm{Z}_{-ij}$ means that the joint conditional density $\pi(z_i,z_j|\bm{z}_{-ij}) = \pi(z_i,z_j,\bm{z}_{-ij}) / \int\pi(z_i',z_j',\bm{z}_{-ij})\mathrm{d}z_i'\mathrm{d}z_j'$ factorizes as the product of the conditional marginals 
\begin{equation}\label{eq:ConditionalIndependenceABC}
 \pi(z_i,z_j|\bm{z}_{-ij}) = \pi(z_i|\bm{z}_{-ij}) \pi(z_j|\bm{z}_{-ij}),
\end{equation}
where $\pi(z_i|\bm{z}_{-ij}) = \int \pi(z_i,z_j|\bm{z}_{-ij}) \mathrm{d}z_j$ and $ \pi(z_j|\bm{z}_{-ij})= \int \pi(z_i,z_j|\bm{z}_{-ij}) \mathrm{d}z_i$.
Thus, $Z_i \ci Z_j | \bm{Z}_{-ij}$ allows the joint density to factorize as
$\pi(\bm{z})= \pi(z_i|\bm{z}_{-ij})\pi(z_j|\bm{z}_{-ij}) \pi(\bm{z}_{-ij})$. If we further assume that $\pi$ is continuously differentiable, this yields
\begin{equation}\label{eq:ConditionalIndependenceABC_partialAB}
 \partial_i\partial_j \log\pi(\bm{z}) = 0 ,
\end{equation}
for any $\bm{z}\in\mathbb{R}^d$. Conversely, if a strictly positive continuously differentiable $\pi(\bm{z})$ satisfies \eqref{eq:ConditionalIndependenceABC_partialAB} for any $\bm{z}\in\mathbb{R}^d$, then $\pi(z_i,z_j|\bm{z}_{-ij})$ necessarily factors as in \eqref{eq:ConditionalIndependenceABC} so that $Z_i \ci Z_j | \bm{Z}_{-ij}$. The characterization \eqref{eq:ConditionalIndependenceABC_partialAB} of conditional independence has been already observed in Lemma~4.1 of \citet{spantini2018inference}. 
Based on \eqref{eq:ConditionalIndependenceABC_partialAB}, we propose to measure the conditional independence
of $Z_i$ and $Z_j$ by the score $\Omega_{ij}\geq0$ defined as%
\begin{equation} \label{eq:ConditionalIndependenceScore}
 \Omega_{ij} \coloneqq \int |\partial_i\partial_j \log\pi(\bm{z}) |^2 \, \pi(\bm{z}) \mathrm{d}\bm{z}.
\end{equation}
A similar measure also appears in \cite{morrison2017beyond} under the name of ``generalized precision,'' the difference being the square inside the integral.\footnote{In this paper, we no longer use the term ``generalized precision,'' but keep the same notation $\Omega$ to denote the new conditional independence score.}
For a strictly positive and continuously differentiable density $\pi$, the condition $\Omega_{ij} = 0$ yields \eqref{eq:ConditionalIndependenceABC_partialAB} so that $Z_i$ and $Z_j$ are conditionally independent. That is, the sparsity pattern of $\Omega$ gives the Markov structure of $\pi$. The entries of $\Omega$ also provide a natural score for conditional independence:  a value of $\Omega_{ij}$ near zero means that $Z_i$ and $Z_j$ are nearly conditionally independent, whereas a large value of $\Omega_{ij}$ means that $Z_i$ and $Z_j$ are strongly conditionally dependent. %
In Section~\ref{ssec:tol}, we will use this interpretation to estimate the Markov structure of $\pi$ by \emph{thresholding} the entries of an estimator for $\Omega$.

We can relate the magnitude of score $\Omega_{ij}$ to another popular measure of conditional independence, the conditional mutual information (CMI). The CMI $I(Z_i ; Z_j|\bm{Z}_{-ij})$ is defined as the expected (with respect to $\bm{Z}_{-ij}$) Kullback--Leibler divergence from the product of the marginal conditionals $\pi(z_i|\bm{z}_{-ij})$ and $ \pi(z_j|\bm{z}_{-ij})$ to the joint conditional $\pi(z_i,z_j|\bm{z}_{-ij})$; that is,
\begin{align}
 I(Z_i ; Z_j|\bm{Z}_{-ij}) 
 &= \int \left( \int \pi(z_i,z_j|\bm{z}_{-ij}) \log\frac{\pi(z_i,z_j|\bm{z}_{-ij})}{\pi(z_i|\bm{z}_{-ij}) \pi(z_j|\bm{z}_{-ij})} \textrm{d}z_i\textrm{d}z_j \right)  \pi(\bm{z}_{-ij}) \textrm{d}\bm{z}_{-ij} \label{eq:definitionCMI}\\
 &= \int \log\left(\frac{\pi(z_i,z_j|\bm{z}_{-ij})}{\pi(z_i|\bm{z}_{-ij}) \pi(z_j|\bm{z}_{-ij})} \right)  \pi(\bm{z}) \textrm{d}\bm{z}.
\end{align}
The CMI is widely adopted in part due to its information theoretic interpretations. The following theorem shows that $ I(Z_i ; Z_j|\bm{Z}_{-ij})$ can in fact be \emph{bounded above} by $\Omega_{ij}$. This result relies on logarithmic Sobolev inequalities.

\begin{definition}
  A probability density function $\pi$ on $\mathbb{R}^d$ satisfies the logarithmic Sobolev inequality if there exists a constant $C<\infty$ such that
  \begin{equation}\label{eq:logSob}
  \int h\log \frac{h}{\int h \mathrm{d}\pi} \mathrm{d}\pi \leq \frac{C}{2} \int \|\nabla\log h\|_2^2 \, h \, \mathrm{d}\pi ,   
  \end{equation}
 holds for any continuously differentiable function $h:\mathbb{R}^d\rightarrow \mathbb{R}_{>0}$. Here $\|\cdot\|_2$ denotes the canonical Euclidean norm on $\mathbb{R}^d$. The smallest constant $C=C(\pi)$ such that \eqref{eq:logSob} holds is called the logarithmic Sobolev constant of $\pi$.
\end{definition}

\begin{theorem}\label{th:OmegaBoundCMI}
 Let $\pi$ be a strictly positive continuously differentiable probability density function on $\mathbb{R}^d$. Assume that all the conditional densities of the form $\pi(\cdot|\bm{z}_{-i})\colon z_i\mapsto\pi(z_i|\bm{z}_{-i})$ and $\pi(\cdot,\cdot|\bm{z}_{-ij})\colon (z_i,z_j)\mapsto\pi(z_i,z_j|\bm{z}_{-ij})$ satisfy the logarithmic Sobolev inequality with constants uniformly bounded by some constant $C_0<\infty$, meaning 
 \begin{equation}\label{eq:AssumptionOmegaBoundCMI}
  C( \pi(\cdot|\bm{z}_{-i}) ) \leq C_0
  \quad\text{and}\quad
  C( \pi(\cdot,\cdot|\bm{z}_{-ij}) ) \leq C_0 ,
 \end{equation} 
 for all $\bm{z}_{-i}\in\mathbb{R}^{d-1}$ and $\bm{z}_{-ij}\in\mathbb{R}^{d-2}$ and for all $1\leq i\neq j\leq d$. Then 
 \begin{equation}\label{eq:OmegaBoundCMI}
  I(Z_i ; Z_j|\bm{Z}_{-ij})  \leq C_0^2\, \Omega_{ij} ,
 \end{equation}
 holds for any $i\neq j$.
\end{theorem}
The proof of Theorem~\ref{th:OmegaBoundCMI} is provided in Appendix~\ref{app:CMI_bound}. 

Let us now comment on assumption \eqref{eq:AssumptionOmegaBoundCMI} of Theorem \ref{th:OmegaBoundCMI}.
In general, there is no simple way to compute exactly the logarithmic Sobolev constant of a density.
The Holley--Stroock perturbation criterion \citep{holley19762} and the Bakry--\'Emery criterion \citep{bakry1985diffusions} are commonly used to bound the logarithmic Sobolev constant. 
Following~\citet{zahm2018certified}, one can combine these two criteria to show that if there exists a log-concave probability density $\pi_0$ and two scalars $\alpha>0$ and $\beta<\infty$ such that $\alpha\pi_0(\bm{z})\leq\pi(\bm{z})\leq \beta\pi_0(\bm{z})$ for any $\bm{z} \in \mathbb{R}^{d}$, then $\pi$ satisfies \eqref{eq:AssumptionOmegaBoundCMI} with constant
\begin{equation}\label{eq:BakryEmery}
 C_0 = \frac{\beta}{\alpha \lambda},
\end{equation}
where $\lambda>0$ is a lower bound for the smallest eigenvalue of $-\nabla^2\log\pi_0(\bm{z})$ for any $\bm{z} \in \mathbb{R}^{d}$.
We refer the reader to~\citet[Section~2.2]{zahm2018certified} for more details and examples of probability densities that satisfy the above conditions.

\begin{remark}[Gaussian case]
 Suppose that $\bm{Z}\sim\mathcal{N}(\bm{m},\Sigma)$ is a Gaussian vector with mean $\bm{m}\in\mathbb{R}^d$ and non-singular covariance $\Sigma\in\mathbb{R}^{d\times d}$. Because $\pi(\bm{z})\propto\exp(-\frac{1}{2}(\bm{z}-\bm{m})^\top\Sigma^{-1}(\bm{z}-\bm{m}))$ we have
 \begin{equation}\label{eq:Omega_Precision_GaussianCase}
  \Omega_{ij} = (\Sigma^{-1})_{ij}^2,
 \end{equation}
and so the sparsity pattern of the precision matrix $\Sigma^{-1}$ gives the Markov structure of $\bm{Z}$. This is a well known property of Gaussian vectors; see for instance Proposition 3.3.6. in \cite{drton2008lectures}. This also explains the name ``generalized precision'' in \cite{morrison2017beyond}.
 
Next we show how sharp the inequality \eqref{eq:OmegaBoundCMI} is for a $d=2$ dimensional Gaussian vector $\bm{Z}=(Z_1,Z_2)$ with covariance given $\Sigma = (\begin{smallmatrix} 1 &\rho \\ \rho & 1 \end{smallmatrix})$ for some correlation $-1\leq\rho\leq1$. One can compute the (conditional) mutual information analytically $I(Z_1 ; Z_2|\bm{Z}_{-12}) = I(Z_1 ; Z_2) = -\frac{1}{2}\log(1-\rho^2)$ and the score $\Omega_{12} = (\frac{\rho}{1-\rho^2})^2$.
It remains to compute $C_0$. Using the formula \eqref{eq:BakryEmery} with $\beta=\alpha=1$ (since $\pi$ is log-concave we can chose $\pi_0=\pi$), we obtain $C_0 = \lambda_{\min}(-\nabla^2\log\pi({\bm z}))^{-1} = \lambda_{\max}(\Sigma)$.
Because $\lambda_{\max}(\Sigma)\leq2$, we then have
$$
I(Z_1 ; Z_2) = -\frac{1}{2}\log(1-\rho^2) \overset{\eqref{eq:OmegaBoundCMI}}{\leq}  C_0^2\Omega_{12} = \lambda_{\max}(\Sigma)^2\left(\frac{\rho}{1 - \rho^2}\right)^2 \leq 4 \left(\frac{\rho}{1 - \rho^2}\right)^2.
$$
\end{remark}

We complete this section by comparing the \emph{computational cost} of evaluating the CMI with that of evaluating the conditional independence score $\Omega_{ij}$, for one variable pair $(Z_{i},Z_{j})$. 
Given the joint density $\pi(\bm{z})$,  computing the CMI requires the ability to evaluate the \emph{normalized} conditional density $\pi(z_i,z_j|\bm{z}_{-ij})$ and the two normalized marginal conditionals, $\pi(z_i|\bm{z}_{-ij})$ and $\pi(z_j|\bm{z}_{-ij})$, for any value of $z_i, z_j \in \mathbb{R}, \ \bm{z}_{-ij} \in \mathbb{R}^{d-2}$. 
Breaking this down further, evaluating $\pi(z_i,z_j|\bm{z}_{-ij})$ requires integrating the joint density with respect to $(z_i,z_j)$, while the marginal densities involve further integration with respect to the $z_i$ and $z_j$ variables individually. Then one must take an outer expectation over $\bm{z}$. Evaluating the CMI therefore requires computing \emph{nested} integrals, which is known to be challenging without additional structure~\citep{rainforth2018nesting}. In practice, information theoretic quantities that involve nested integration, such as CMI, are approximated using nested Monte Carlo (NMC) estimators. Given a budget of $n$ samples, the best-case root-mean-squared error of NMC converges at a slow asymptotic rate of $\mathcal{O}(n^{-1/3})$. Moreover, as these Monte Carlo estimators involve the repeated computation of normalizing constants, controlling variance requires the construction of suitable biasing distributions for importance sampling~\citep{gelman1998simulating}, often in a way that depends on the conditioning variables~\citep{feng2019layered}.

On the other hand, computing the score entry $\Omega_{ij}$ requires evaluating a single mixed derivative of the log density $\log \pi(\bm{z})$ for any $\bm{z} \in \mathbb{R}^{d}$. In computing derivatives of the log density, we only require access to the unnormalized $\pi(\bm{z})$. Moreover, the outer expectation of the score can be approximated with standard non-nested Monte Carlo estimators, converging at the usual $\mathcal{O}(n^{-1/2})$ rate. Samples can be drawn directly from the target $\pi$, and no importance sampling of any kind is required. \rev{For these reasons, in this work we focus on efficiently estimating the score matrix rather than computing the CMI.}%

In either case, evaluating the CMI or the conditional independence score requires an explicit functional form of the joint density $\pi$. In a setting where we only have access to samples from $\pi$, many structure learning algorithms rely on density estimation techniques as a  step to learn the Markov structure.
In general, density estimation in high dimensions,
especially for non-Gaussian data, can become  computationally expensive. Thus, a major  question
becomes: What is an efficient or advantageous way to represent the density given the goal of
learning the Markov structure of a distribution? To answer this, we rely on a particular method of density estimation, using transport maps.

\section{Transport maps} \label{sec:tm} 
Given a target probability measure $\bm{\nu}_{\pi}$ and a reference probability measure $\bm{\nu}_{\eta}$, both on $\mathbb{R}^{d}$, a transport map $S\colon \mathbb{R}^{d} \rightarrow \mathbb{R}^{d}$ is a measurable function %
that defines a deterministic coupling between these two measures~\citep{villani2008optimal}. For random variables $\bm{Z} \sim \bm{\nu}_{\pi}$ and $\bm{X} \sim \bm{\nu}_{\eta}$, this map ensures $S(\bm{Z}) = \bm{X}$ in distribution. If this map is invertible, we can sample from $\pi$ by generating i.i.d.\thinspace samples $\{\textbf{x}^{l}\}_{l > 0}$ from $\eta$ and evaluating the inverse map at these samples to obtain i.i.d.\thinspace samples $S^{-1}(\textbf{x}^{l})$ from $\pi$. 
The \rev{measure} of $S(\bm{Z})$ is known as the pushforward measure of $\bm{\nu}_{\pi}$ through $S$ and is denoted by $S_{\sharp}\bm{\nu}_{\pi}$. If the measure $\bm{\nu}_{\pi}$ has density $\pi$, the density of the pushforward measure is denoted by $S_{\sharp}\pi$. %
Similarly, the \rev{measure} of $S^{-1}(\bm{X})$ is known as the pullback measure of $\bm{\nu}_{\eta}$ through $S$ and is denoted by $S^{\sharp}\bm{\nu}_{\eta}$. If the measure $\bm{\nu}_{\eta}$ has density $\eta$, the density of the pullback measure is denoted by $S^{\sharp}\eta$. %
For a diffeomorphism $S$ (i.e., a differentiable bijective map with differentiable inverse), %
the pullback density $S^{\sharp}\eta$ is given by: %
\begin{equation}
    S^\sharp \eta(\bm{z}) = \eta \circ S(\bm{z})\vert \det \nabla S(\bm{z}) \vert, \label{eq:pullback}
\end{equation}
where $\det \nabla S(\bm{z})$ denotes the determinant of the Jacobian of the map $S$ at $\bm{z}$, \rev{and $\circ$ represents the composition operator}. %

In general, there may exist many different transport maps that couple two arbitrary probability distributions. Optimal transport identifies one such map by minimizing an integrated transportation cost that represents the effort of ``moving'' samples from one distribution to another~\citep{peyre2019computational}. We will instead consider transport maps with \emph{triangular} structure, and show how to learn such maps given only samples from the target density $\pi$. 
Triangular structure will confer several advantages, %
which we detail below.  Besides the structure of the map, another important degree of freedom is the choice of reference distribution. For the remainder of this paper, we choose a standard Gaussian reference, $\bm{\nu}_\eta = \mathcal{N}(\bm{0}, I_d)$. As we will see below, this choice, coupled with triangular structure, leads to useful simplicity in the optimization problem used to build estimators of the map. Moreover, the fact that $\bm{\nu}_\eta$ is a product measure guarantees that the triangular transport map inherits sparsity from the Markov structure of $\bm{\nu}_{\pi}$~\citep{spantini2018inference}. This connection will be exploited when we discuss our iterative algorithm for learning the graphical model in Section~\ref{sec:sing}.

\subsection{Lower triangular transport maps }\label{ssec:lttm}

We consider triangular transports between smooth and strictly
    positive densities $\pi$ and $\eta$ on $\mathbb{R}^{d}$. %
    In this setting,\footnote{More generally, we only need the measures $\bm{\nu}_{\pi}$, $\bm{\nu}_{\eta}$ to be absolutely continuous with respect to the Lebesgue measure on $\mathbb{R}^{d}$~\citep[see][]{santambrogio2015optimal}.} \citet{rosenblatt1952remarks,knothe1957contributions,bogachev2005triangular} showed that there exists a unique monotone %
    lower triangular map $S$---known as the Knothe-Rosenblatt (KR) rearrangement---such that
    $S^\sharp \eta = \pi$. %
    A monotone lower triangular map $S$ is a multivariate function of the form
\[ S(\bm{z}) = \begin{bmatrix*}[l]
    S^1(z_1)\\S^2(z_1,z_2)\\S^3(z_1,z_2,z_3)\\\,\,\vdots\\S^d(z_1,\dots
\dots,z_d)\end{bmatrix*},\]
where the $k$th component $S^{k}$ only depends on the first $k$ input variables $\bm{z}_{1:k} \coloneqq (z_1,\dots,z_k)$, and the map $\xi \mapsto S^{k}(z_1,\dots,z_{k-1},\xi)$---the restriction of the component onto its first $k-1$ inputs---is  monotone increasing for all $(z_{1},\dots,z_{k-1}) \in \mathbb{R}^{k-1}$. We denote the space of such lower triangular maps as $\mathcal{S}_\Delta$.

For differentiable triangular maps, the Jacobian determinant of $S$ can be easily evaluated as the product of its partial derivatives, i.e., $\det \nabla S = \prod_{k=1}^{d} \partial_{k} S^{k}$. This enables the pullback density in~\eqref{eq:pullback} to be efficiently evaluated. For more discussion on the advantages of a triangular structure for $S$ we refer the reader to~\citet{marzouk2016sampling}.

\begin{remark}[Gaussian case] \label{rem:gaussian_chol}
Suppose that $\bm{Z} \sim \pi = \mathcal{N}(\bm{0},\Sigma)$ is a Gaussian vector with non-singular covariance and $\bm{X} \sim \eta = \mathcal{N}(\bm{0},I_d)$. Let $ L^T L = \Sigma^{-1}$ be the Cholesky decomposition of the inverse covariance matrix of $\bm{Z}$. 
Then, $L$ is a linear operator that maps samples $\mathbf{z}\sim \pi$ from the target density to samples $\mathbf{x}\sim \eta$ from the reference density, and similarly, $L^{-1}$ maps $\mathbf{x}$ to $\mathbf{z}$. That is,
\[L\mathbf{z} = \mathbf{x}, \quad\quad L^{-1}\mathbf{x} = \mathbf{z}.\] 
Thus, $L\bm{z}$ is an example of a linear lower triangular transport map $S(\bm{z})$. %
We note that other non-triangular transformations also map $\mathbf{z}$ to $\mathbf{x}$ (e.g., any inverse square root of $\Sigma^{-1}$), but $L$ is the unique lower triangular map. %
\end{remark} 

More generally, an affine transport map is sufficient to represent Gaussian target distributions. On the other hand, %
nonlinear maps $S$ can represent %
non-Gaussian
target distributions.

\subsection{Diagonal transport maps} \label{ssec:diag}

Diagonal maps $D\colon \mathbb{R}^{d} \rightarrow \mathbb{R}^{d}$ are one class of lower triangular transport maps where each map component $D^k$ depends only on the $k$th input variable $z_{k}$.  

Let $\bm{\nu}_{\rho}$ be an arbitrary measure with density $\rho$. While the pullback measure of $\bm{\nu}_{\rho}$ through a general lower triangular transport map $S$ has a different Markov structure than $\bm{\nu}_{\rho}$, the pullback measure of $\bm{\nu}_{\rho}$ through a diagonal map has the same Markov structure as $\bm{\nu}_{\rho}$. %
This is formalized in the following proposition, which we prove in Appendix~\ref{app:other_proofs}. %

\begin{proposition} %
\label{prop:diag}
    Let $\bm{\nu}_{\rho}$ be a measure with strictly positive density $\rho$ that is Markov with respect to $\mathcal{G}$, and let $D$ be a differentiable diagonal transport map. %
    Then, the pullback measure $D^\sharp \bm{\nu}_{\rho}$ is also Markov with respect to $\mathcal{G}$.%
\end{proposition}

If the pullback density $\pi = D^{\sharp}\rho$ satisfies the hypotheses in Theorem~\ref{th:OmegaBoundCMI}, we can also observe this property of diagonal transformations %
in the conditional independence score of $\pi$. The following proposition shows that the entries of the score matrix are related to the mixed partial derivatives of the log-density of $\rho$ and the derivatives of the diagonal map components. %

\begin{proposition}%
\label{prop:gpnp}
    Let $\rho$ be a strictly positive continuously differentiable density and let $D$ and its inverse be differentiable diagonal transport maps. Then, the conditional independence score of the pullback density $D^{\sharp}\rho$ is: %
    \begin{equation} \label{eq:CIS_npn}
        \Omega_{ij} = \int \left| \partial_{i}\partial_{j} \log \rho(\bm{x}) \, \partial_i (D^i)^{-1}(x_i) \partial_j (D^j)^{-1}(x_j) \right|^2 \rho(\bm{x}) d\bm{x}. %
    \end{equation}
\end{proposition}
This result is proved in Appendix~\ref{app:other_proofs}.

\begin{remark}[Gaussian case]
Suppose that $\rho$ is a multivariate Gaussian density with non-singular covariance $\Sigma \in \mathbb{R}^{d \times d}$. Then the conditional independence score of $D^{\sharp}\rho$ %
has the form:
\begin{equation*} \Omega_{ij} = (\Sigma^{-1})_{ij}^2 ~ \int \left| \partial_i(D^i)^{-1}(x_i) \partial_j (D^j)^{-1}(x_j)
        \right|^2 \rho(\bm{x}) d\bm{x}.
\end{equation*}
Thus, if the transformation $D$ is strictly increasing so that $D^{i}(z_i) \neq 0$ for all $i$, then the support of $\Omega$ is identical to the support of the inverse covariance matrix, i.e., $\Omega_{ij} = 0$ if and only if $(\Sigma^{-1})_{ij} = 0$. Furthermore, each nonzero entry of the conditional independence score $\Omega_{ij}$ is proportional to $(\Sigma^{-1})_{ij}^2$. 
\end{remark}

The pullback of a multivariate Gaussian density $\rho$ through a diagonal map $D$ is an example of a Gaussian copula. It is well known that Gaussian copulas preserve the Markov properties of $\rho$, while introducing non-Gaussianity via nonlinear diagonal transformations. This class of distributions was considered in the context of structure learning in~\cite{liu2009nonparanormal} and will be studied numerically in Section~\ref{sec:ex}.%

\subsection{Monotone transport map parameterizations} \label{ssec:par}

To approximate the KR rearrangement between a pair of densities on $\mathbb{R}^{d}$, we consider a differentiable and monotone increasing representation for the map. 
Monotonicity  %
is enforced component-wise 
by ensuring the derivative of the $k$th component $S^k$ with respect to the $k$th variable is a strictly positive function, i.e., %
$\partial_{k} S^{k}(z_{1},\dots,z_{k}) > 0$ for all $(z_{1},\dots,z_{k}) \in \mathbb{R}^{k}$. %
A general parameterization that guarantees monotonicity of $S^k$ is:
\begin{equation}
    S^{k}(z_1,\dots,z_k) = c_k(z_1,\dots,z_{k-1}) + \int_0^{z_k}
    g \left(h_k \left(z_1,\dots,z_{k-1},t\right)\right) dt,\label{eq:monotone_map}%
\end{equation}
for some positive function $g\colon \mathbb{R} \rightarrow \mathbb{R}_{+}$ and functions $c_k\colon \mathbb{R}^{k-1} \rightarrow \mathbb{R}$ and $h_k\colon \mathbb{R}^{k} \rightarrow
\mathbb{R}$~\citep{ramsay1998estimating}. Two common choices for $g$ in~\eqref{eq:monotone_map} are $g(x) = \exp(x)$ and $g(x) = x^2$, which result in the ``integrated exponential'' and the the ``integrated squared'' parameterizations, respectively. In this work, we use the integrated squared parameterization because of its computational advantage of closed-form integration under certain choices for $h_{k}$~\citep{spantini2018inference, bigoni2019greedy}.  %
Following~\citet{baptista2020}, 
we parameterize the functions $c_{k}$ and $h_{k}$ using linear %
expansions
 \begin{equation} \label{eq:Hermite_expansions}
     c_{k}(\bm{z}) = \sum_{j} c_{k,j} \psi_{j}(\bm{z}), \hspace{1cm} h_{k}(\bm{z}) = \sum_{j} h_{k,j} \phi_{j}(\bm{z}),
 \end{equation}
in terms of tensor product Hermite functions $(\psi_{j}, \phi_{j})$ and unknown coefficients $\bm{\alpha} = (c_{k,j},h_{k,j})$. We note that recent methods for autoregressive density estimation alternatively represent the functions $c_{k}$ and $h_{k}$ using neural networks~\citep{papamakarios2017masked, jaini2019sum}. 

In practice, we truncate the expansions in~\eqref{eq:Hermite_expansions} by prescribing a maximum total  degree $\beta$ for the multivariate Hermite functions in $c_{k}$ and $h_{k}$. We denote the space of lower-triangular maps with total degree $\beta$ as $\mathcal{S}_\Delta^\beta$. As expected, a higher degree provides a richer basis for density estimation, but requires more computational effort to optimize and more samples to accurately estimate the coefficients. %
Here we follow the convention in~\cite{spantini2018inference} where a map of degree $\beta$ uses %
basis functions up to degree $\beta$ for $c_{k}$ and $\beta-1$ for $h_{k}$, since the latter is then integrated once. To include affine maps within the space $\mathcal{S}_\Delta^\beta$, we also include constant and linear functions with respect to each variable in the expansions~\eqref{eq:Hermite_expansions}. Computations in Section~\ref{sec:ex} using the transport map parameterizations above are performed using the publicly-available software \textsc{TransportMaps}.\footnote{\url{http://transportmaps.mit.edu}}

\subsection{Optimization of the transport map}\label{ssec:opt}
We complete the discussion of transport maps by describing the optimization procedure for finding the map. After defining a maximum polynomial degree for the basis functions within each map component, the resulting map $S_{\bm{\alpha}} \in \mathcal{S}_\Delta^\beta$ is parameterized by a finite number of coefficients $\bm{\alpha} \in \mathbb{R}^{p}$. %
In this subsection we include the subscript $\bm{\alpha}$ on $S$ to emphasize the map's parametric dependence.

A computational approach to find the KR rearrangement that was explored in~\citet{marzouk2016sampling} is to minimize the Kullback--Leibler divergence $D_{\text{KL}}(\pi||S^{\sharp}\eta) = \mathbb{E}_{\pi}[\log(\pi/S^{\sharp}\eta)]$ from the pullback density $S^{\sharp}\eta$ to $\pi$ over the space of monotone increasing triangular maps $\mathcal{S}_{\Delta}$. After parameterizing the maps with coefficients $\bm{\alpha}$, this is equivalent to solving
\begin{align}
    \bm{\alpha}^* &= \argmin_{\bm{\alpha}} \, D_{\text{KL}} \left(\pi || S_{\bm{\alpha}}^\sharp \eta \right) \nonumber \\
             &= \argmax_{\bm{\alpha}} \, \mathbb{E}_\pi \left[\log S_{\bm{\alpha}}^\sharp \eta(\bm{Z}) \right] \label{eq:KL_minimization} %
\end{align} 

As shown in \citet{marzouk2016sampling,parno2014transport}, for standard Gaussian $\eta$ and lower
triangular $S$, this optimization problem is separable across the map components $S^{1}, \ldots,
S^{d}$. In addition, when using the parameterizations in Section~\ref{ssec:par} for each map
component, the optimization problem is unconstrained and differentiable with respect to the
coefficients $\bm{\alpha}$. Therefore, in practice we can use an iterative method such as BFGS to
find the optimal solution~\citep{nocedal2006numerical}. While the problem is not in general convex
in $\bm{\alpha}$, suitable choices of $g$  in \eqref{eq:monotone_map} can ensure that the problem
has a unique global minimizer~\citep{baptista2020}.

Given i.i.d.\thinspace data $\{\textbf{z}^{l}\}_{l=1}^{n}$ from $\pi$, we can approximate the expectation in~\eqref{eq:KL_minimization} and maximize the likelihood associated with this data set. That is,
\begin{align} \label{eq:mle}
    \widehat{\bm{\alpha}} &= \argmax_{\bm{\alpha}}\, \frac{1}{n}\sum_{l=1}^n \log S_{\bm{\alpha}}^\sharp \eta(\textbf{z}^{l}), \\
    &= \argmax_{\bm{\alpha}}\, \frac{1}{n}\sum_{l=1}^n \sum_{k=1}^{d} \left[ -\frac{1}{2} S_{\bm{\alpha}}^k(\textbf{z}_{1:k}^l)^2 + \log \partial_k S_{\bm{\alpha}}^k(\textbf{z}_{1:k}^l) \right]. \nonumber%
\end{align} where the last equality follows from the form of %
the standard Gaussian reference density $\eta$. %
The optimal solution $\widehat{\bm{\alpha}}$ in~\eqref{eq:mle} is the maximum likelihood estimate (MLE) of $\bm{\alpha}^{*}$. Here we assume that the solutions of~\eqref{eq:KL_minimization} and~\eqref{eq:mle} are unique.

Under suitable regularity conditions on the log-likelihood in~\eqref{eq:mle}, %
$\widehat{\bm{\alpha}}$ is a consistent estimator of $\bm{\alpha}^{*}$. Furthermore, it is a random variable that converges in distribution as $n \rightarrow \infty$ to a normal random
vector given by
\begin{equation} \sqrt{n} \left(\widehat{\bm{\alpha}} - \bm{\alpha}^* \right) \overset{d}\longrightarrow
\mathcal{N}\left(\mathbf{0}, \Gamma(\bm{\alpha}^*)^{-1}\right),\label{eq:mle-norm}\end{equation}
where %
$\Gamma(\bm{\alpha}) \in \mathbb{R}^{p \times p}$ is the non-singular Fisher information matrix~\citep{casella2002statistical} for the transport map representation of the density $\pi$ with coefficients $\bm{\alpha}$. %
Entry $(i,j)$ of the Fisher information matrix is given by %
\begin{equation}\label{eq:FisherMiamMiam}
    \Gamma(\bm{\alpha})_{ij} := -\mathbb{E}_{\pi} \left[\partial_{\alpha_{i}}\partial_{\alpha_{j}} \log S_{\bm{\alpha}}^{\sharp}\eta(\bm{Z}) \right].
\end{equation}

We conclude this section with a description of the closed-form solution to~\eqref{eq:mle} when the coefficients parameterize transport maps $S_{\bm{\alpha}}(\bm{z})$ that are affine in $\bm{z}$. A proof of this result is presented in Appendix~\ref{app:other_proofs}.

\begin{proposition}[Affine map optimization] \label{prop:mo}
Suppose $\pi$ is an arbitrary continuous density on $\mathbb{R}^{d}$, and let $\{\mathbf{z}^{l}\}_{l=1}^{n}$ be a sample drawn from $\pi$ with $n \geq d$. If the map components are restricted to be affine functions of the input variables (i.e., polynomial degree $\beta = 1$), maximizing the log-likelihood function that follows from the pullback density in~\eqref{eq:mle} yields a Gaussian approximation to $\pi$ given by $S_{\widehat{\alpha}}^{\sharp}\eta =  \mathcal{N}(\widehat{\mathbf{m}},\widehat{\Sigma})$ with empirical mean $\widehat{\mathbf{m}} = \frac{1}{n} \sum_{l=1}^{n} \mathbf{z}^{l}$ and empirical covariance matrix $\widehat{\Sigma} = \frac{1}{n} \sum_{l=1}^{n} (\mathbf{z}^{l} - \widehat{\mathbf{m}})(\mathbf{z}^{l} - \widehat{\mathbf{m}})^{T}$. %
\end{proposition}

\section{Learning the Markov structure} %
\label{sec:msing}
In this section, we present our first algorithm for learning the edge set of a minimal I-map for
$\bm{\nu}_\pi$. This algorithm uses the transport map representation of the target density $\pi$ to
compute the conditional independence score $\Omega$. In Section~\ref{ssec:gpfs} we define a
sample-based estimator for $\Omega$ and in Section~\ref{ssec:tol} we present a thresholding
procedure to identify the sparsity of $\Omega$. Section~\ref{ssec:nsing} presents the complete
algorithm for learning the edges in the graph, and Section~\ref{sec:scaling} shows that this procedure is consistent.

\subsection{Computation of $\Omega$}\label{ssec:gpfs}
After optimizing the coefficients, the pullback density $S_{\bm{\widehat\alpha}}^\sharp\eta$ defines an approximation to the target density $\pi$. The conditional independence score $\Omega_{ij}$ in~\eqref{eq:ConditionalIndependenceScore} is then estimated as %
\begin{equation} \label{eq:exp_ci} 
 \widehat{\Omega}_{ij} = \mathbb{E}_{\pi} |\partial_{i}\partial_{j}\log S_{\bm{\widehat\alpha}}^\sharp \eta(\bm{Z})|^2. 
\end{equation}
We can approximate the expectation above using $n$ i.i.d.\thinspace samples $\{\textbf{z}^{l}\}_{l=1}^{n}$ from $\pi$, yielding a sample-based estimator of $\widehat{\Omega}$:
\begin{equation} 
\widetilde{\Omega}_{ij} = \frac{1}{n} \sum_{l=1}^n |\partial_{i}\partial_{j} \log S_{\bm{\widehat\alpha}}^\sharp
      \eta(\textbf{z}^{l})|^2. \label{eq:ci} %
\end{equation}
In this work we estimate $\widehat{\Omega}$ using the same $n$ samples from $\pi$ as those used in
\eqref{eq:mle} to estimate the map coefficients $\widehat{\bm{\alpha}}$.  Reusing the samples 
produces a biased approximation of the non-zero values of $\widehat{\Omega}$ at finite $n$.
We observe in our numerical experiments, however, that this bias has no significant impact on the estimation of the sparsity pattern of $\Omega$.

\subsection{Threshold estimator of $\Omega$}%
\label{ssec:tol}

As a result of the sample-based approximations of $S$ and $\widehat{\Omega}$, the sparsity pattern
of $\widetilde{\Omega}$ will not exactly match that of $\Omega$. For instance, a
zero entry in $\Omega$ may result in a small but numerically non-zero entry in $\widetilde{\Omega}$.
To account for this mismatch, we introduce a threshold estimator $\overline{\Omega}$ defined as
\begin{equation}
\overline{\Omega}_{ij} = \begin{cases}\widetilde{\Omega}_{ij}, \quad &\widetilde{\Omega}_{ij} \geq \tau_{ij}\\
0, & \text{otherwise} \end{cases}, \label{eq:Omega_bar}\end{equation}
for some $\tau_{ij}>0$.
Threshold estimators are commonly used for sparse covariance matrix estimation (see \citet{cai2011adaptive}) and $\tau_{ij}$ is usually chosen proportional to the standard deviation of $\widetilde{\Omega}_{ij}$. 
The rationale behind this choice is to threshold the entries of $\widetilde{\Omega}$ whose standard deviation makes them indistinguishable from zero.

To compute the standard deviation or variance of $\widetilde{\Omega}_{ij}$, empirical estimation is not feasible since we only have a unique realization of $\widetilde{\Omega}_{ij}$. Instead we approximate its variance with
\begin{equation} \mathbb{V}(\widetilde{\Omega}_{ij}) \approx \frac{1}{n} \left(\nabla_{\bm{\alpha}} \widetilde{\Omega}_{ij} \right)^T
\Gamma(\bm{{\alpha}})^{-1} \left(\nabla_{\bm{{\alpha}}}
\widetilde{\Omega}_{ij}\right)\Big|_{\bm{{\alpha}}=\widehat{\bm{\alpha}}} \coloneqq \frac{\widetilde{\upsilon}_{ij}^2}{n}, \label{eq:rho}\end{equation}
where $\Gamma(\bm{{\alpha}})$ is the Fisher information in~\eqref{eq:FisherMiamMiam} and $\nabla_{\bm{\alpha}} \widetilde{\Omega}_{ij} \vert_{\bm{\alpha} = \widehat{\bm{\alpha}}}$ denotes the gradient of $\bm{\alpha} \mapsto \widetilde{\Omega}_{ij}(\bm{\alpha}) = \frac{1}{n}\sum_{i=1}^{n} |\partial_i \partial_j \log S_{\bm{\alpha}}^{\sharp}\eta(\textbf{z}^l)|^2$ evaluated at $\widehat{\bm{\alpha}}$. We assume here that $\bm{\alpha} \mapsto \widetilde{\Omega}_{ij}(\bm{\alpha})$ is a continuously differentiable function of the parameters $\bm{\alpha}$ for each entry $(i,j)$. 
The variance approximation in~\eqref{eq:rho} is inspired by the delta method \citep{oehlert1992note}, which exploits the fact that, given a sequence of random variables $\theta_{n} \in \mathbb{R}^{p}$ satisfying $\sqrt{n} \left( \theta_{n} - \theta \right)
    \overset{D}{\longrightarrow}
\mathcal{N}\left(\bm{0}, \Lambda \right)$ and a continuously differentiable function $g\colon \mathbb{R}^{p} \rightarrow \mathbb{R}$ such that $\nabla g(\theta)\neq 0$, we have $\sqrt{n} \left(
g\left(\theta_{n}\right) - g\left(\theta\right) \right) \overset{D}{\longrightarrow}
\mathcal{N}\left(0, \nabla g(\theta)^{T} \Lambda \nabla g(\theta) \right)$. Here we consider $\theta_{n}$ to be the map coefficients $\widehat{\bm{\alpha}}$ and $g$ to be the sample-based score estimator $\widetilde{\Omega}_{ij}$. %
The main difference between~\eqref{eq:rho} and the variance predicted by the delta method is that we evaluate the gradients and the Fisher information at the estimated map coefficients $\bm{\widehat{\alpha}}$ because the limiting coefficients $\bm{\alpha}^*$ are unknown in practice.

For the threshold in \eqref{eq:Omega_bar}, we then use %
\begin{equation} 
 \tau_{ij} = f(n)  \frac{\widetilde{\upsilon}_{ij}}{\sqrt{n}} ,\label{eq:tau}
\end{equation}
where $f(n)$ is a function that increases slowly with $n$. In our numerical experiments, we will choose $f(n) \propto \sqrt{\log n }$.
In Section~\ref{sec:scaling}, we will justify this choice by identifying a class %
of functions $f$ that make the resulting threshold estimator $\overline{\Omega}$ a \emph{consistent} estimator of the sparsity pattern of $\Omega$ as $n \rightarrow \infty$.

\subsection{Non-iterative \textsc{sing} algorithm} \label{ssec:nsing}

The tools above suffice to build an algorithm to learn the Markov structure of a continuous,
non-Gaussian distribution. %
Algorithm~\ref{alg:1sing} learns the graph structure from the support of the threshold estimator
$\overline{\Omega}$ for the conditional independence score. 
We refer to this sequence of steps as the non-iterative Sparsity Identification in Non-Gaussian
distributions (\textsc{n-sing}) algorithm. In Section~\ref{sec:sing} we propose an iterative version
of this algorithm that uses the sparsity of $\overline{\Omega}$ to define improved estimators for
the map and the resulting score matrix $\Omega$.

\begin{algorithm}[!ht]
\setstretch{1.2}
\caption{Non-iterative Sparsity Identification in Non-Gaussian distributions (\textsc{n-sing})\label{alg:1sing}}
\DontPrintSemicolon
        \SetKwInOut{Input}{Input}\SetKwInOut{Output}{Output}\SetKwInOut{Define}{define}
\BlankLine
        \Input{i.i.d.\thinspace sample $\{\textbf{z}^{l}\}_{l=1}^{n} \sim \pi$, maximum polynomial degree $\beta$, threshold scaling function $f$} %
        \Output{Edge set $\widehat{E}_n$ of minimal I-map for $\bm{\nu}_\pi$}
\BlankLine
        Compute transport map: $S_{\widehat{\bm{\alpha}}} = \argmax_{S_{\bm{\alpha}} \in \mathcal{S}_\Delta^\beta} \sum_{l=1}^{n} \log S_{\bm{\alpha}}^{\sharp}\eta(\mathbf{z}^{l})$\; %
        Estimate $\Omega$: %
        $\widetilde{\Omega}_{ij} = \frac{1}{n}\sum_{l=1}^{n} |\partial_{i}\partial_{j} \log S_{\widehat{\bm{\alpha}}}^\sharp
        \eta(\textbf{z}^l)|^2$\;
        Threshold $\widetilde{\Omega}$, with $\tau_{ij} = f(n) \widetilde{\upsilon}_{ij} / \sqrt{n} $, to yield $\overline \Omega$\;
        Compute $\widehat{E}_n$: $(i,j) \in \widehat{E}_n$ if $\overline{\Omega}_{ij} \neq 0$\;
\end{algorithm}

\subsection{Analysis of consistency} \label{sec:scaling}

Here we establish conditions under which the proposed threshold estimator is 
consistent for recovering the edge set $E$ of the minimal I-map for $\bm{\nu}_{\pi}$.
For simplicity, we consider a variant of \textsc{n-sing} that uses an exact expectation $\widehat{\Omega}_{ij}
= \mathbb{E}_{\pi}|\partial_i \partial_j \log S_{\bm{\widehat\alpha}}^{\sharp}\eta(\bm{Z})|^2$ in step 2 of Algorithm~\ref{alg:1sing}. 
Using the finite-sample approximation $\widetilde{\Omega}_{ij}$ would complicate but not fundamentally change the following analysis. 
We also assume that the map parameterization is sufficiently rich to recover the target density exactly, i.e., that there exists a map $S_{\bm{\alpha}^*} \in \mathcal{S}_{\Delta}^{\beta}$ with coefficients $\bm{\alpha}^* \in \mathbb{R}^p$, for some polynomial degree $\beta$, such that $\pi = (S_{\bm{\alpha}^*})^{\sharp}\eta$, where $\bm{\alpha}^*$ is obtained from the solution of~\eqref{eq:KL_minimization}. 
Intuitively, since the maximum likelihood estimate $\widehat{\bm{\alpha}}$ converges to $\bm{\alpha}^*$ as $n\rightarrow\infty$ (recall \eqref{eq:mle-norm}), each entry $\widehat{\Omega}_{ij}$ of the estimated score %
matrix should then converge to the true score $\Omega_{ij}$:
$$
 \widehat{\Omega}_{ij} = \mathbb{E}_{\pi}|\partial_i \partial_j \log S_{\bm{\widehat\alpha}}^{\sharp}\eta(\bm{Z})|^2  
 \quad\underset{n\rightarrow\infty}{\longrightarrow}\quad
 \mathbb{E}_{\pi}|\partial_i \partial_j \log S_{\bm{\alpha}^*}^{\sharp}\eta(\bm{Z})|^2 = \Omega_{ij}.
$$
We will formalize this notion within the analysis below. %

To recover the support of $\Omega$ from $\widehat{\Omega}$, we consider the threshold estimator 
\begin{equation}
\overline{\Omega}_{ij} = \widehat{\Omega}_{ij} \, \mathbbm{1}(\widehat{\Omega}_{ij} > \tau_{ij}),
\qquad\text{with } \tau_{ij} = f(n)\widehat{\upsilon}_{ij}/\sqrt{n},
\label{eq:threshsimple}
\end{equation}
where $\widehat{\upsilon}_{ij}^2 \coloneqq (\nabla_{\bm{\alpha}} \widehat{\Omega}_{ij})^{T} \Gamma(\bm{\alpha})^{-1}(\nabla_{\bm{\alpha}} \widehat{\Omega}_{ij}) \vert_{\bm{\alpha} = \widehat{\bm{\alpha}}}$. 
A proper choice of $f$ is critical to guaranteeing that the support of the threshold estimator $\overline{\Omega}$ converges to the support of $\Omega$ with increasing $n$. 
For instance, when $\Omega_{ij} = 0$, both $\widehat{\Omega}_{ij}$ and $\widehat{\upsilon}_{ij}/\sqrt{n}$ go to zero at the same rate as $n \rightarrow \infty$. As a result, the event $\{\widehat{\Omega}_{ij} > \widehat{\upsilon}_{ij}/\sqrt{n}\}$ asymptotically occurs with a constant non-zero probability, resulting in false positive edges. The role of $f(n)$ in this case is to adjust the rate of convergence of the threshold to ensure that $\widehat{\Omega}_{ij} < \tau_{ij}$ asymptotically, i.e., that there are no false positives. A similar argument holds for false negatives. The following proposition gives sufficient conditions on $f$ to guarantee the recovery of the edge set in the minimal I-map for $\boldsymbol{\nu}_\pi$. The proof is provided in Appendix~\ref{app:consistency}.

\begin{proposition} \label{prop:sample_complexity} 
For the threshold estimator~\eqref{eq:threshsimple}, let $f$ be a function such that $f(n)
    \rightarrow \infty$ and $f(n)/\sqrt{n}\rightarrow0$ as $n\rightarrow\infty$. Then the edge set
    $\widehat{E}_n$ returned by the associated \textsc{n-sing} algorithm is a consistent estimator
    of $E$, i.e., $\mathbb{P} ( \widehat{E}_n = E ) \rightarrow 1$ as $n \rightarrow \infty$. 
\end{proposition}

The main idea behind this result is to show that the probability of false positives (i.e., type 1 errors) or false negatives (i.e., type 2 errors) converges to zero with the prescribed threshold. As demonstrated above, the occurrence of these events is determined by the magnitude of the estimated score $\widehat{\Omega}_{ij}$ relative to the threshold $\tau_{ij}$, or, equivalently, by the magnitude of the ratio $\sqrt{n}\widehat{\Omega}_{ij}/\widehat{\upsilon}_{ij}$ in comparison to $f(n)$. %
For each pair $(i,j)$, we consider the function $g(\bm{\alpha}) = \mathbb{E}_{\pi}|\partial_i \partial_j \log S_{\bm{\alpha}}^{\sharp}\eta(\bm{z})|^2$ and analyze the asymptotic statistics of the ratio $\sqrt{n}\widehat{\Omega}_{ij}/\widehat{\upsilon}_{ij} = \sqrt{n}g(\widehat{\bm{\alpha}})/(\nabla_{\bm{\alpha}} g(\widehat{\bm{\alpha}})^T \Gamma(\widehat{\bm{\alpha}})^{-1} \nabla_{\bm{\alpha}} g(\widehat{\bm{\alpha}}))^{1/2}$ as a function of the estimated map coefficients $\widehat{\bm{\alpha}}$. %
When the variables $Z_i$ and $Z_j$ are conditionally dependent, we have $g(\bm{\alpha}^*) = \Omega_{ij} \neq 0$, and we assume that the gradient $\nabla_{\bm{\alpha}} g(\bm{\alpha}^*) \neq 0$. Then, the limiting distribution of the ratio is Gaussian by an application of the delta method. On the other hand, when $Z_i$ and $Z_j$ are conditionally independent, we have not only $g(\bm{\alpha}^*) = \Omega_{ij} = 0$ but also $\nabla_{\bm{\alpha}} g(\bm{\alpha}^*) = 0$, because both of these terms depend on $\partial_i \partial_j \log \pi(\bm{z})$, which is zero for all $\bm{z} \in \mathbb{R}^{d}$. %
Thus both $\widehat{\Omega}_{ij}$ and $\widehat{\upsilon}_{ij}$ approach zero, and their ratio becomes singular as $n \rightarrow \infty$. In this case, the delta method is no longer valid; instead, we must consider the asymptotic distribution of singular Wald statistics, as analyzed in~\cite{drton2016wald, pillai2016unexpected}. Here, to characterize the limiting distribution, we assume that $\nabla_{\bm{\alpha}}^2 g(\bm{\alpha}^*) \neq 0$, but higher-order derivatives could also be considered if this condition does not hold. Under the asymptotic distributions for these two scenarios, we show that the probabilities of false positives and false negatives converge to zero given the conditions above on the function $f$ in~\eqref{eq:threshsimple}.

While Proposition~\ref{prop:sample_complexity} guarantees the threshold estimator is consistent for \emph{any} $f$ satisfying the criteria above, the selected $f$ may affect the algorithm's finite-sample performance. 
A function $f$ that grows more quickly with $n$ will produce higher thresholds and reduce the probability of false positive edges, while a more slowly growing function will produce lower thresholds and reduce the probability of false negative edges. 
Future work will investigate the impact of these choices on finite-sample bounds, i.e., the number of samples required by the algorithm to recover the graph with high probability. %

\section{Improved estimator for the Markov structure}%
\label{sec:sing}
In Section~\ref{sec:msing} we showed how to estimate the conditional independence score using a transport map representation of the target density. This transport map is a lower-triangular function and thus is a sparse map where each component does not depend on all of its input variables. However, when the target measure $\bm{\nu}_{\pi}$ satisfies conditional independence properties, the transport map we seek to approximate %
can inherit additional sparse structure~\citep[see][Section 5]{spantini2018inference}. %
In this Section, we take advantage of this connection between the sparsity of the graph and the sparsity of the transport map to present an iterative algorithm for learning the graph.

\subsection{Sparsity of the transport map}\label{ssec:spar-tm} %
For a lower triangular function $S$, 
the sparsity pattern of the map, $\mathcal{I}_S$, is defined in
\citet{spantini2018inference} as:
\begin{equation} \mathcal{I}_S \coloneqq \{(j,k):j<k, \partial_j S^k = 0 \}.\end{equation}
That is, the sparsity pattern is the set of all integer pairs $(j,k)$ with $j<k$, such that the $k$th
component of the map does not depend on the $j$th input variable.\footnote{The lower triangular function also satisfies $\partial_{j} S^{k} = 0$ for all $j > k$ by construction.} The complement of this set, i.e.,
\begin{equation} \mathcal{I}_S^c \coloneqq \{(j,k):j<k, \partial_j S^k \neq 0 \},\end{equation}
determines the \emph{active variables} of the map. That is, if $(j,k) \in \mathcal{I}^c_S$, then the $k$th component of the map must depend on the $j$th  input variable. We denote the set of lower triangular maps that respect the sparsity pattern given by $\mathcal{I}_{S}$ as $\mathcal{S}_{\mathcal{I}_{S}} \subset \mathcal{S}_{\Delta}$.

Given a target density $\pi$,  \citet{spantini2018inference} showed that the Markov structure of $\bm{\nu}_\pi$ yields a \emph{tight lower
bound} on the sparsity pattern $\mathcal{I}_S$ of the KR rearrangement that pulls back $\eta$ to $\pi$. Knowledge of this sparsity can be used when solving the variational problem in~\eqref{eq:KL_minimization} by restricting the feasible domain to transport maps with a reduced set of active variables. To determine this sparsity pattern, we perform a series of graph operations on the minimal I-map $\mathcal{G}$ of the target measure $\bm{\nu}_{\pi}$. These operations define the active variables for each map component based on a sequence of intermediate graphs $(\mathcal{G}^{k})$. The graph $\mathcal{G}^{k}$ is identical to the graph obtained in the variable elimination algorithm before marginalizing node $k$ according to the elimination ordering $(d,d-1,\dots,1)$. However, we emphasize that this sparsity pattern is identified only by inspecting the graph, without actually performing variable elimination or additional computation (e.g., marginalization) on the joint density.
We restate the relevant part of this result below.\footnote{Parts 2 and 3 of Theorem 3  in~\citet{spantini2018inference} provide sparsity bounds for the transport map $S^{-1}$, which are related to the marginal independence of $\bm{\nu}_{\pi}$ and do not concern the current work.}
\begin{theorem}[\citet{spantini2018inference}, Theorem~3 (Part 1)]%
\label{thm:map_sparsity}
    Let $\bm{X} \sim \bm{\nu}_\eta$, $\bm{Z} \sim \bm{\nu}_\pi$ with Lebesgue absolutely continuous measures $\bm{\nu}_\eta$, $\bm{\nu}_\pi$, and let $\bm{\nu}_\eta$ be a product measure on $
    \mathbb{R}^{d}$. Moreover, assume that $\bm{\nu}_{\pi}$ is globally Markov with respect to $\mathcal{G}$,
    and define, recursively, the sequence of graphs $(\mathcal{G}^k)_{k=1}^d$ as: (1) $\mathcal{G}^d \coloneqq \mathcal{G}$ and (2) for all $1 \leq k < d$, $\mathcal{G}^{k-1}$ is obtained from
    $\mathcal{G}^k$ by removing node $k$ and by turning its neighborhood $\nb(k, \mathcal{G}^k)$ into a clique. Then the following holds:
    \begin{enumerate}
        \item If $\mathcal{I}_S $ is the sparsity pattern of the transport map $S$, then
            \begin{equation}
            \widehat{\mathcal{I}}_S \subset \mathcal{I}_S,\end{equation} where $\widehat{\mathcal{I}}_S$ is
            the set of integer pairs $(j,k)$ such that $j \notin \nb(k, \mathcal{G}^k)$.
    \end{enumerate}
\end{theorem}

\rev{
\begin{remark} The sequence of elimination steps performed on the graph in Theorem~\ref{thm:map_sparsity} are equivalent to the steps taken by the \emph{symbolic} Cholesky factorization of a sparse symmetric matrix to identify the nonzero structure of its triangular Cholesky factor; see~\cite{lulfesmann2010interactively} for more details on this equivalence. Thus, the sparsity bound $\widehat{\mathcal{I}}_S$ identified by Theorem~\ref{thm:map_sparsity} corresponds to the symbolically predicted sparsity of the Cholesky factor of a symmetric matrix whose $(j,k)$th entry is zero if $Z_j \ci Z_k | \bm{Z}_{-jk}$.
\label{rem:cholesky}
\end{remark}}

\subsection{Ordering variables in the map}\label{ssec:ordering}
\begin{figure}
  \centering
    \begin{subfigure}{.45\textwidth}
  \includegraphics[width=.8\textwidth]{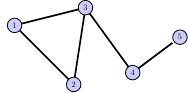}
        \caption{\label{fig:spar-tm-a}}
    \end{subfigure}
    \begin{subfigure}{.45\textwidth}
  \includegraphics[width=.8\textwidth]{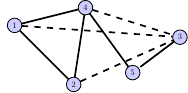}
        \caption{\label{fig:spar-tm-b}}
    \end{subfigure}
    \caption{(a) A sparse graph with an optimal node ordering; (b) Suboptimal ordering induces extra
    edges.\label{fig:spar-tm}}
\end{figure}

In the process above, the sparsity pattern of the map decreases (relative to that of the original I-map $\mathcal{G}$) when adding edges to the intermediate graphs $\mathcal{G}^{k}$ to create cliques. These edges produce \emph{fill-in}. Fill-in will occur unless $\mathcal{G}$ is chordal \emph{and} the variable ordering corresponds to the perfect elimination ordering~\citep{rose1976algorithmic}. 
Whether or not $\mathcal{G}$ is chordal, the amount of fill-in is dependent on the ordering of the input variables.
For example, consider the graph and variable ordering in Figure~\ref{fig:spar-tm-a}.
The corresponding lower bound for the sparsity pattern of the map is
\begin{equation}
\widehat{\mathcal{I}}_S = \{(1,4),(2,4),(1,5),(2,5),(3,5)\},\end{equation}
and the dependence of each map component on the input variables is
\begin{equation} S(\bm{z}) =
\begin{bmatrix*}[l]
    S^1(z_1)\\S^2(z_1,z_2)\\S^3(z_1,z_2,z_3)\\S^4(\quad\quad\;\,\,\,z_3,z_4)\\S^5(\quad\quad\quad\quad\,z_4,z_5)\end{bmatrix*}.
\end{equation}
With the variable ordering shown in Figure~\ref{fig:spar-tm-a}, no edges are added during the process
to identify $\widehat{\mathcal{I}}_S$ and the resulting transport map is more sparse than a dense lower-triangular map. 
In contrast, using the suboptimal ordering shown in
Figure~\ref{fig:spar-tm-b}, edges must be added to the induced graph, shown in dashed lines. The
associated sparsity pattern is now $\widehat{\mathcal{I}}_S =
\{(1,5),(2,5)\}$, and hence the transport map is predicted to be less sparse.

With a larger set $\widehat{\mathcal{I}}_S$, we can simplify the parameterization of the map and reduce the size of the coefficient vector $\bm{\alpha}$; this in turn reduces the variance of the estimator $\widehat{\bm{\alpha}}$~\eqref{eq:mle}, for any given sample size $n$~\citep[see demonstrations in][Section 3.2]{morrison2017beyond}.
Thus, it is
worthwhile to find a variable ordering that maximizes the sparsity of the map.
A variable ordering is equivalent to a permutation $\varphi\colon [d] \rightarrow [d]$ of the nodes in the graph. \cite{spantini2018inference} presented an optimization problem for identifying the permutation that induces the \rev{least fill-in---specifically, the 
largest cardinality of the resulting sparsity bound $\vert \widehat{\mathcal{I}}_S \vert $.} 
This optimization problem is in general  NP-complete~\citep{yannakakis1981computing}. %
\rev{Nevertheless, several heuristics have been proposed to construct permutations based only on the graph $\mathcal{G}$, including weighted min-fill and min-degree orderings~\citep{koller2009probabilistic}.}

\rev{Given a matrix whose sparsity defines the Markov structure of $\bm{\nu}_{\pi}$, an optimal variable ordering can be identified from its sparse Cholesky factorization. From Remark~\ref{rem:cholesky}, we have that the variable dependence in the transport map identified by Theorem 10 corresponds to the fill-in added to the Cholesky factor $L$ of a sparse matrix $\Sigma^{-1} = L^TL$ encoding the Markov structure. As a result, finding an ordering that minimizes fill-in in the graph can be cast as seeking a permutation matrix $P$ so that $P\Sigma^{-1}P^T = L^TL$ where $L$ is as sparse as possible; see~\citet{raskutti2018learning} for consistency guarantees when solving this ideal problem with estimators of $\Sigma^{-1}$.}

\rev{
\begin{remark} In the setting of Gaussian Markov random fields~\citep{rue2005gaussian}, we can take the matrix $\Sigma^{-1}$ to be the (sparse) precision matrix of the Gaussian distribution $\mathcal{N}(0,\Sigma)$. Following  Remark~\ref{rem:gaussian_chol}, the lower-triangular Cholesky factor $L$ of $\Sigma^{-1}$ defines a linear transport map $S(\bm{z}) = L\bm{z}$ that pulls back the reference distribution $\mathcal{N}(\mathbf{0},I_d)$ to the target $\mathcal{N}(\mathbf{0},\Sigma)$. Our goal is to seek a variable ordering that yields the map with sparsest variable dependence, i.e., the Cholesky factor $L$ with the largest number of zero entries.
\end{remark}
}

\rev{For arbitrary non-Gaussian distributions, the sparsity of the conditional independence score after thresholding, $\overline{\Omega}$, provides an estimate of the Markov structure of $\bm{\nu}_{\pi}$. We therefore can find a good ordering by seeking a permutation that induces minimum fill-in in the symbolic Cholesky factorization of $\overline{\Omega}$. We use standard heuristics to solve this problem, which are implemented in software packages such as \textsc{cholmod}~\citep{davis2008user}. The \textsc{cholmod} package uses both the approximate minimum degree algorithm and the nested dissection algorithm implemented in the \textsc{metis} graph partitioning library~\citep{karypis1998fast} to identify candidate orderings, and then chooses the variable ordering that induces the fewest non-zero entries in the Cholesky factor of $\overline{\Omega}$. %
We use this ordering scheme for the numerical examples in Section~\ref{sec:ex}.}

\subsection{The iterative \textsc{sing} algorithm}

In Section~\ref{ssec:spar-tm} we saw how sparsity in the graph implies sparsity in the transport map. To take advantage of this sparsity, we need to know the Markov structure of the target density $\pi$. This Markov structure is not available \textit{a priori} given only samples from $\pi$. However, the \textsc{n-sing} algorithm from Section~\ref{sec:msing} can provide an initial estimate of this Markov structure, based on a dense lower triangular map $S$ (i.e., with $\mathcal{I}_{S} = \emptyset$). This Markov structure can be used to identify a new variable ordering, and a corresponding bound $\widehat{\mathcal{I}}_S$ on the sparsity pattern of the transport map via Theorem~\ref{thm:map_sparsity}. We can then enforce this sparsity bound to compute a new estimate of the transport map, and in turn obtain a new threshold estimate of the Markov structure of $\pi$. 
We repeat this procedure until the sparsity of the estimated graph no longer changes. This defines an iterative application of \textsc{n-sing} that we call the \textsc{sing} algorithm.\footnote{The \textsc{sing} algorithm first appeared with slight modifications in~\citet{morrison2017beyond}.} The formal steps of \textsc{sing} are presented in Algorithm~\ref{alg:sing}. %

The first steps of \textsc{sing} are identical to the \textsc{n-sing} algorithm. The rationale for the remaining steps, and for subsequent iterations, is essentially to exploit sparsity for variance reduction. Sparsity in the graph, coupled with a good variable ordering, leads to sparsity in the map, i.e., a smaller number of active variables. As noted above, such a map can be described by fewer coefficients $\bm{\alpha}$; a maximum likelihood estimate of these coefficients---based on the same finite sample from $\pi$ as in previous iterations---in turn has a smaller variance, as observed in~\cite{morrison2017beyond}. Thus, the iterations of the \textsc{sing} algorithm provide improved estimators of the target density and of the conditional independence score for the goal of learning the graph.

\begin{remark}
    For a multivariate Gaussian target density $\pi = \mathcal{N}(\bm{0},\Sigma)$, the \textsc{sing} algorithm with affine maps (i.e., polynomial degree $\beta = 1$) alternates between estimating a sparse Cholesky factor of the inverse covariance matrix (see Proposition~\ref{prop:mo} and Remark~\ref{rem:gaussian_chol}) and defining a threshold estimator for $\Sigma^{-1}$ to learn the Markov structure of $\pi$. The sparsity of the Cholesky factor is dependent on the sparsity of $\Sigma^{-1}$ and the ordering of the input variables. Recently, several methods have also been proposed to learn sparse Cholesky factors of a sparse inverse covariance matrix for the goal of density estimation in the Gaussian setting. These methods are based on $\ell_{1}$-penalized maximum likelihood estimation~\citep{huang2006covariance}, banded sparsity patterns for the Cholesky factor~\citep{bickel2008regularized, levina2008sparse}, and combinations of multiple variable orderings~\citep{kang2020improved}.%
\end{remark}

To conclude, let us comment on the stopping criterion used in Algorithm~\ref{alg:sing}. During the procedure, the edges in the estimated graph can change freely. For instance, we do not impose any constraint on the edge set $\widehat E^t$ at iteration $t$ (e.g., $\widehat E^t \subseteq \widehat E^{t-1}$) or on the sparsity pattern of the map. %
Nevertheless, we have observed in most of our numerical experiments that the cardinality of estimated edges $|\widehat E^t|$ decreases until the algorithm finds a good estimate for $E$. Thus, checking that $|\widehat{E}^t|$ is non-decreasing works well as a practical stopping criterion.

\begin{algorithm}[!ht]
\setstretch{1.2}
    \caption{Sparsity Identification in Non-Gaussian distributions (\textsc{sing})
    \label{alg:sing}}
\DontPrintSemicolon
    \SetKwInOut{Input}{Input}\SetKwInOut{Output}{Output}\SetKwInOut{Define}{Define}
\BlankLine
    \Input{i.i.d.\thinspace sample $\{\textbf{z}^{l}\}_{l=1}^{n} \sim \pi$, maximum polynomial degree $\beta$, threshold scaling function $f$} %
    \Output{Edge set $\widehat{E}_n$ of minimal I-map for $\bm{\nu}_\pi$}
\BlankLine
    \Define{$\mathcal{S}_{\widehat{\mathcal{I}}^1}^{\beta} = \mathcal{S}_{\Delta}^{\beta}$, $|\widehat E^0| = d(d-1)/2$, $t=1$}%
    \While{$|\widehat{E}^{t}|$ is  decreasing}{%
    Compute transport map: $S_{\widehat{\bm{\alpha}}} = \argmax_{S_{\bm{\alpha}} \in \mathcal{S}_{\widehat{\mathcal{I}}^t}^\beta} \sum_{l=1}^{n} \log S_{\bm{\alpha}}^\sharp \eta(\textbf{z}^l)$\;
    Estimate $\Omega$: %
    $(\widetilde{\Omega}^t)_{ij}  = \frac{1}{n} \sum_{l=1}^{n} | \partial_{i}\partial_{j} \log S_{\widehat{\bm{\alpha}}}^\sharp
    \eta(\textbf{z}^l)|^2$\; %
    Threshold $\widetilde{\Omega}^t$ with $\tau_{ij} = f(n) (\widetilde{\upsilon}^t)_{ij} / \sqrt{n} $, to yield $\overline{\Omega}^t$\;
    Compute $|\widehat{E}^t|$ where $(i,j) \in \widehat{E}^t$ if $(\overline{\Omega}^t)_{ij} \neq 0$\; %
  Find permutation of the variables $\varphi^{t+1}$ (using, e.g., \rev{\textsc{cholmod}})\;
    Identify sparsity pattern of subsequent map $\widehat{\mathcal{I}}^{t+1}$\;
$t \leftarrow t+1$\;
}
Set $\widehat{E}_n = \widehat{E}^{t}$\;
\end{algorithm}

\section{Examples} \label{sec:ex}
In this section we apply the \textsc{sing} algorithm to learn the Markov structure of several non-Gaussian datasets. Section~\ref{ssec:but} presents
results for the butterfly distribution and  demonstrates the value of an iterative algorithm for recovering the Markov structure.
Section~\ref{ssec:nonp} applies \textsc{sing} to data from nonparanormal distributions that were considered in
\citet{liu2009nonparanormal}. Surprisingly, affine transport map approximations to the target density (i.e., with polynomial degree $\beta = 1$) work well for these nonparanormal examples, even with highly non-Gaussian marginals.
In Section~\ref{ssec:x3} we use an analytical example to further investigate why a linear map might still work for non-Gaussian data. We then examine a generalization of the  nonparanormal setting in Section~\ref{ssec:beta2}, where the target distribution is given by a diagonal transformation of a non-Gaussian base distribution. In these examples, we use the approximation class of the transport map that represents the base density also to approximate the target density, and still recover the correct graph. %
Finally, in Sections~\ref{ssec:l96} and \ref{ssec:cells} we consider \rev{two application-oriented examples:} a higher-dimensional physics-based dataset arising from the Lorenz-96 dynamical system \rev{and a data set that captures interactions in a protein-signaling network.}

In all of our numerical experiments, we parameterize the transport maps using the integrated squared representation introduced in Section~\ref{ssec:par}, thereby enforcing the monotonicity of each map component by construction. Within the \textsc{sing} algorithm, we use the ordering heuristics described in Section~\ref{ssec:ordering} to identify a good permutation at each iteration. Following the analysis in Section~\ref{sec:scaling}, %
we set the threshold to $f(n) = c \sqrt{\log n }$ where $c \in \mathbb{R}$ is a constant.
While $c$ can be chosen via cross-validation to improve empirical performance, any value is consistent for learning the graph, and we set $c = 1$ in our numerical investigations. %
Before running \textsc{sing}, we standardize each variable in the dataset by subtracting the empirical mean and dividing by the empirical standard deviation. This normalization step ensures all of the variables are on a similar scale and improves empirical performance. %

To quantitatively evaluate the results of the \textsc{sing} algorithm for recovering the Markov structure using samples from $\pi$, we measure the errors in the estimated edge sets $\widehat{E}$ \rev{(in examples where the true edge set $E$ is available)}. For each graph, we measure the number of false positives: edges that are in $\widehat{E}$ and not in $E$ (Type 1 errors) and the number of false negatives: edges in $E$ that are not in $\widehat{E}$ (Type 2 errors). In the figures below, we report the mean type 1 and 2 errors across 25 runs of the algorithm with independent batches of samples, as well as the 95\% confidence interval for the mean.

\subsection{Butterfly distribution}\label{ssec:but}
The first example consists of $r$ i.i.d.\thinspace pairs of random variables $(P_{i},Q_{i})$, where:
\begin{align}
    P_{i} &\sim \mathcal{N}(0,1)\\
    Q_{i} &= W_{i} P_i, \quad \text{with } W_{i} \sim \mathcal{N}(0,1),\quad W_{i} \ci P_{i},\quad i=1,\dots,r.
\end{align}
One such pair of random variables, or a variation of the above, is a commonly used example to
illustrate how two random variables can be uncorrelated but not independent.

\begin{figure}[!htb]
  \centering
  \begin{subfigure}{.2\textwidth}
    \centering
    \includegraphics[width=\textwidth]{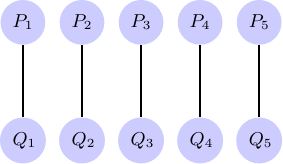}
      \caption{\label{fig:modrad_graph}}
  \end{subfigure}%
  \begin{subfigure}{.4\textwidth}
  \centering
   \includegraphics[width=\textwidth]{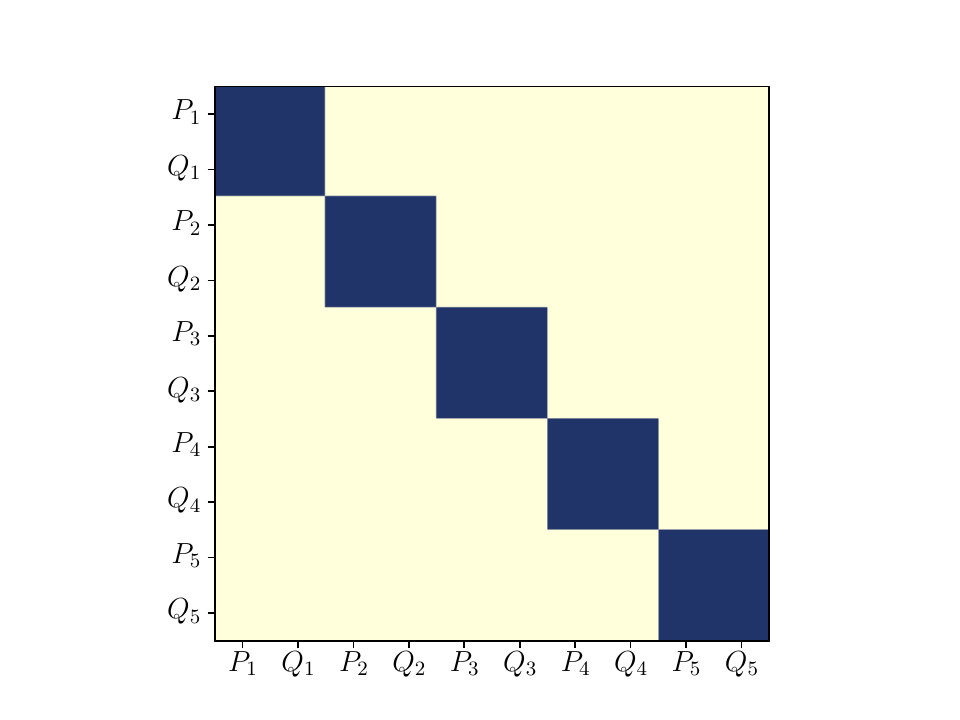}
      \caption{\label{fig:modrad_adj}}
  \end{subfigure}\hspace{-1em}
  \begin{subfigure}{.33\textwidth}
   \centering
    \includegraphics[width=\textwidth]{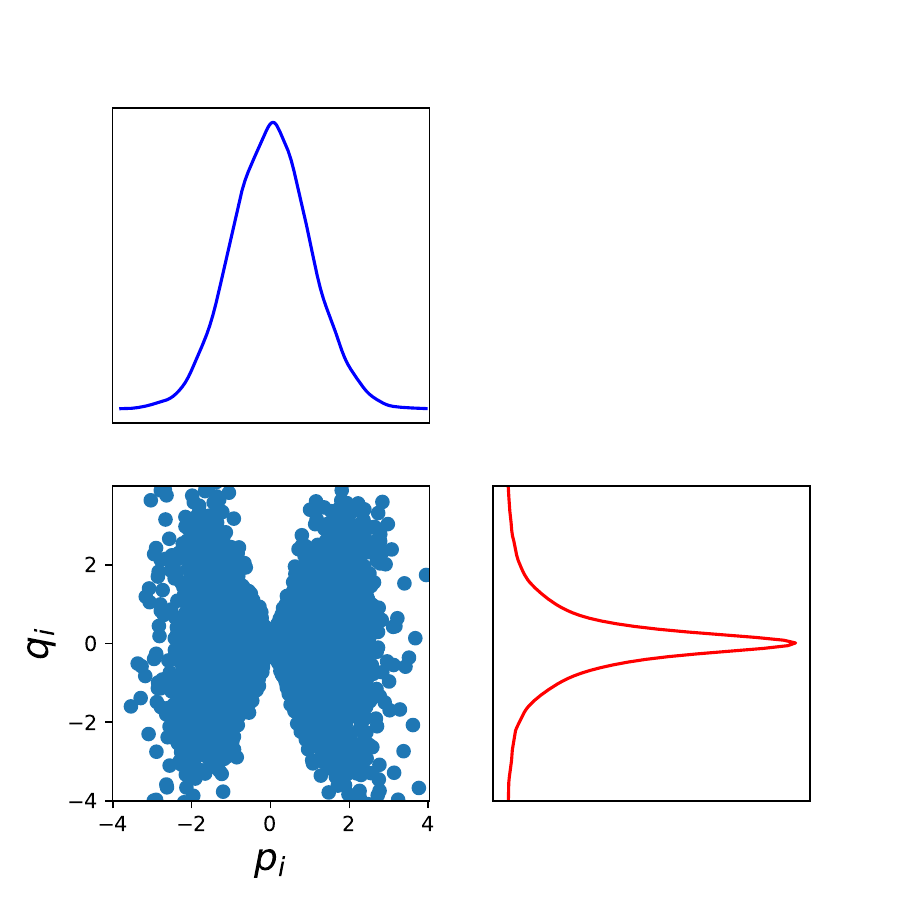}
      \caption{\label{fig:butterfly_density}}
  \end{subfigure}\hspace{-1em}
\caption{(a) The undirected graphical model; (b) Adjacency matrix of true graph (dark blue corresponds to
    an edge, off-white to no edge); (c) One- and two-dimensional marginal densities
    for one pair $(P_i,Q_i)$.\label{fig:modrad}}
\end{figure}

Figures~\ref{fig:modrad_graph}--\ref{fig:modrad_adj} show the minimal I-map and corresponding adjacency matrix of the graph for $r=5$ pairs, with the variables ordered as $P_1, Q_1, \dots, P_5, Q_5$. Figure~\ref{fig:butterfly_density} shows the one- and two-dimensional marginal densities for one pair $(P_i,Q_i)$.
Each one-dimensional marginal is symmetric and unimodal, but the two-dimensional marginal (shown as
samples) displays strongly non-Gaussian behavior.

\subsubsection{Graph identification}
\begin{figure}[!htb]
\centering
\hspace{-0.5em}
\begin{subfigure}{.22\textwidth}
  \centering
  \includegraphics[width=\textwidth,trim={1.9cm 0 2.1 0},clip]{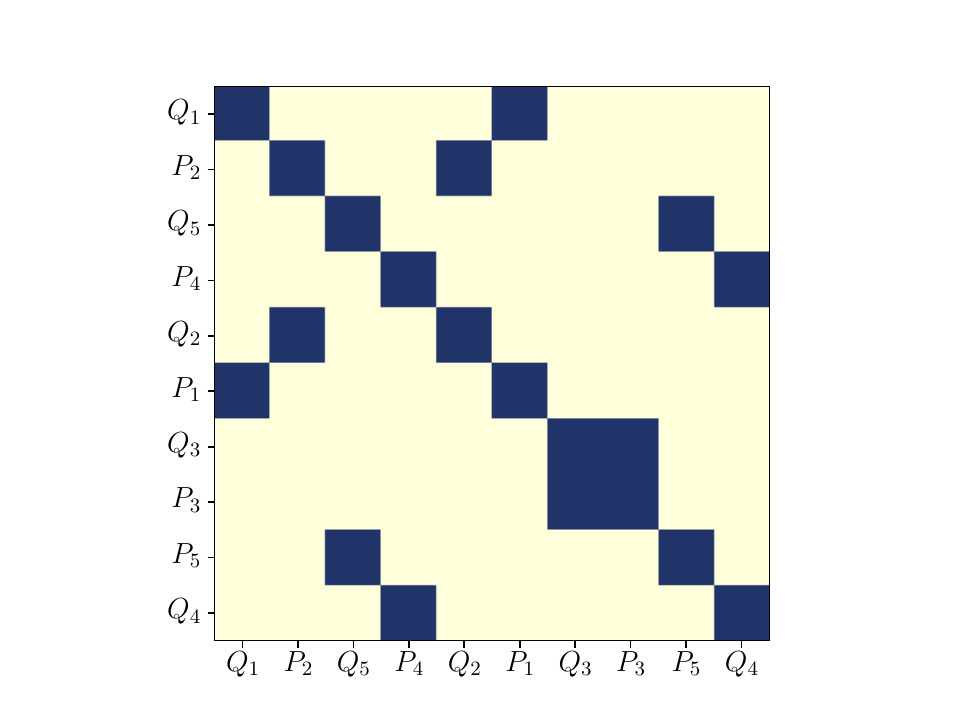}
  \caption{}
\end{subfigure}
\hspace{-1.6em}
\begin{subfigure}{.22\textwidth}
  \centering
  \includegraphics[width=\textwidth,trim={1.9cm 0 2.1 0},clip]{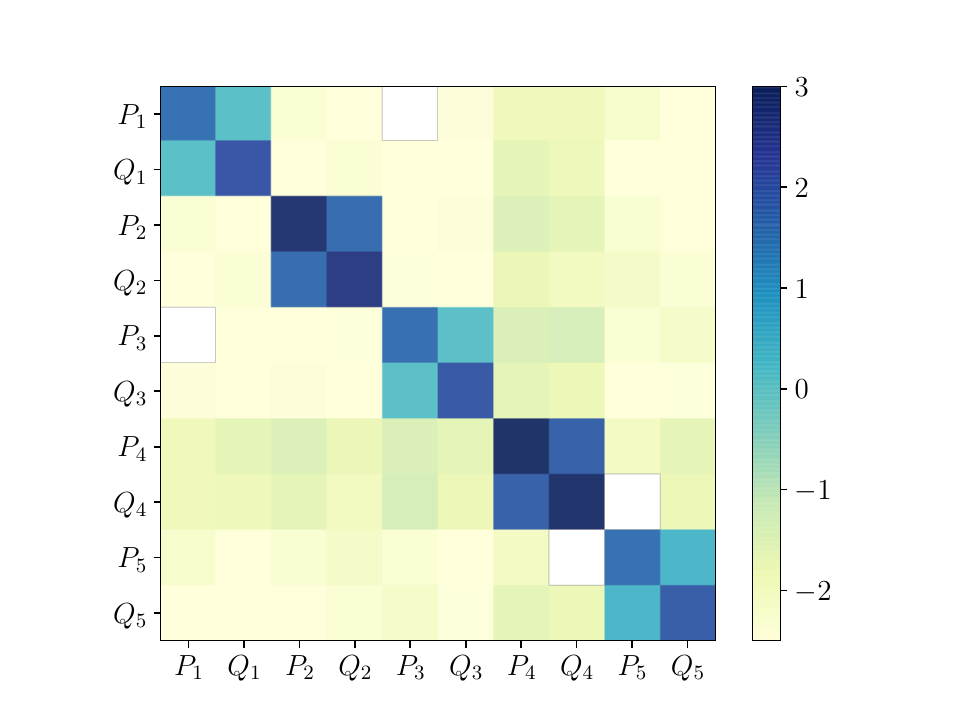}
  \caption{\label{fig:alg-mr_iter1}}
\end{subfigure}
\hspace{-1.3em}
\begin{subfigure}{.22\textwidth}
  \centering
  \includegraphics[width=\textwidth,trim={1.9cm 0 2.1 0},clip]{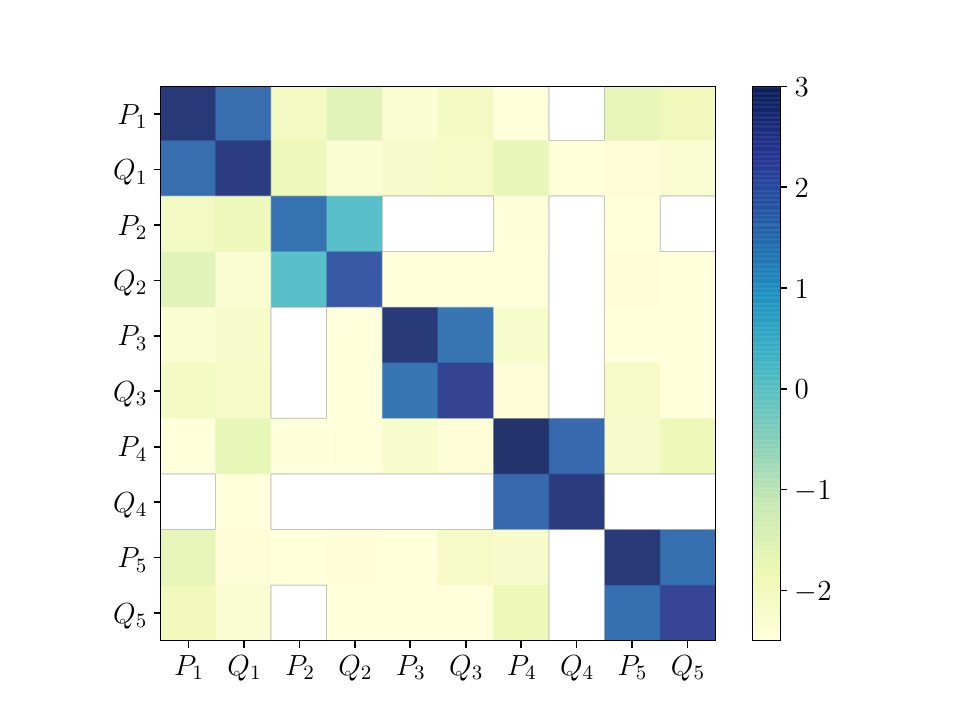}
  \caption{\label{fig:alg-mr_iter2}}
\end{subfigure}
\hspace{-1.3em}
\begin{subfigure}{.22\textwidth}
   \centering
   \includegraphics[width=\textwidth,trim={1.9cm 0 2.1 0},clip]{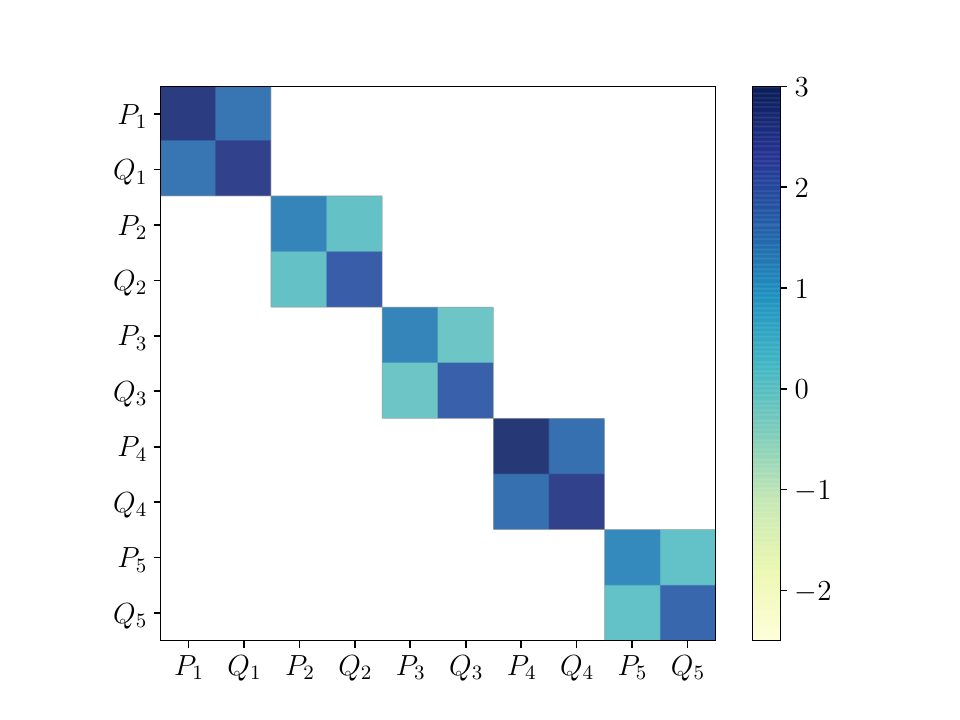}
   \caption{\label{fig:alg-mr_iter3}}
\end{subfigure}
\hspace{-1.6em}
\begin{subfigure}{.22\textwidth}
  \centering
  \includegraphics[width=\textwidth,trim={1.9cm 0 2.1 0},clip]{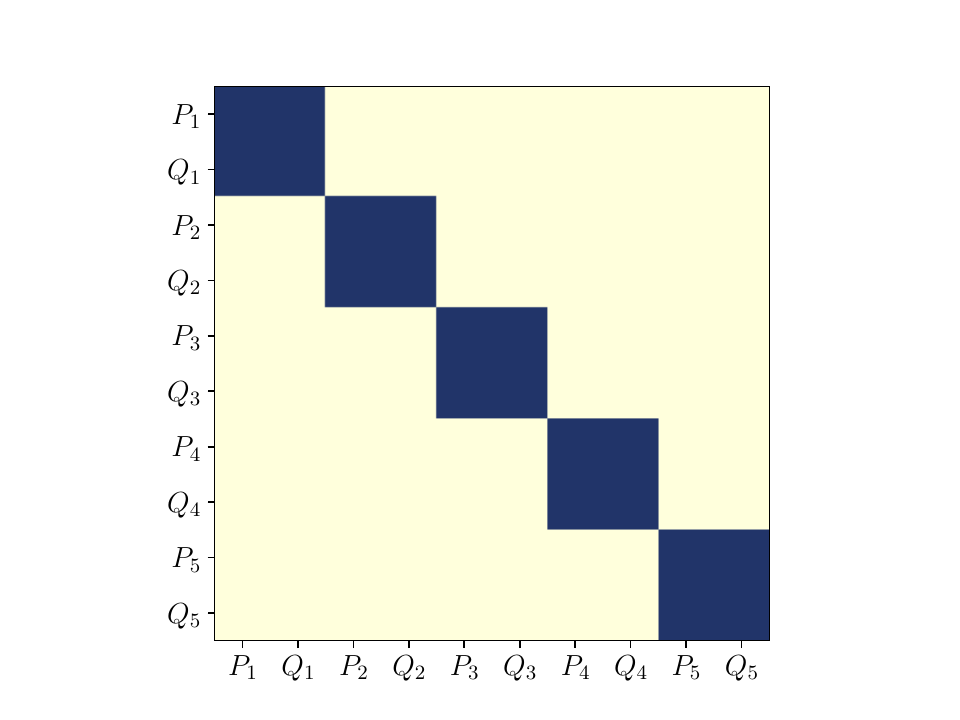}
  \caption{\label{fig:alg-mr_iter4}}
\end{subfigure}%
\hspace{-0.5em}
\caption{(a) Adjacency matrix of original graph under random permutation; (b) Entrywise logarithm of the thresholded score matrix $\overline{\Omega}$ (white indicates a zero entry for $\overline{\Omega}$) after iteration 1; (c) Iteration 3; (d) Iteration 5; (e) Adjacency matrix of the graph after iteration 5. \textsc{sing} returns the correct graph with $\beta=3$.
\label{fig:alg-mr}}
\end{figure}

Figure~\ref{fig:alg-mr} shows the progression of the identified graph (based on the sparsity of the estimated conditional independence score) over the iterations of
\textsc{sing}, with $n=3000$ samples and a polynomial degree $\beta=3$. The variables in the data set are initially permuted, %
to verify that \textsc{sing} identifies a good ordering. After the first iteration of the algorithm (the output of the \textsc{n-sing} algorithm), the estimator for the conditional independence score has the block diagonal pattern in Figure~\ref{fig:alg-mr_iter1}, but the off-diagonals of $\overline{\Omega}$ are not yet zero, resulting in many extra edges. In the next iterations, the algorithm leverages the sparsity of the graph estimated thus far to reveal sparsity in the transport map and improve the estimator for $\Omega$ in Figures~\ref{fig:alg-mr_iter2} and \ref{fig:alg-mr_iter3}, thereby removing all erroneous edges. %
After the
fifth iteration, the sparsity of the graph (and thus the size of the edge set) has not changed and
the algorithm returns the the correct graph in Figure~\ref{fig:alg-mr_iter4}. %

\begin{figure}
    \centering
    \begin{subfigure}{.24\textwidth}
    \centering
    \includegraphics[width=\textwidth,trim={2cm 0 2 0},clip]{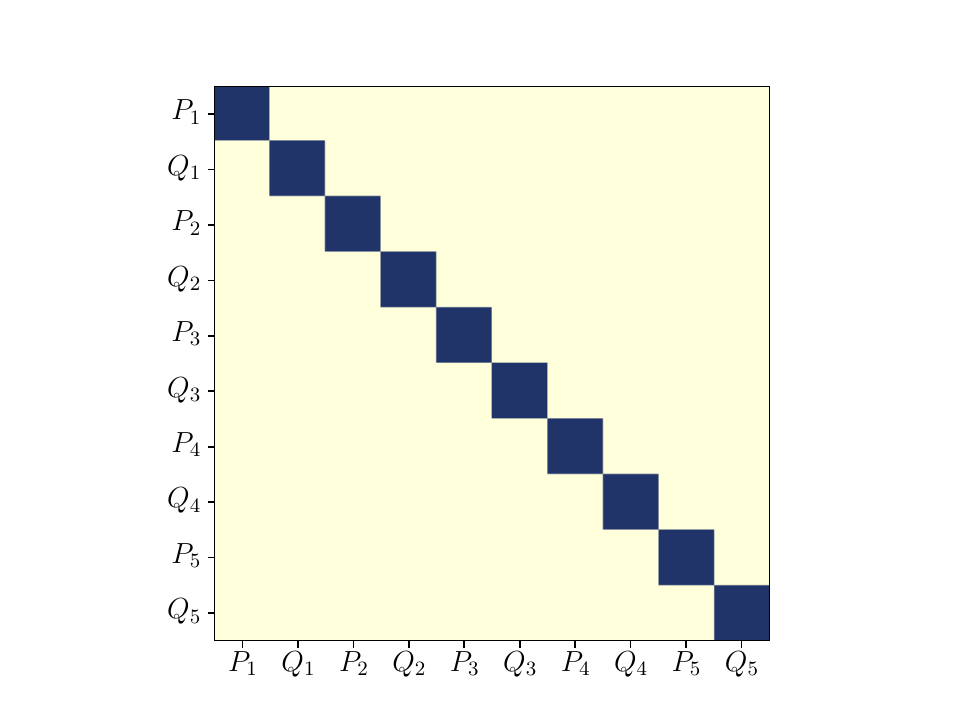}%
    \caption{}
    \end{subfigure}
    \hspace{-1.6em}
    \begin{subfigure}{.24\textwidth}
      \centering
      \includegraphics[width=\textwidth,trim={1.9cm 0 2.1 0},clip]{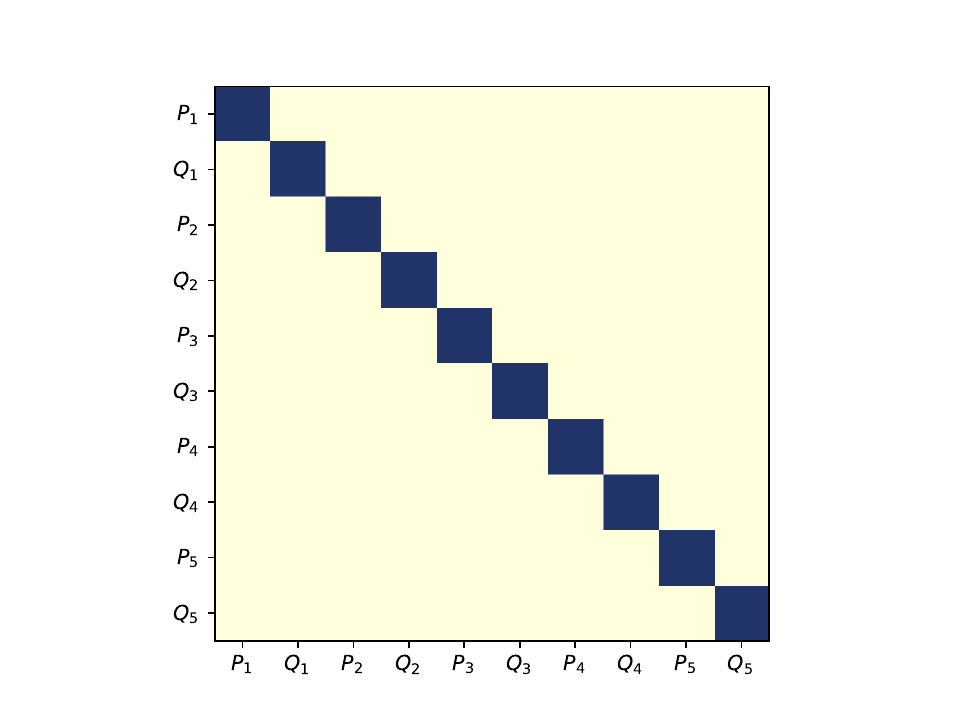}
    \caption{\label{fig:butterfly_npn_ecdf}}
    \end{subfigure}
    \hspace{-1.3em}
    \begin{subfigure}{.24\textwidth}
      \centering
      \includegraphics[width=\textwidth,trim={1.9cm 0 2.1 0},clip]{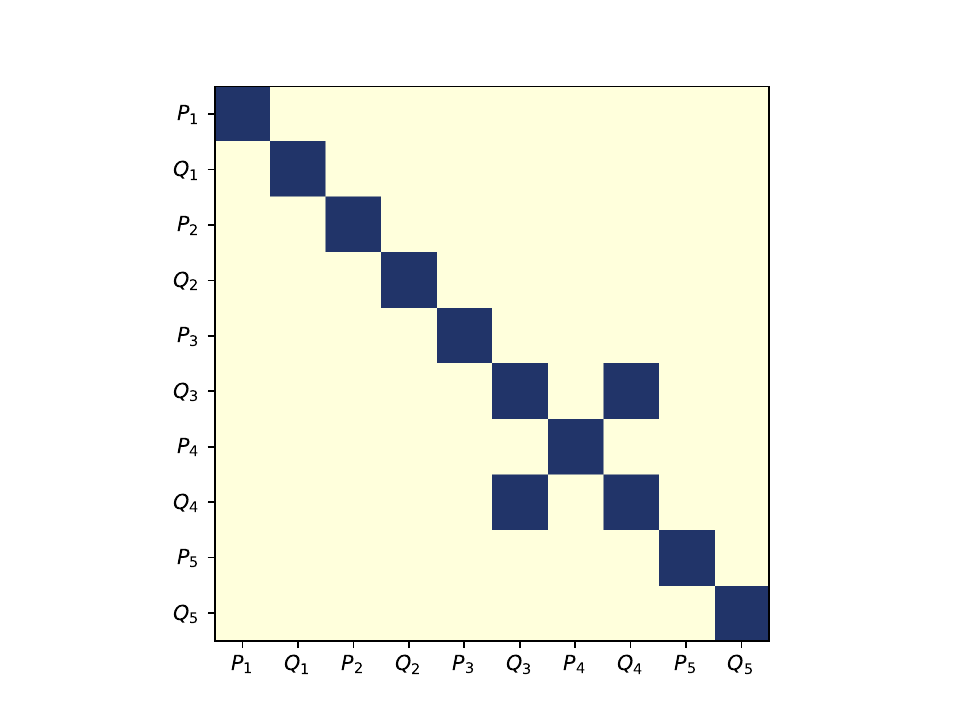}
      \caption{\label{fig:butterfly_npn_spearman}}
    \end{subfigure}
    \hspace{-1.3em}
    \begin{subfigure}{.24\textwidth}
      \centering
      \includegraphics[width=\textwidth,trim={1.9cm 0 2.1 0},clip]{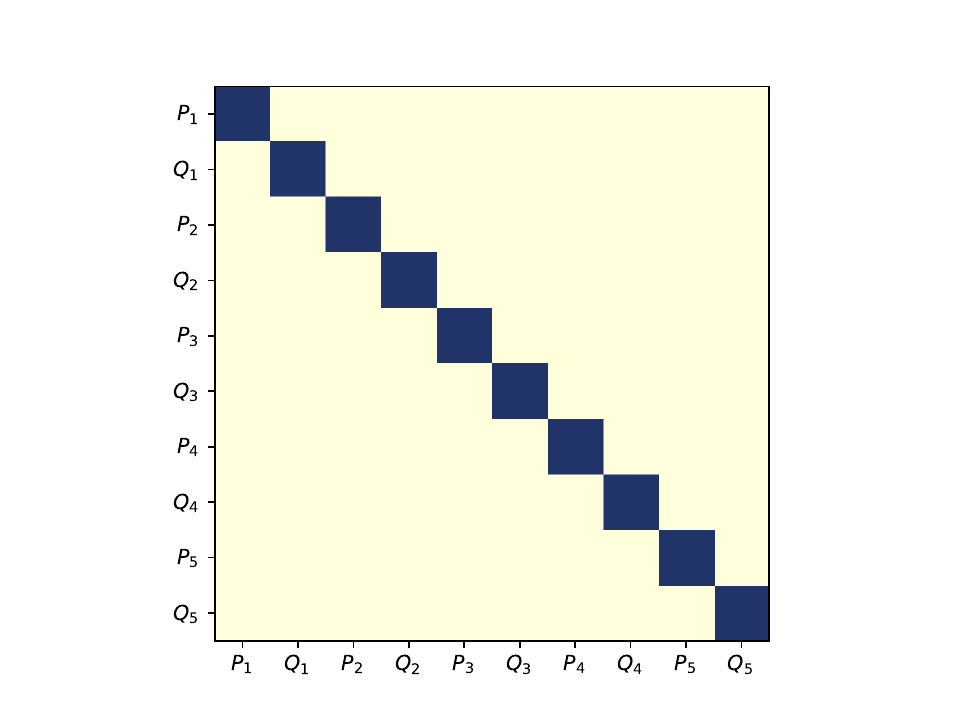}
      \caption{\label{fig:butterfly_npn_ktau}}
    \end{subfigure}
    \caption{\rev{When the data are assumed (a) normal (i.e., using $\beta = 1$) or nonparanormal using (b) empirical CDF marginal transformations, (c) Spearman correlations or (d) Kendall-Tau correlations, all ten variables are incorrectly found to be independent or to have spurious dependencies}.\label{fig:modrad-d}}
\end{figure}

In contrast, assuming that 
\rev{the data follow a normal or nonparanormal distribution}
returns the incorrect graph, as illustrated by the experiments in Figure~\ref{fig:modrad-d}. If the data were normal, then a linear map ($\beta = 1$) would suffice.
However, \textsc{sing} with $\beta=1$ (Figure~\ref{fig:modrad-d}(a)) yields not only an incorrect graph, but in fact fails to
detect any edges at all. This result does not improve with increasing $n$, and incorrectly implies that all ten variables are marginally independent. Furthermore, this result indicates that methods based on precision matrix estimation that assume the data to be Gaussian (e.g., \textsc{glasso}) will also fail to recover the correct graph in this example, as demonstrated in \cite{morrison2017beyond}. \rev{ Similarly, algorithms that assume the data to be Gaussian after a (possibly nonlinear) diagonal transformation of the marginals do not recover the correct graph for this dataset; see Figures~\ref{fig:butterfly_npn_ecdf}-\ref{fig:butterfly_npn_ktau}. %
These algorithms transform the data marginal using either the empirical cumulative distribution functions (CDFs) in Figure~\ref{fig:butterfly_npn_ecdf}, or rank-based quantities such as the Spearman's rho and Kendall's tau statistics (that are only based on the sample ordering and do not require computing CDFs) in Figures~\ref{fig:butterfly_npn_spearman} and~\ref{fig:butterfly_npn_ktau}, respectively. After computing these transformations,~\citet{liu2009nonparanormal,liu2012nonparanormal} showed that a \textsc{glasso} estimator applied with a correlation coefficient matrix of the transformed data will consistently estimate the graph for nonparanormal data. For the distribution considered in Figure~\ref{fig:modrad-d} which is outside of this class, however, these algorithms consistently miss the edges between each pair of non-Gaussian variables.} %

Another important comment on this example is that the underlying density estimation problem is quite challenging: the exact transport map that characterizes $\pi$ in this
case would in fact require an infinite expansion of polynomials. That is,
no transport map with finite polynomial degree $\beta$ can perfectly represent $\pi$. However, the
correct graph can be identified with $\beta=3$. Thus, in this case, allowing for some nonlinearity
in the map is sufficient. A linear map does not work at all, but $\beta$ also need not be
excessively high.\footnote{A tutorial on this example is provided online by~\citet{tmv1}.}

\subsubsection{Computational cost}
\label{sec:costsmain}
\rev{Here we provide details on the runtime and memory requirements of applying \textsc{sing} to the butterfly dataset. %
We have observed that memory usage is relatively consistent across different data sets and computing setups. Runtime can vary more widely, although it tends to be on the order of seconds to just a few minutes.}

\rev{The tables in Appendix~\ref{app:memory} provide specific memory usage with $n = 1000$ samples for different dataset dimensions $d$ and maximum polynomial degrees $\beta$. The tables break down the memory required for: (i) maximum likelihood estimation of the map coefficients, (ii) estimating the score, and (iii) estimating the variance of the score, all for the first two iterations of \textsc{sing}. We observe that variance estimation (see equation \eqref{eq:rho}) is the main contributor to memory usage. This is not surprising because the variance estimate relies on computing derivatives of $\widetilde{\Omega}$ with respect to map parameters, and $\widetilde{\Omega}$ already contains second derivatives of the log density with respect its arguments. 
Evaluations of the resulting derivatives rely on a tensor whose size has quartic growth with respect to the number of variables $d$. %
We also note that the first iteration of \textsc{sing} uses the most memory by far. After each iteration, \textsc{sing} may remove active variables from the map, making subsequent iterations less memory intensive. During the first iteration, we do not assume any sparsity of the map or the graph, and hence this iteration is the most costly.}

\rev{Though our current implementation of the algorithm can be memory intensive, we emphasize that there are no theoretical limits to recovering the exact graph with \textsc{sing} (assuming the transport map class is appropriate); indeed, even the one-step algorithm will recover the correct graph in the limit of infinite samples. (This is in contrast with many existing methods that make distributional assumptions, usually related to Gaussianity.) 
Though our present focus is on establishing essential features of the provably consistent algorithm, it is natural to then seek approximations that could lessen computational cost in practice. Some options are to use a lower $\beta$ than absolutely necessary, to devise further approximations to the variance of $\widetilde{\Omega}$, or to exploit prior knowledge that imposes some sparsity on the initial transport map. Another natural extension, which would significantly reduce memory usage, is to devise a \emph{local} version of \textsc{sing} that employs neighborhood selection. We comment further on this possibility in Section~\ref{sec:con}.} %

\subsection{Nonparanormal data: Gaussian CDF and power transformations}\label{ssec:nonp}
\begin{figure}[!tbh]
     \begin{subfigure}[b]{.45\textwidth}
       \includegraphics[width=\textwidth]{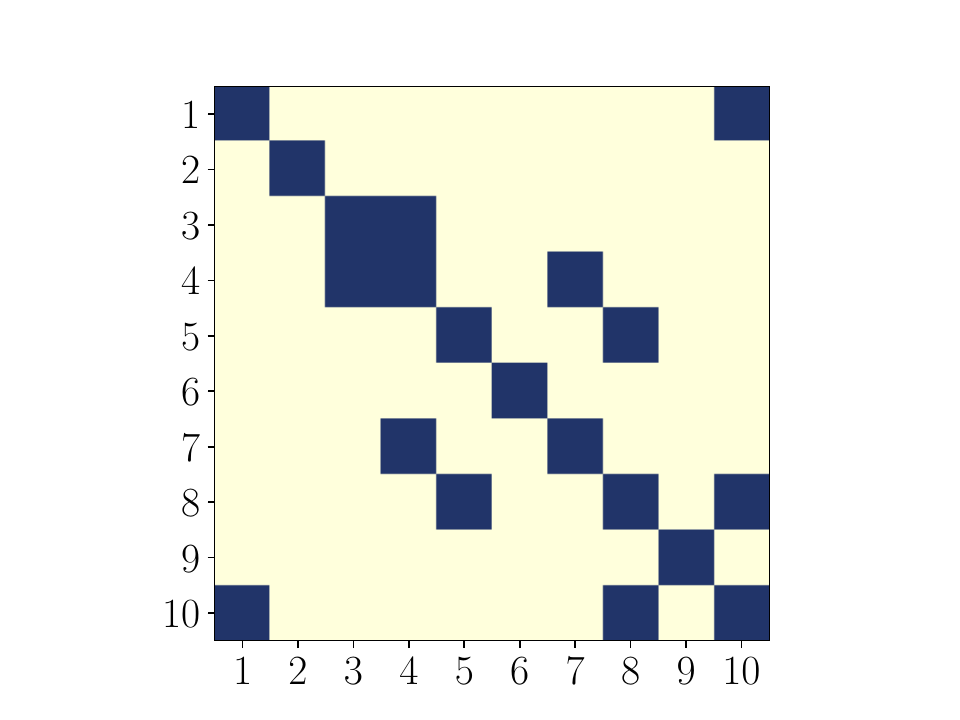} %
       \caption{\label{fig:cdf-graph}}
     \end{subfigure}
     \hspace{0.5cm}
     \begin{subfigure}[b]{.45\textwidth}
       \includegraphics[width=0.85\textwidth]{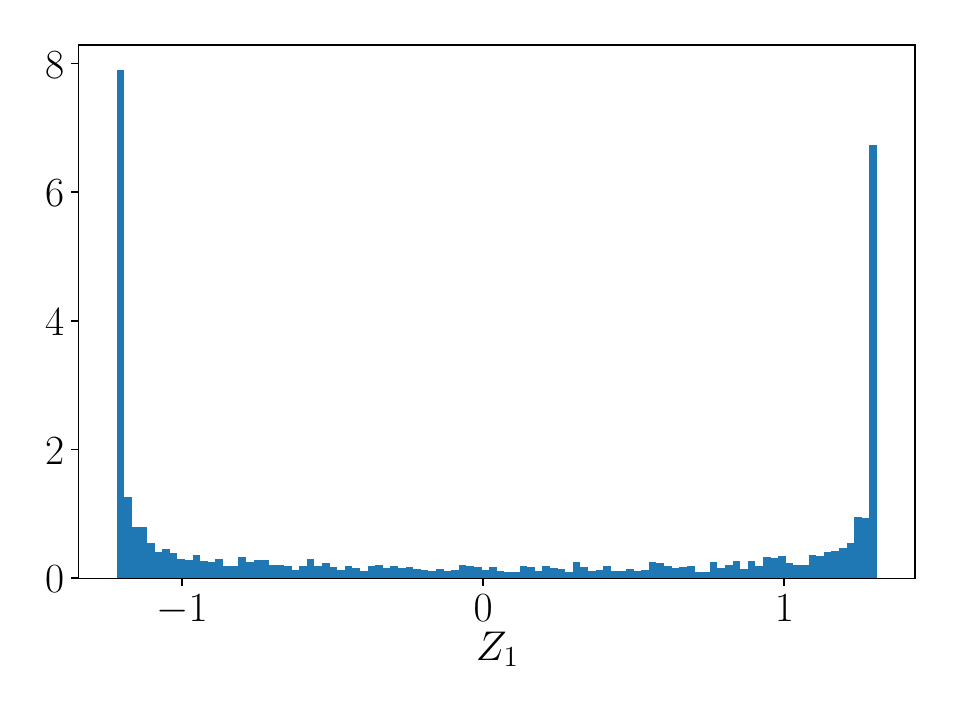} %
       \caption{\label{fig:nonphist}}
     \end{subfigure}
    \hfill
     \caption{(a) The true minimal I-map of $\bm{\nu}_\pi$. (b) A histogram of one of the marginals from the nonparanormal dataset. In this case, the
     data has bimodal, non-Gaussian behavior.}
 \end{figure}

Now we test \textsc{sing} on data from nonparanormal distributions with arbitrary Markov structures. Let $D^k: \mathbb{R} \rightarrow \mathbb{R}$ %
be a monotone and differentiable transformation for $k=1,\dots,d$. Following~\cite{liu2009nonparanormal}, we say that $\bm{Z} = (Z_1, Z_2, \dots, Z_d)$ has a nonparanormal distribution with measure $\bm{\nu}_\pi$ and density $\pi$ when %
$D(\bm{Z}) = (D^1(Z_1), D^2(Z_2), \dots, D^d(Z_d))$ follows a multivariate Gaussian distribution %
with measure $\bm{\nu}_\rho$ and density $\rho = \mathcal{N}(\bm{0},\Sigma)$. We refer to this Gaussian as the ``base'' distribution in the following subsections.  %
The transformation $D$ acts component-wise on each variable, and thus defines a diagonal transport map that pushes forward $\pi$ to $\rho$ (or equivalently $D$ pulls back $\rho$ to $\pi$). From the result in Proposition~\ref{prop:diag} for diagonal maps, the 
minimal I-maps of $\bm{\nu}_\pi$ and of $\bm{\nu}_\rho$ are equivalent. %
Thus the Markov structure of the target distribution %
is prescribed by the sparsity of the precision matrix $\Sigma^{-1}$ of $\rho$, %
but the data can have very non-Gaussian features (see Figure~\ref{fig:nonphist} for one example).  Within the field of structure learning from non-Gaussian data, this type of nonparanormal distribution (also known as a Gaussian copula) is a test case for algorithms that handle non-Gaussianity. %

To generate data from a nonparanormal distribution with an arbitrary graph structure, we follow the steps in~\citet{liu2009nonparanormal}.
First, a random sparse graph $\mathcal{G} = (V, E)$ is generated. For each node $i \in (1, \dots, d)$, we associate a pair of random variables $(Y_i^{(1)}, Y_i^{(2)}) \in [0,1]^2$ to $i$ where
\begin{equation}
    Y_1^{(l)}, Y_2^{(l)},\dots Y_d^{(l)} \sim \mathcal{U}[0,1]
\end{equation}
for $l = 1,2$. Then, each pair of nodes $(i,j)$ is included in the edge set $E$ with probability
\begin{equation}
    \mathbb{P}((i,j) \in E) = \frac{1}{\sqrt{2 \pi}}\exp\left(\frac{\|y_i - y_j\|_{2}^2}{2s}\right),
\end{equation}
where $s$ is a parameter that controls the sparsity of the graph, $y_k \equiv
(y_k^{(1)},y_k^{(2)})$ is a sample of $(Y_k^{(1)}, Y_k^{(2)})$, and $\| \cdot\|_{2}$
represents the Euclidean norm. In our numerical experiments we set $s = 3$ and limit the maximum degree of the graph, i.e., the number of the edges incident to each node, to be four. %
A realization of a graph generated according to this procedure is shown in Figure~\ref{fig:cdf-graph}. After defining the graph, the entries of the inverse covariance 
$\Sigma^{-1}$ are given by:
\begin{equation}
    \Sigma^{-1}_{ij} = \begin{cases} 1 \quad& i = j\\
        0.245 \quad& (i,j)\in E\\
        0 \quad &\text{otherwise}.
    \end{cases}
\end{equation}
We note that with maximum degree four, the value $0.245$ ensures that the inverse covariance matrix is positive definite by the Gershgorin circle theorem. %

To sample from $\bm{\nu}_{\pi}$, we generate i.i.d.\thinspace samples $\mathbf{x}^{l}$ from $\rho$ and apply the inverse diagonal transformation to generate i.i.d.\thinspace samples $\mathbf{z}^{l} = D^{-1}(\mathbf{x}^{l})$ from $\pi$. In this work we consider two possibilities for the function $D$ %
as in~\cite{liu2009nonparanormal}: a scaled Gaussian CDF and a power transformation. We now detail how we construct the two functions $D^k$.

\paragraph{Gaussian CDF transformation.}  Let $f_0\colon \mathbb{R} \rightarrow [0,1]$ be the univariate Gaussian CDF with mean $\mu_{f_0}$ and standard deviation $\sigma_{f_0}$, i.e.,
    \begin{equation}
        f_0(t) = \Phi \left(\frac{t - \mu_{f_0}}{\sigma_{f_0}} \right),
    \end{equation}
        where $\Phi$ is the univariate standard Gaussian CDF. The inverse CDF transformation $F^{k} = (D^{k})^{-1}$ applied to the $k$th variable is defined as:
    \begin{equation} \label{eq:CDF_transformation}
    F^{k}(x_k) = 
        \sigma_{k} \left( \frac{f_0(x_k) - \int f_0(t) \Phi'{\textstyle \left(\frac{t -
    \mu_k}{\sigma_k}\right)}dt}{\sqrt{\int\left(f_0(y) - \int f_0(t) \Phi'{\textstyle \left(\frac{t - \mu_k}{\sigma_k}\right)}dt
    \right) ^2 \Phi'{\textstyle \left(\frac{y - \mu_k}{\sigma_k}\right)}dy}}\right) + \mu_k.\end{equation}
    In our experiments we apply the same transformation to each marginal by setting  $\mu_{f_0} = 0.05$, $\sigma_{f_0} = 0.4$, $\mu_k = 0$ and $\sigma_k =
    \sqrt{\Sigma_{kk}}$. %
    A %
representative marginal PDF of $\pi$ is shown (as a histogram) in Figure~\ref{fig:nonphist}. Each marginal displays very non-Gaussian behavior as a result of the nonlinearity in the function $F^{k}$.

\paragraph{Power transformation.} Let $f_0(t)= \text{sign} (t) |t|^a$, for $a>0$. The inverse power transformation $F^{k} = (D^{k})^{-1}$ applied to the $k$th variable is defined as
\begin{equation} F^{k}(x_k) = \sigma_k \left( \frac{f_0(x_k - \mu_k)}{\sqrt{\int (f_0(t - \mu_k))^2
\Phi\left(\frac{t-\mu_k}{\sigma_k}\right)dt}}\right) + \mu_k .\end{equation}
We set $\mu_k$ and $\sigma_k$ as above in the CDF transformation, and set $a = 3$.\\

\begin{figure}[tbh]
    \centering
 \begin{subfigure}{.24\textwidth}
   \centering
     \includegraphics[width=\textwidth,trim={1.9cm 0 2.1 0},clip]{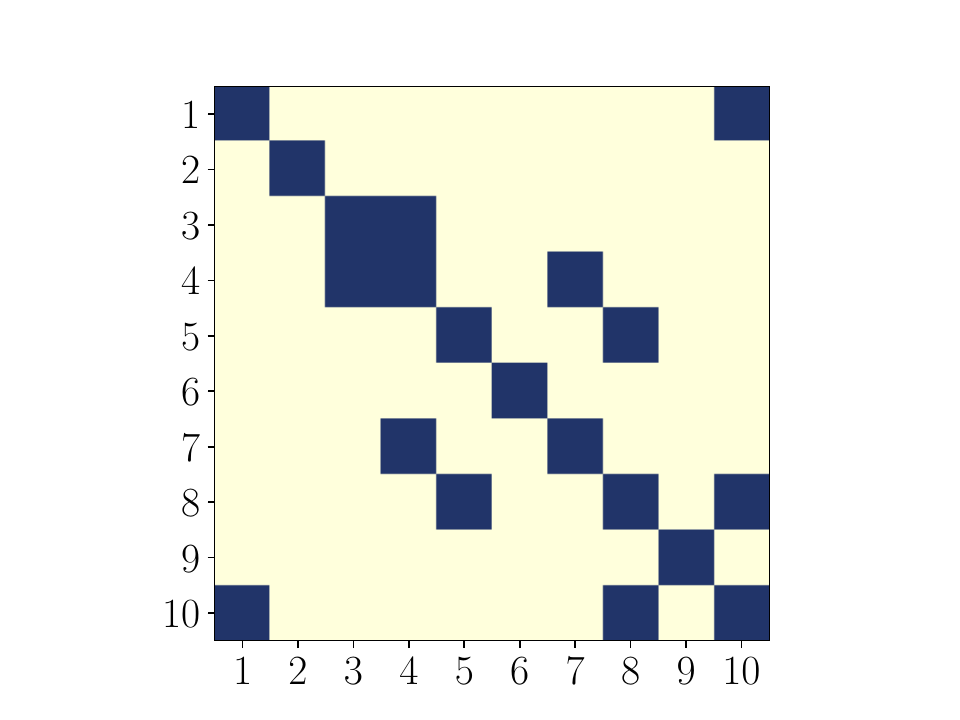}
     \caption{\label{fig:nonp-beta1}}
 \end{subfigure}
    \hfill
 \begin{subfigure}{.24\textwidth}
   \centering
     \includegraphics[width=\textwidth,trim={1.9cm 0 2.1 0},clip]{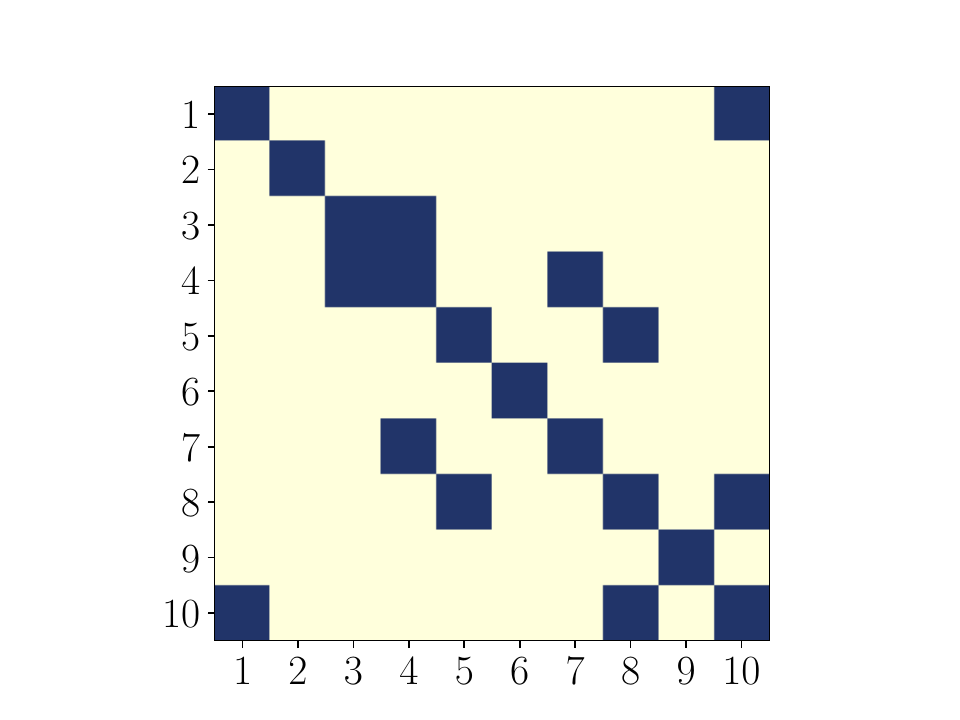}%
     \caption{\label{fig:nonp-beta2}}
 \end{subfigure}
    \hfill
 \begin{subfigure}{.24\textwidth}
   \centering
     \includegraphics[width=\textwidth,trim={1.9cm 0 2.1 0},clip]{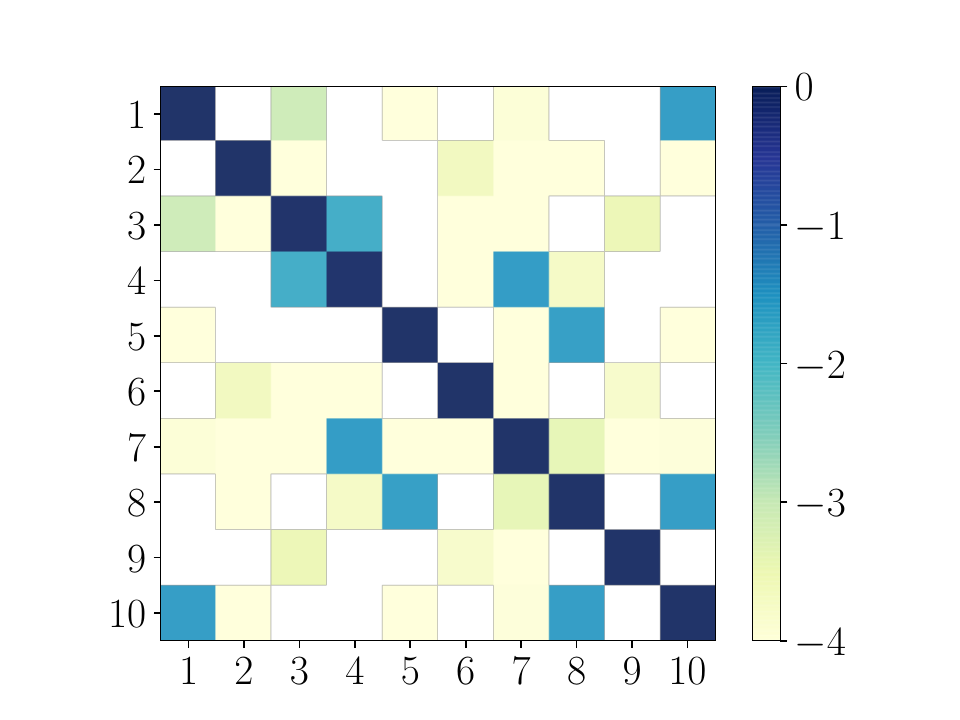} %
     \caption{\label{fig:nonp-iv}}
 \end{subfigure}
    \hfill
 \begin{subfigure}{.24\textwidth}
   \centering
     \includegraphics[width=\textwidth,trim={1.9cm 0 2.1 0},clip]{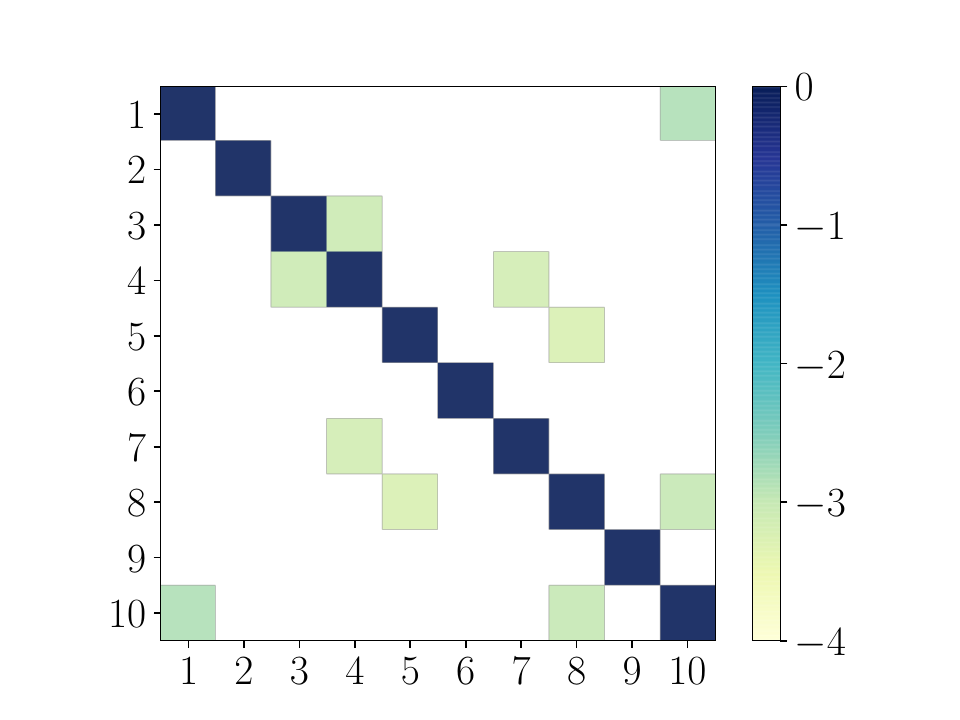} %
     \caption{\label{fig:nonp-omega-beta1}}
 \end{subfigure}
    \caption{(a) Recovered graph with $\beta = 1$. The correct graph is returned. (b) The recovered
    graph with $\beta = 2$ is also correct, but Hermite functions of polynomial degree two in the transport map are
    unnecessary. (c) Entrywise logarithm of the inverse of the empirical covariance matrix of the data. (d) Entrywise logarithm of $\overline{\Omega}$ for $\beta = 1$.\label{fig:nonpresults}}
 \end{figure}

Surprisingly, the conditional independence properties of these nonparanormal distributions are recovered with a linear map. As
seen in Figure~\ref{fig:nonp-beta1} for the Gaussian CDF transformation, we recover the correct graph with $n = 3000$ samples and a polynomial degree $\beta = 1$.
Figure~\ref{fig:nonp-beta2} shows that the correct graph is also returned with $\beta=2$. While a transport map with higher polynomial degree could be used, a biased approximation to $\pi$ based on a $\beta=1$ map 
in this case is sufficient for learning the graph. %
We note that the dominant entries
of the inverse of the empirical covariance matrix of the data, shown in Figure~\ref{fig:nonp-iv}, for this example also reveal the true graph---just as the sparsity of the precision matrix would in the Gaussian case. The next subsection will
investigate this connection further. %
But the inverse of the empirical covariance contains many noisy and non-zero entries as compared to the final score matrix $\overline{\Omega}$, as shown in Figure~\ref{fig:nonp-omega-beta1};
the latter benefits from the thresholding process in \textsc{sing} and the resulting sparsification of the transport map used to estimate the target density.

\begin{figure}[tbh]
    \centering
 \begin{subfigure}{.45\textwidth}
   \centering
     \includegraphics[width=\textwidth]{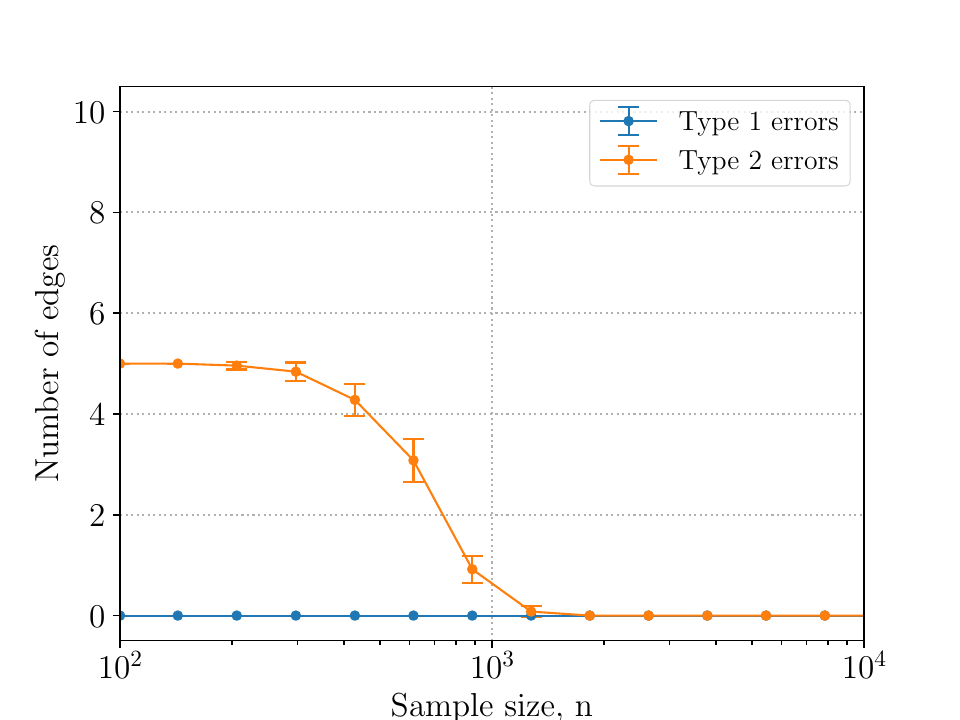}
     \caption{\label{fig:err-cdf}}
 \end{subfigure}
  \hspace{0.5cm} 
 \begin{subfigure}{.45\textwidth}
   \centering
     \includegraphics[width=\textwidth]{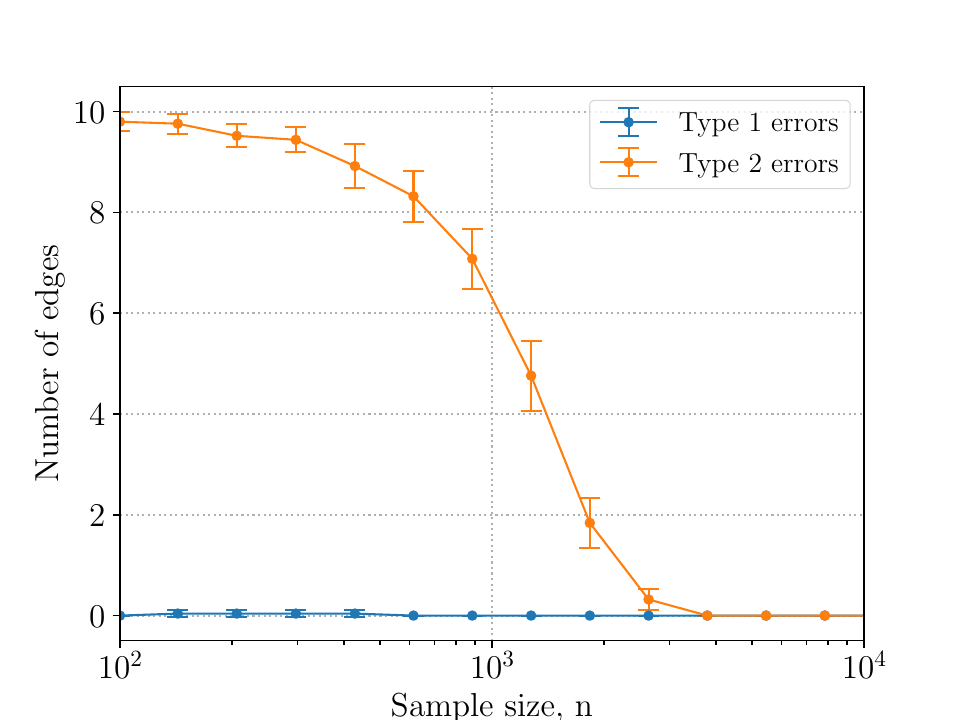}
     \caption{\label{fig:err-power}}
 \end{subfigure}
    \caption{Type 1 and 2 errors for recovering the graph of a nonparanormal distribution versus sample size $n$ for the (a) Gaussian CDF, and (b) power transformation using $\beta=1$. \label{fig:nonpresults_vsN}}
 \end{figure}

Figure~\ref{fig:nonpresults_vsN} shows the errors made by the \textsc{sing} algorithm versus the sample size $n$ for both the Gaussian CDF and the power transformation, with $\beta = 1$. 
For both transformations, the number of type 1 errors (i.e., erroneous edges that are not present in the true graph) is always zero for all $n$, and the number of type 2 errors (i.e., undetected edges that are present in the true graph) decreases to zero as $n$ increases.

\subsection{Nonparanormal data: cubic transformation}\label{ssec:x3}%
In this section we investigate the effect of biased approximations to the target density when learning the graph. We consider a simplified form of the power transformation applied to a 3-dimensional Gaussian vector $\bm{X} \in \mathbb{R}^{3}$. Let $\bm{X} \sim \rho = \mathcal{N}(\mathbf{0},\Sigma_{\rho})$ where the precision and the covariance matrix are given by
\begin{equation}
\Sigma_{\rho}^{-1} = \begin{bmatrix}
1 & 0.2 & 0 \\
0.2 & 1 & 0.2 \\
0 & 0.2 & 1
\end{bmatrix}, \;\;\;\; \Sigma_{\rho} = 0.92 \begin{bmatrix}
0.96 & -0.2 & 0.04 \\
-0.2 & 1 & -0.2 \\
0.04 & -0.2 & 0.96
\end{bmatrix}
\end{equation}
From the sparsity pattern of the precision matrix, the random variables satisfy $X_{1} \ci X_{3} | X_{2}$. The
corresponding graph %
is a chain connecting nodes $X_1$ to $X_2$ and $X_2$ to $X_3$.

We now consider a component-wise monotonic transformation (i.e., a diagonal transport map) %
given by $F^{k}(x_{k}) = x_{k}^3$ for $k=1,2,3$. The transformed random variable $\bm{Z} = F(\bm{X}) = (F^{1}(X_{1}), F^{2}(X_{2}), F^{3}(X_3))$ is non-Gaussian with density $\pi = F_{\sharp}\rho$. By the result of Proposition~\ref{prop:gpnp} for diagonal transformations, $\pi$ satisfies the conditional independence property $Z_{1} \ci Z_{3} | Z_{2}$, and the Markov structure of $\pi$ is equivalent to the Markov structure of $\rho$. 

To characterize the target density, suppose we approximate $\pi$ by the pullback of a standard Gaussian density through a triangular transport map with polynomial degree $\beta = 1$. %
Using Proposition~\ref{prop:mo} with $n \rightarrow \infty$, the approximate density is a multivariate Gaussian distribution $\mathcal{N}(\mu_{\pi},\Sigma_{\pi})$ with mean
$\mu_{\pi} = \mathbb{E}_{\pi}[\bm{Z}] = \mathbb{E}_{\rho}[F(\bm{X})] = \bf{0}$
and covariance $\Sigma_{\pi} = \mathbb{E}_{\pi}[\bm{Z} \bm{Z}^{T}] = \mathbb{E}_{\rho}[F(\bm{X})F(\bm{X})^T]$.  Each entry
of the covariance matrix can be expressed analytically in terms of $\sigma_{ij} = (\Sigma_{\rho})_{ij}$ as
\begin{equation}\label{eq:tmp2645}
(\Sigma_{\pi})_{ij} = 9\sigma_{ii}\sigma_{jj}\sigma_{ij} + 6\sigma_{ij}^3,
\end{equation}since each entry is in fact a higher-order moment of the multivariate Gaussian distribution $\bm{\nu}_{\rho}$. As a result, we can also analytically compute the inverse of $\Sigma_{\pi}$, which is given by
\begin{equation}
\Sigma_{\pi}^{-1} = \begin{bmatrix}
5.96 \times 10^{-2} & 6.99 \times 10^{-3} & -5.56 \times 10^{-4} \\
6.99 \times 10^{-3} & 5.36 \times 10^{-2} & 6.99 \times 10^{-3} \\
-5.56 \times 10^{-4} & 6.99 \times 10^{-3} & 5.96 \times 10^{-2} \\
\end{bmatrix}, \label{eq:prec_gauss_approx}
\end{equation}
up to three significant digits.
The (1,3) entry of $\Sigma_\pi^{-1}$ is not zero, so in principle a Gaussian approximation to $\pi$ will \emph{not} recover 
the correct graph. The (1,3) entry is still quite small, however---and, importantly, the relative magnitudes of entries in
$\Sigma_\pi^{-1}$ are similar to those in $\Sigma_\rho^{-1}$  \rev{(see \cite{morrison2022diagonal} for a more theoretical explanation for why these entries are still small)}. Thus, when using a numerical approximation, the small $(1,3)$ entry
can easily be thresholded and set to zero. And after this thresholding, the correct graph is in fact returned. 

\begin{figure}[!tbh]
  \centering
  \begin{subfigure}{.45\textwidth}
    \centering
    \includegraphics[width=\textwidth]{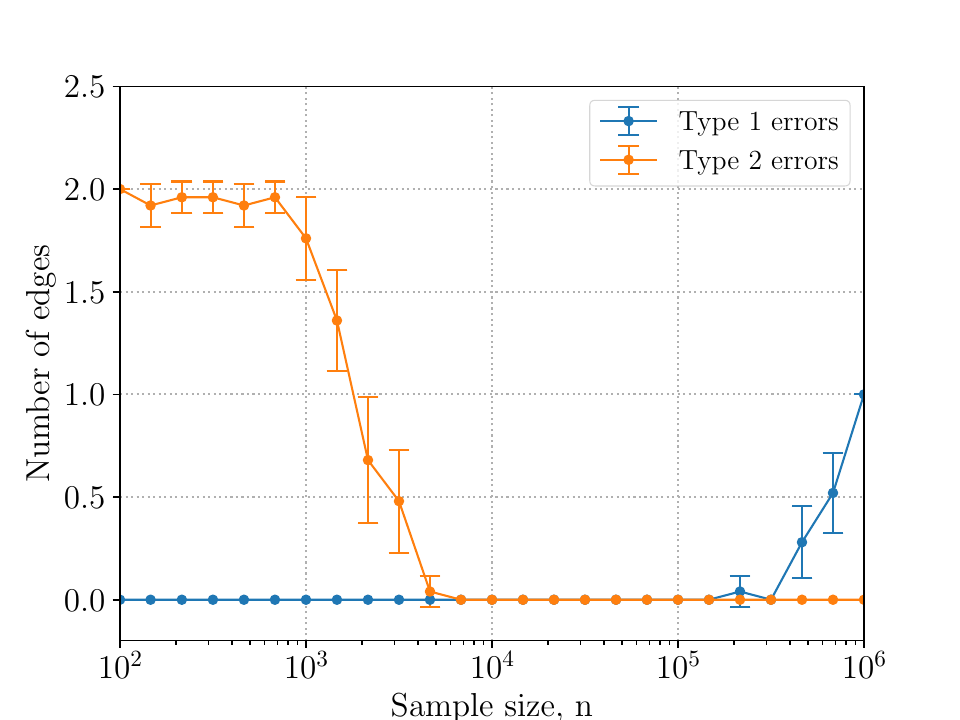}
    \caption{\label{fig:cubic_error_beta1}}
  \end{subfigure}
  \hspace{0.5cm} 
  \begin{subfigure}{.45\textwidth}
    \centering
    \includegraphics[width=\textwidth]{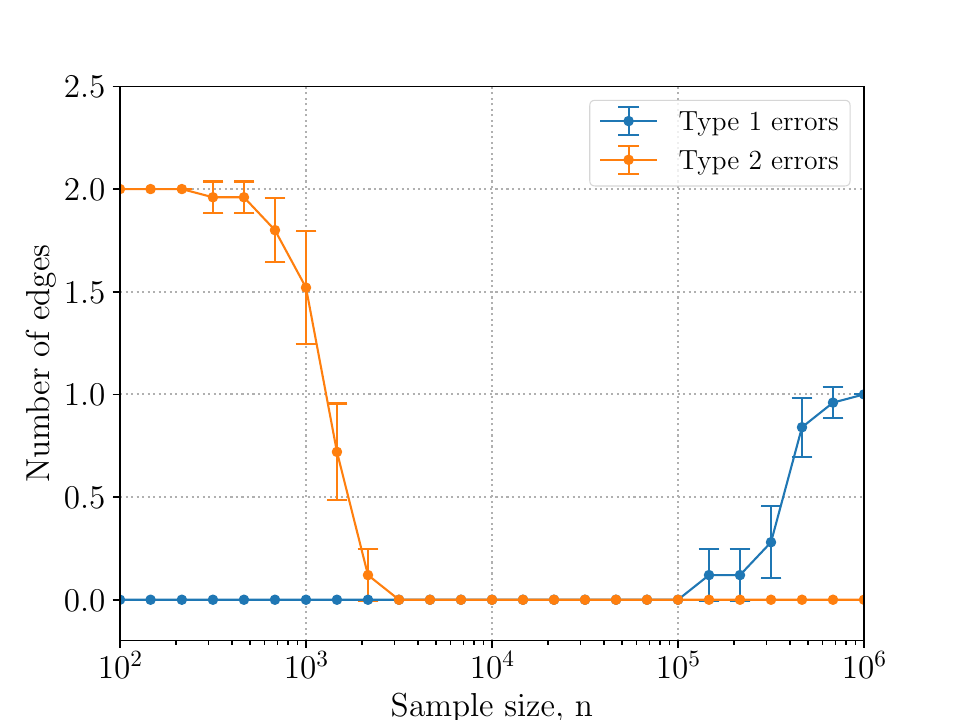}
    \caption{\label{fig:cubic_error_beta2}}
  \end{subfigure}
  \caption{Type 1 and 2 errors for recovering the graph of the non-paranormal distribution with the cubic transformation versus sample size $n$ using (a) $\beta = 1$, and (b) $\beta = 2$. \label{fig:cubic_results_samplesize}}
\end{figure}

Figures~\ref{fig:cubic_error_beta1} and~\ref{fig:cubic_error_beta2} investigate this phenomenon, by showing the errors made by \textsc{sing} for sample sizes $n \in [10^2, 10^6]$ using $\beta = 1$ and $\beta = 2$, respectively. %
The correct graph is returned for a broad range of sample sizes.
This is a case where the finite-sample estimators of $S$ and $\Omega$ and an under-parameterized map (i.e., a biased approximation to $\pi$) interact in a surprisingly beneficial way to correctly learn the graph. We believe the same phenomenon explains the success of
the linear $\beta = 1$ transport maps in the previous subsection. When the sample size becomes large enough to resolve the smallest entries in the precision matrix \eqref{eq:prec_gauss_approx} with sufficiently high confidence, however, we observe in Figure~\ref{fig:cubic_results_samplesize} that the \textsc{sing} algorithm with $\beta = 1$ also includes $(1,3)$ in the estimated edge set of the graph. %
A similar trend is observed for $\beta = 2$. Note that the class of $\beta = 2$ transport maps is still insufficient to 
exactly capture $\pi$ in this example (the exact transport map would be the composition of a linear map with a diagonal map that applies component-wise cubic root transformations). For sufficiently large $n$, the bias for $\beta = 2$ yields the type 1 errors seen at the right of Figure~\ref{fig:cubic_error_beta2}. 

\subsection{Diagonal transformations of a non-Gaussian base distribution}
\label{ssec:beta2}

In this example, we consider the class of \rev{probability distributions with measure} $\bm{\nu}_{\pi}$ defined as pullbacks of a \textit{non-Gaussian} base \rev{measure} $\bm{\nu}_{\rho}$ through a nonlinear diagonal transport map. When the base measure $\bm{\nu}_{\rho}$ is represented as the pullback of a standard Gaussian measure through a triangular transport map that uses polynomials of maximum degree $\beta$, we refer to $\bm{\nu}_{\rho}$ as a \rev{distribution} in base class $\beta$. The nonparanormal distributions in Sections~\ref{ssec:nonp}--\ref{ssec:x3} are a subset of this class, where $\rho$ is a multivariate Gaussian density (i.e., $\rho$ is in base class $\beta = 1$). For $\beta > 1$, the class considered here is a \emph{generalization} of the nonparanormal family of distributions.

\begin{figure}[!ht]
    \centering
 \begin{subfigure}{.3\textwidth}
    \centering
    \begin{tikzpicture}[transform shape]
      \tikzstyle{every node}=[draw,shape=circle,thick,scale = 0.75];
        \node[fill=blue!20] at (360:0cm) (1) {\scriptsize $X_{1}$};
        \foreach \a in {2,...,5}{
          \node[fill=blue!20] at (\a*360/4-45:1.4cm) (\a) {\scriptsize $X_{\a}$};
          \draw[-, thick, >=stealth'] (1)--(\a);
        }
    \end{tikzpicture}
 \end{subfigure}
    \hspace{0.8cm}
 \begin{subfigure}{.3\textwidth}
    \centering
    \begin{tikzpicture}
      \begin{scope}[transform shape, scale = 1 ]
      \matrix [draw ,column sep=.15cm, row sep=.15cm, ampersand replacement=\& ]
        {
      \node[entriesMatrix] { };\& \node[entriesMatrixZeroDashed]{ };\& \node[entriesMatrixZeroDashed]{ }; \& \node[entriesMatrixZeroDashed] { }; \& \node[entriesMatrixZeroDashed] { }; \\
      \node[entriesMatrix] { };\& \node[entriesMatrix]{ };\& \node[entriesMatrixZeroDashed]{ }; \& \node[entriesMatrixZeroDashed] { }; \& \node[entriesMatrixZeroDashed] { }; \\
      \node[entriesMatrix] { };\& \node[entriesMatrixZeroDashed]{ };\& \node[entriesMatrix]{ }; \& \node[entriesMatrixZeroDashed] { }; \& \node[entriesMatrixZeroDashed] { }; \\
      \node[entriesMatrix] { };\& \node[entriesMatrixZeroDashed]{ };\& \node[entriesMatrixZeroDashed]{ }; \& \node[entriesMatrix] { }; \& \node[entriesMatrixZeroDashed] { }; \\
      \node[entriesMatrix] { };  \& \node[entriesMatrixZeroDashed]{ };     \& \node[entriesMatrixZeroDashed]{ }; \& \node[entriesMatrixZeroDashed] { }; \& \node[entriesMatrix] { }; \\
        };
      \end{scope} 
    \end{tikzpicture}
 \end{subfigure}
 \caption{Markov structure and sparsity pattern $\mathcal{I}_{S}$ of the transport map $S$ for the $d = 5$ dimensional pullback density $S^{\sharp}\eta$~\label{fig:banana_graph}.}
\end{figure}
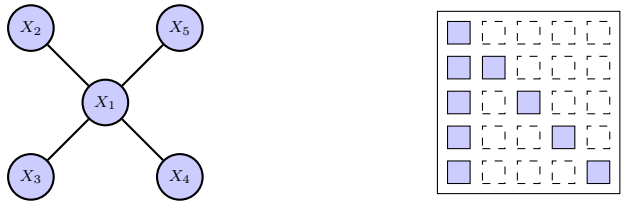

As an example, we consider a \rev{distribution} $\bm{\nu}_{\rho}$ in base class $\beta = 2$. Its density $\rho$ is given by the pullback of a standard Gaussian density $\eta$ through a sparse transport map $S$ of the form
\begin{equation}
S^{1}(x_{1}) = ax_{1}, \;\;\;\;\; S^{k}(x_{1},x_{k}) = (x_{1}^2 + b) + ax_{k} \;\; \text{ for } \;\; k = 2,\dots,d,
\end{equation}
where we set the parameters $a = 1$ and $b = 1$ to adjust the moments of the distribution $S^{\sharp}\bm{\nu}_{\eta}$. The transformed random variable $\bm{X} = S^{-1}(\bm{Y})$ for $\bm{Y} \sim \eta$ %
has the density $S^{\sharp}\eta$ and its Markov structure is displayed in Figure~\ref{fig:banana_graph} for a $d=5$ dimensional problem. The star graph associated with the conditional independence structure of $\rho$ is a commonly used graph benchmark for structure learning algorithms~\citep{jalali2011learning}. %

After defining the base density $\rho$, we apply the nonlinear inverse CDF transformation in~\eqref{eq:CDF_transformation} to each component of $\bm{X}$ in order to define the random variable $\bm{Z} = D^{-1}(\bm{X})$. Our target density is that of $\bm{Z}$: the pullback of a standard Gaussian density $\eta$ through the composition of $S$ and the diagonal map $D$, which we denote as $\pi = (S \circ D)^{\sharp}\eta$. To sample from $\pi$, we generate i.i.d.\thinspace samples $\mathbf{y}^{l}$ from the standard Gaussian reference density $\eta$ and apply the composition of the inverse maps $D^{-1} \circ S^{-1}\colon \mathbb{R}^{d} \to \mathbb{R}^{d}$ to each sample, thus generating i.i.d.\thinspace samples $\mathbf{z}^{l} = D^{-1} \circ S^{-1}(\mathbf{y}^{l})$.

\begin{figure}[!tbh]
  \centering
  \begin{subfigure}{.32\textwidth}
    \centering
    \includegraphics[width=\textwidth]{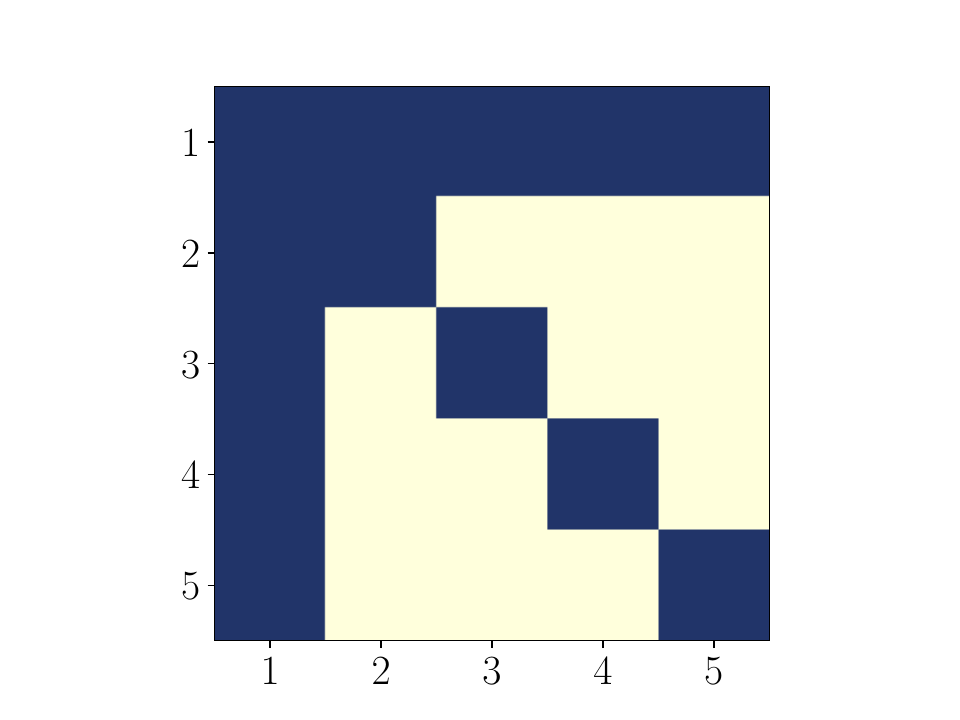}
    \caption{\label{fig:banresults_truegraph}}
  \end{subfigure}
  \begin{subfigure}{.32\textwidth}
    \centering
    \includegraphics[width=\textwidth]{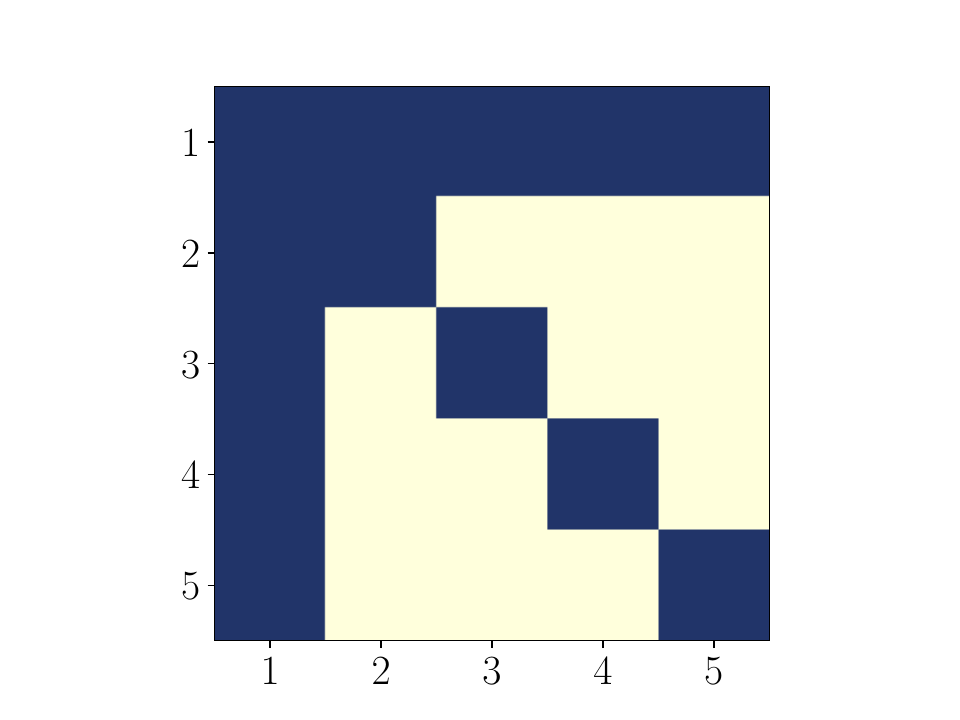}
    \caption{\label{fig:banresults_beta2}}
  \end{subfigure}
  \begin{subfigure}{.32\textwidth}
    \centering
    \includegraphics[width=\textwidth]{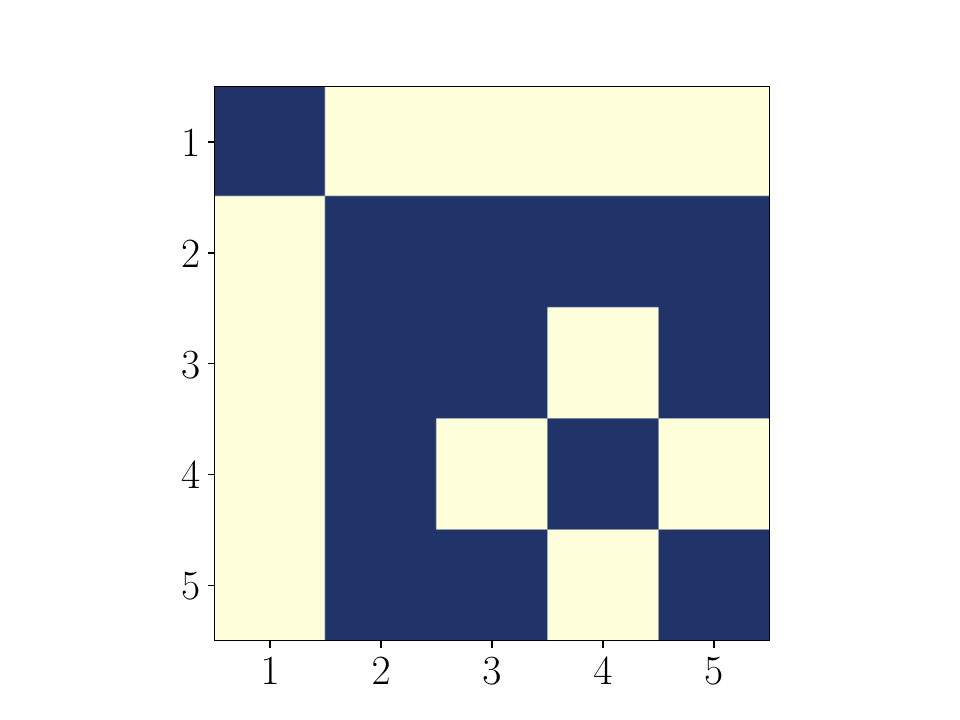}
    \caption{\label{fig:banresults_beta1}}
  \end{subfigure}
  \caption{(a) The true graph structure. (b) The recovered graph with $\beta = 2$ is correct. (c) Recovered graph with $\beta = 1$ is incorrect. \label{fig:banresults}}
\end{figure}

To learn the graph structure of $\pi$, we run the \textsc{sing} algorithm with $\beta = 2$ using $n = 10^{4}$ samples. The true graph and the recovered graph are displayed in Figures~\ref{fig:banresults_truegraph} and~\ref{fig:banresults_beta2}, respectively. The graph structure is learned correctly. On the other hand, Figure~\ref{fig:banresults_beta1} shows that using $\beta = 1$ does not recover the true graph of $\pi$.

\begin{figure}[tbh]
  \centering
  \begin{subfigure}{.45\textwidth}
    \centering
    \includegraphics[width=\textwidth]{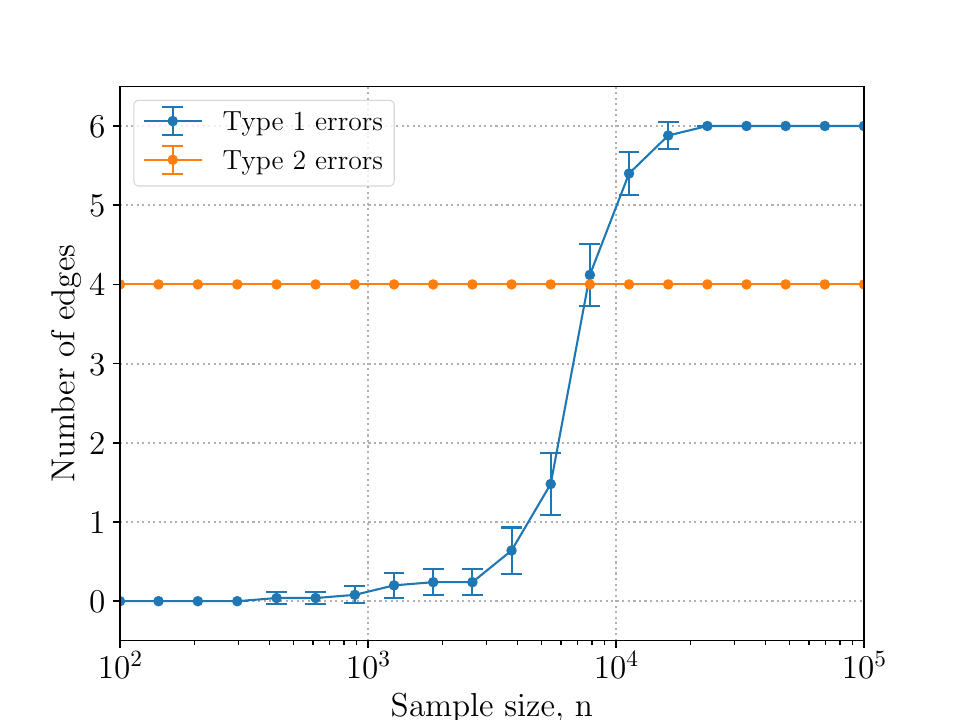}
    \caption{\label{fig:banresults_error_beta1}}
  \end{subfigure}
  \hspace{0.5cm} 
  \begin{subfigure}{.45\textwidth}
    \centering
    \includegraphics[width=\textwidth]{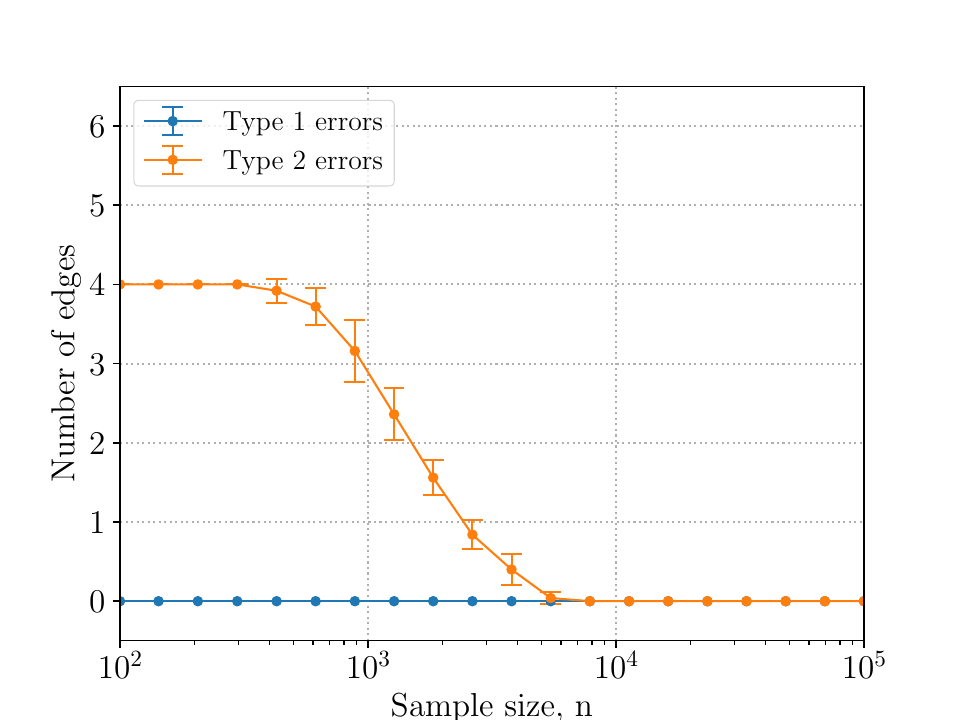}
    \caption{\label{fig:banresults_error_beta2}}
  \end{subfigure}
  \caption{Type 1 and 2 errors for recovering the graph of a nonlinear marginal transformation of a \textit{non-Gaussian} base distribution,  versus sample size $n$, using (a) $\beta = 1$, and (b) $\beta = 2$. \label{fig:banresults_samplesize}}
\end{figure}

To demonstrate the sample sizes needed to learn the true graph in this example, we run the \textsc{sing} algorithm for sample sizes $n \in [10^{2},10^{5}]$. Figures~\ref{fig:banresults_error_beta1} and~\ref{fig:banresults_error_beta2} display the average type 1 and type 2 errors in the estimated graphs for $\beta=1$ and $\beta = 2$, respectively, at each sample size $n$. %
For $n \geq 10^4$ samples roughly, the number of type 1 and type 2 errors using $\beta = 2$ remains close to zero and represents successful graph recovery. %
In contrast, Figure~\ref{fig:banresults_error_beta1} shows that using $\beta = 1$, there is no sample size within the tested range where one can recover the exact graph.
The conclusion here is analogous to the nonparanormal case of subsection~\ref{ssec:nonp}: the \textsc{sing} algorithm can return the correct Markov structure when %
the $\beta$ parameter matches the polynomial degree %
necessary to represent the base distribution and not the target.

\subsection{Lorenz-96 dynamical system}\label{ssec:l96}
We now apply \textsc{sing} to a physics-based dataset generated by a chaotic dynamical system. In particular, we consider the Lorenz-96 model, which is commonly used to represent features of the atmosphere (e.g., temperature or vorticity) on a mid-latitude circle of the Earth~\citep{lorenz1996predictability}. %
The state at $d$ discrete locations represents the discretization of a spatially \emph{periodic} domain, described by the state vector 
$\bm{Z}(t) = (Z_1(t),\dots,Z_d(t)) \in \mathbb{R}^{d}$ at time $t$. 
The evolution of this state in time is defined by the set of coupled nonlinear ODEs
\begin{align} \label{eq:lorenz96_ODE}
    \frac{dZ_j}{dt} = \left(Z_{j+1} + Z_{j-2}\right)Z_{j-1} - Z_j + F, \quad j=1,\dots,d,
\end{align}
where $Z_{-1} \equiv Z_{d-1}$, $Z_{0} \equiv Z_{d}$, $Z_{1} \equiv Z_{d+1}$. 
In our experiments we use $d=15$ and $F=8$, which leads to chaotic dynamics~\citep{reich2015probabilistic}. 

\begin{figure}
  \centering
  \begin{subfigure}{.49\textwidth}
    \includegraphics[width=\linewidth]{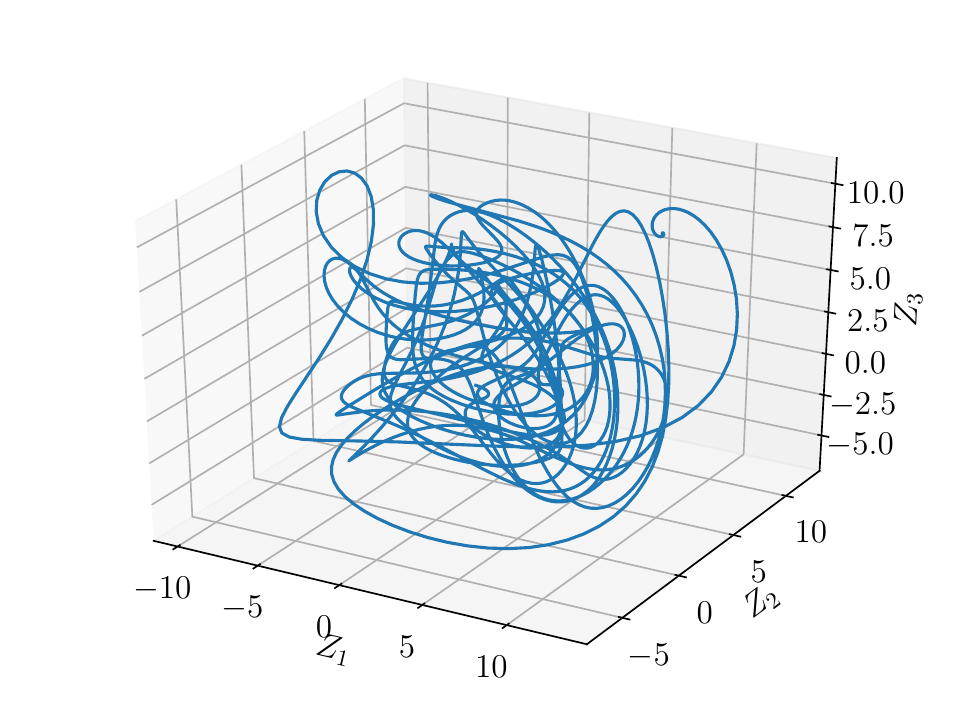}
    \caption{\label{fig:l96-3var}}
  \end{subfigure}
  \begin{subfigure}{.49\textwidth}
    \includegraphics[width=\linewidth]{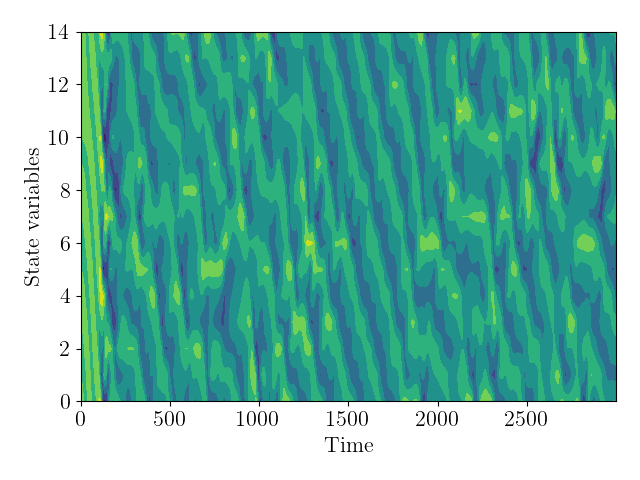}
    \caption{\label{fig:l96-15var}}
  \end{subfigure}
  \caption{15-dimensional Lorenz-96 dataset: (a)~A sample trajectory of three neighboring state variables $(Z_1,Z_2,Z_3)$. (b)~Magnitude of all 15 state variables over time. \label{fig:l96-dyn}}
\end{figure}

Our goal is to characterize the conditional independence properties of the invariant distribution of the state; thus we collect data from long-time trajectories of the system.
\rev{Figure~\ref{fig:l96-3var} shows a sample trajectory of three consecutive variables, and Figure~\ref{fig:l96-15var} displays a heat map of all 15 variables. While there appears to be some dependence among neighboring nodes, the dependence structure is certainly not revealed by data visualization alone. Indeed, this example does not have a known graph representing its ``true'' conditional dependence, which is also the case with other data sets representative of real atmospheric dynamics. 
Nonetheless, approximating the Markov properties of the state of such systems is important to localization schemes for data assimilation (see \citet{spantini2019coupling}) and to learning the dynamics directly (see \citet{ott2002atmospheric}).}%

To generate a trajectory, we sample a random initial condition $\bm{Z}(0) \sim \mathcal{N}(\bm{0},I_{d})$ and use a 4th order Runge-Kutta method with a time step of $\Delta t = 0.01$ for $t \in [0,1600]$ to approximate the state vector $\bm{Z}(k \Delta t)$ of the ODE in~\eqref{eq:lorenz96_ODE} as $\bm{Z}^{k}$, for $k \in \mathbb{N}_{0}$. To reduce correlation between the samples, we sub-sample the trajectory by collecting samples only for $k = 40m$,  $m \in \mathbb{N}_{0}$. Furthermore, to reduce the effect of the initial condition, we discard the first $1000$ samples of the trajectory.

Given this dataset, we then learn the Markov structure of the invariant distribution of the Lorenz-96 system. We run the \textsc{sing} algorithm with $\beta = 2$ using $n = 3000$ samples from this dataset. In this problem, the dynamics at each time step introduce local interactions between each state variable $Z_{j}$ and its neighboring variables on the discretized periodic domain, as seen in the structure of the ODE in~\eqref{eq:lorenz96_ODE}. As a result, the repeated application of the dynamics results in full dependence amongst the variables. However, the invariant distribution of the system can be well approximated by a Markov random field where each variable conditioned on its closest few neighbors in the physical grid is independent of the others. To account for the weak conditional independence between distant states in the grid, we use a threshold for each entry of the conditional independence score given by $\tau_{ij} = \tau_0 + f(n) \widetilde{\upsilon}_{ij} / \sqrt{n} $, where $\tau_0 \geq 0$ is a constant offset for all $(i,j)$. This can be seen as a generalization of the threshold proposed in subsection~\ref{ssec:tol} and applied in the previous numerical examples, where we simply used $\tau_0 = 0$. Here we set $\tau_0 = 0.1$.  %

\begin{figure}[!ht]
  \centering
  \begin{subfigure}{.35\textwidth}
    \includegraphics[width=\linewidth]{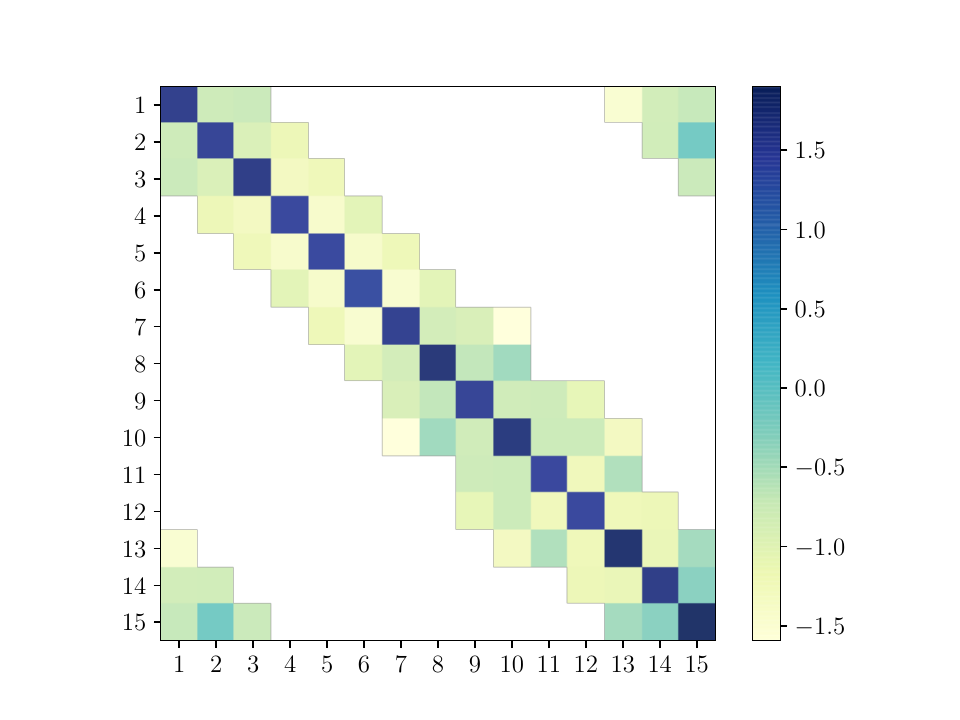}
    \caption{\label{fig:l96-a}}
  \end{subfigure}
  \begin{subfigure}{.35\textwidth}
    \includegraphics[width=\linewidth]{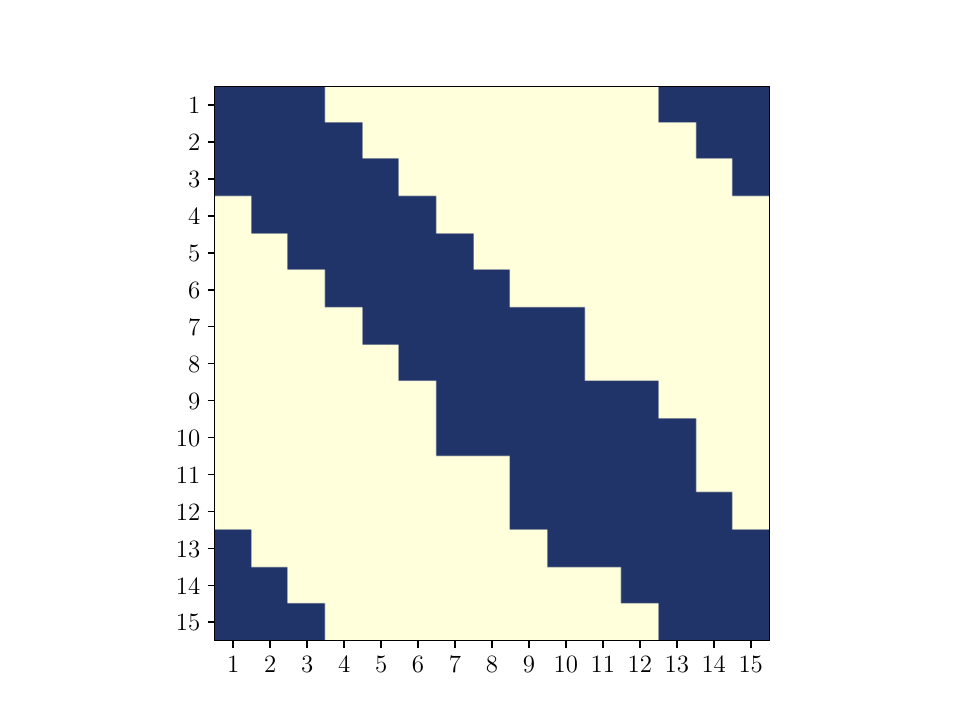}
    \caption{\label{fig:l96-adj-beta2}}
  \end{subfigure}
    \begin{subfigure}{.35\textwidth}
    \includegraphics[width=\linewidth]{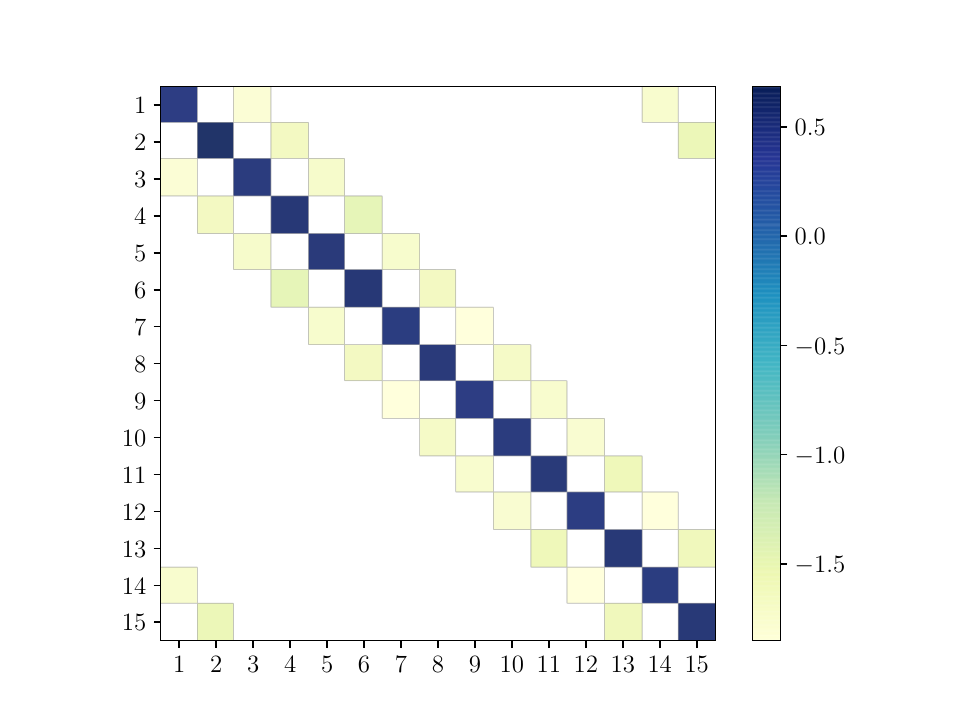}
    \caption{\label{fig:l96-a-beta1}}
  \end{subfigure}
  \begin{subfigure}{.35\textwidth}
    \includegraphics[width=\linewidth]{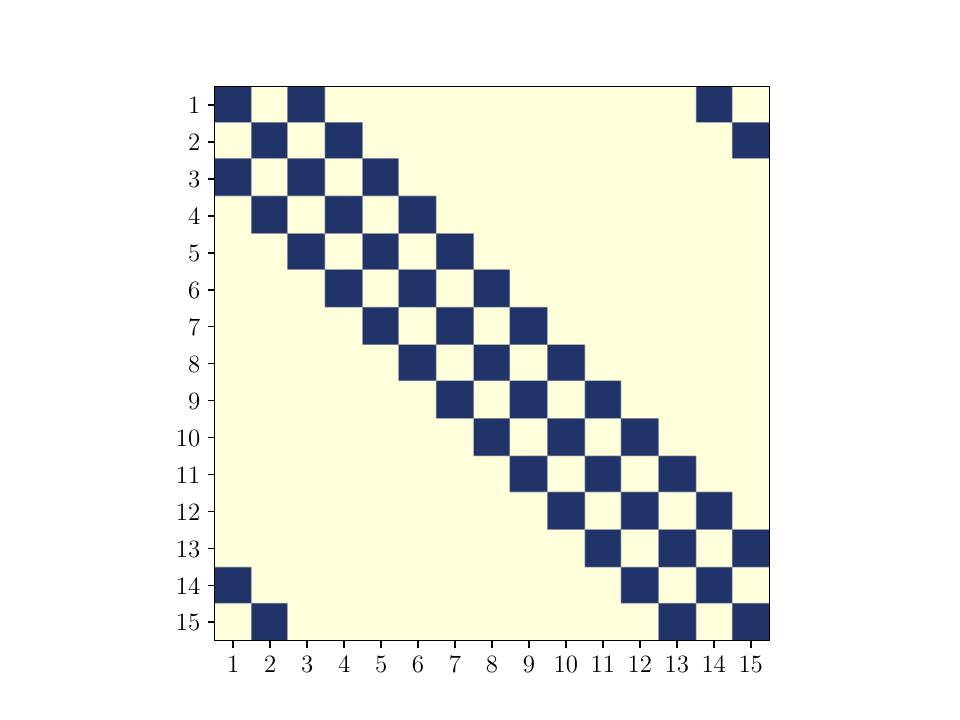}
    \caption{\label{fig:l96-adj-beta1}}
  \end{subfigure}
  \caption{15-dimensional Lorenz-96 dataset: Top row: $\beta = 2$; Bottom row: $\beta=1$. Left column: Logarithm of nonzero entries in $\overline{\Omega}$, at the final iteration of \textsc{sing}; Right column: Adjacency matrix of the graph learned using \textsc{sing}. (d)~In particular, the adjacency matrix obtained with $\beta = 1$ incorrectly suggests that each variable $Z_{j}$ is conditionally independent of its immediate neighbors, $Z_{j-1}$ and $Z_{j+1}$. \label{fig:l96}}
\end{figure}

The thresholded conditional independence score $\overline{\Omega}$ and the corresponding adjacency matrix of the graph found with \textsc{sing} are shown in Figures~\ref{fig:l96-a}-\ref{fig:l96-adj-beta1}, for both $\beta = 2$ and $\beta=1$. To emphasize the decay of the entries in $\overline{\Omega}$ away from the diagonal, we plot the entrywise logarithm of $\overline{\Omega}$. Figure~\ref{fig:l96-a} demonstrates the banded dependence of each variable on neighboring variables separated by at most $3$ nodes in either direction, as well as the periodic structure of the graph. On the other hand, the \textsc{sing} algorithm with $\beta = 1$ (i.e., a Gaussian approximation to the target density $\pi$) entirely misses the conditional dependence of each variable on its immediate neighbors, as seen in Figure~\ref{fig:l96-adj-beta1}.

\subsection{Cell signaling data}
\label{ssec:cells}
\rev{Now we apply \textsc{sing} to learn conditional dependence relationships among the components of a cellular signaling network, based on single-cell data obtained via flow cytometry \citep{sachs2005causal}. The data comprise simultaneous measurements of $d = 11$ phosphorylated proteins and phospholipids in $n = 7446$ individual primary human immune system cells. This dataset was used in~\citet{sachs2005causal} to learn a Bayesian network (i.e., a directed acyclic graph) and in~\citet{friedman2008sparse} to fit a Gaussian graphical model using the \textsc{glasso} algorithm. Here we investigate the effect of accounting for non-Gaussianity, via a nonlinear transport map, on the learned dependencies.}

\rev{
As the measurements are all non-negative, we first apply a marginal log transformation to each variable. This pre-processing step transforms the support of the samples from $\mathbb{R}_{+}^d$ to $\mathbb{R}^d$. Hence, it makes the transport map pulling back the standard Gaussian reference to the target distribution better behaved and easier to approximate. Note that this diagonal transformation does not change the graph structure; see Section~\ref{ssec:diag}.}

\begin{figure}[!ht]
  \centering
  \begin{subfigure}{.35\textwidth}
    \includegraphics[width=\linewidth]{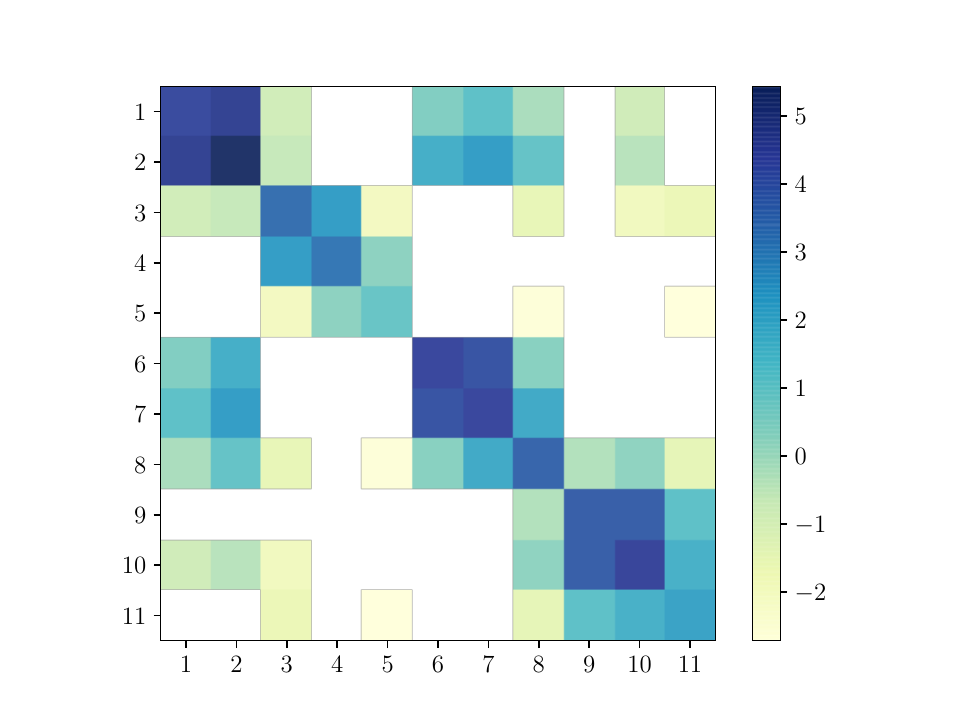}
    \caption{\label{fig:sachs-gp}}
  \end{subfigure}
  \begin{subfigure}{.35\textwidth}
    \includegraphics[width=\linewidth]{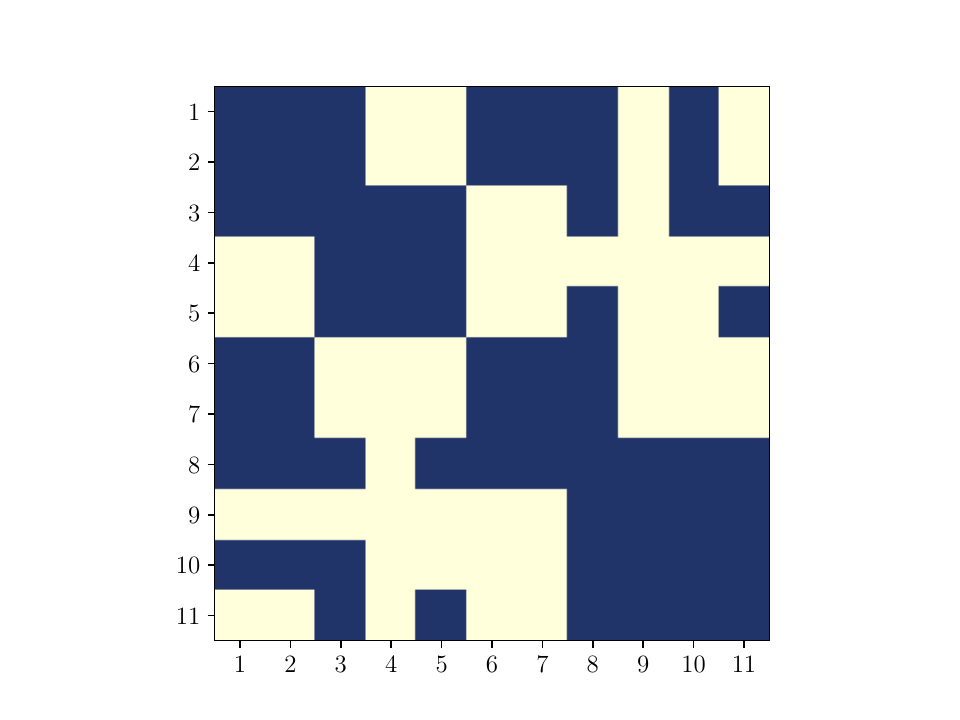}
    \caption{\label{fig:sachs-beta2}}
  \end{subfigure}
  \begin{subfigure}{.35\textwidth}
    \includegraphics[width=\linewidth]{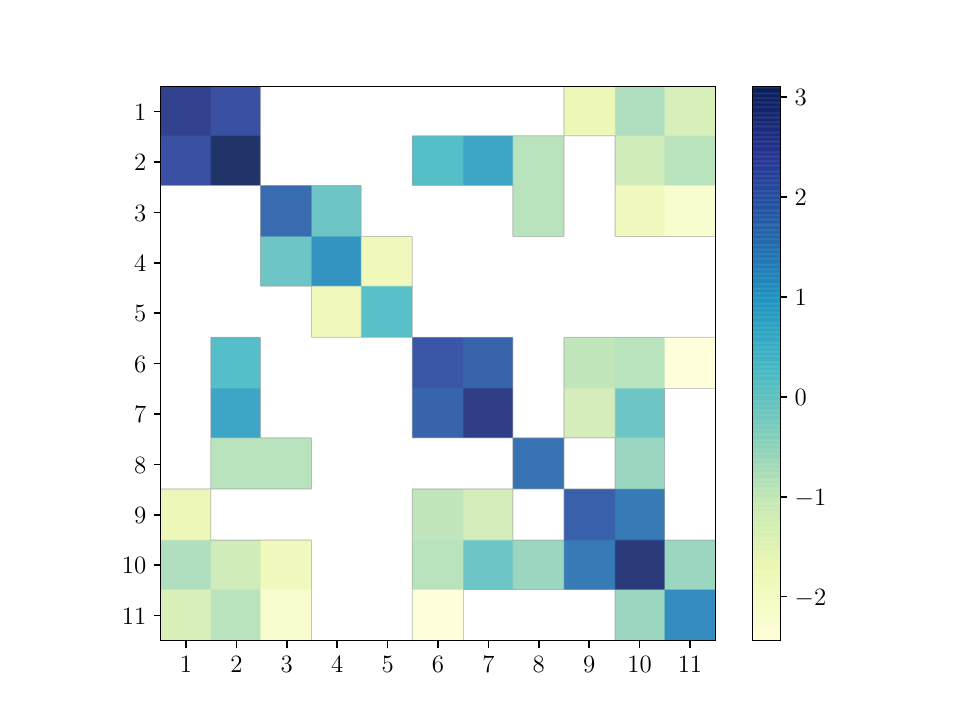}
    \caption{\label{fig:sachs-gp1}}
  \end{subfigure}
  \begin{subfigure}{.35\textwidth}
    \includegraphics[width=\linewidth]{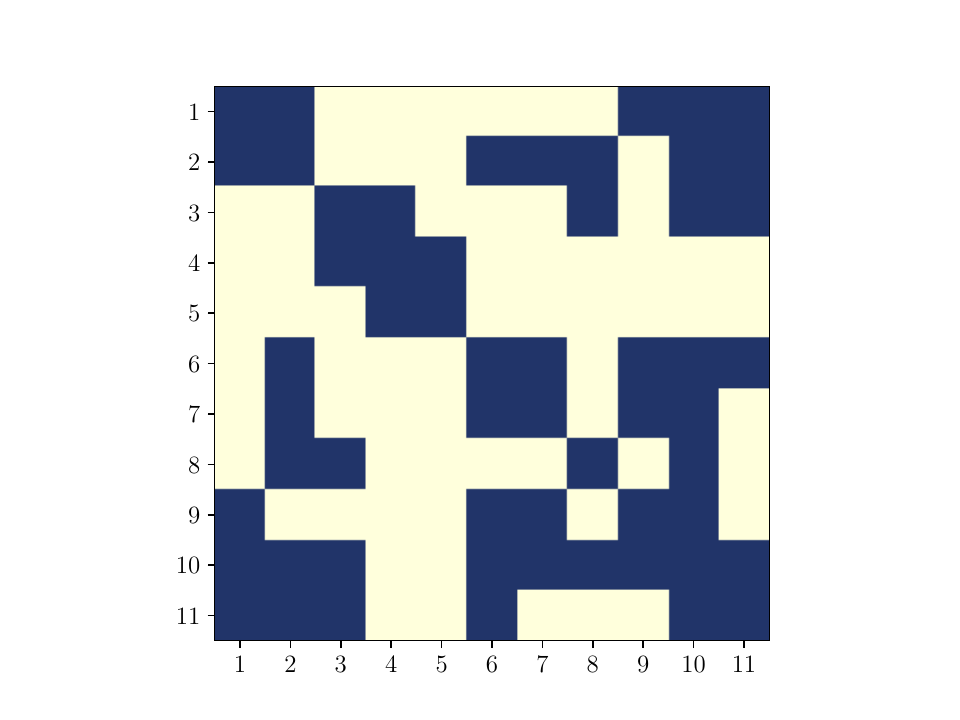}
    \caption{\label{fig:sachs-beta1}}
  \end{subfigure}
  \caption{\rev{11-dimensional cell signaling dataset. Top row: $\beta = 2$; Bottom row: $\beta=1$. Left column: Logarithm of nonzero entries in $\overline{\Omega}$, at the final iteration of \textsc{sing}; Right column: Adjacency matrix of the graph learned using \textsc{sing}.} \label{fig:sachs}}
\end{figure}

\rev{Figures~\ref{fig:sachs-gp} and \ref{fig:sachs-beta2} display the log of the estimated conditional independence score $\overline{\Omega}$ and the adjacency matrix, respectively, from running \textsc{sing} with $\beta = 2$ and a variance threshold based on $f(n) = \delta\log(n)/\sqrt{n}$. Figures~\ref{fig:sachs-gp1} and~\ref{fig:sachs-beta1} display the corresponding conditional independence score and adjacency matrix with $\beta = 1$, i.e., a linear transport map. For both map parameterizations, we select $\delta \in \{1.0,1.25,1.5,1.75,2.0\}$ by performing a 10-fold cross-validation procedure to identify the value which yields, after thresholding,  
an approximate density $S_{\widehat{\boldsymbol{\alpha}}}^\sharp\eta$ that maximizes the average log-likelihood on the test sets.}

\rev{In Figure~\ref{fig:sachs-gp}, and in both Figures~\ref{fig:sachs-beta2} and~\ref{fig:sachs-beta1}, we observe %
strong conditional dependence (i.e., larger values for the estimated score $\overline{\Omega}$), and hence the presence of edges, between variable pairs $(Z_1,Z_2)$, $(Z_3,Z_4)$, $(Z_6,Z_7)$ and $(Z_9,Z_{10})$. These pairs are also connected by edges in the directed graph of Figure~\ref{fig:sachs_true}, whose relationships were verified experimentally by~\cite{sachs2005causal} and thus can be considered as a \emph{true} reference. For variable pairs with weaker dependence, however, the \textsc{sing} algorithm finds different edge sets when accounting for non-Gaussian structure (with $\beta = 2$) than in the Gaussian case ($\beta=1$). For example, the non-Gaussian approach identifies edges between $Z_8$ and the set $(Z_6,Z_7)$, which are also %
found %
in the directed graph by~\cite{sachs2005causal}. In contrast, the Gaussian graphical model fails to identify edges between $Z_8$ and $(Z_6,Z_7,Z_9)$. These edges are present when accounting for non-Gaussianity with $\beta = 2$, and they follow from
the true directed graph after moralization.}

\begin{figure}[!ht]
    \centering
    \begin{tikzpicture}[transform shape]
        \foreach \a in {1,...,11}{
          \node[draw,circle,fill=blue!20] at (\a*360/11-360/11+90:1.2cm) (\a) {};
          \node at (\a*360/11-360/11+90:1.6cm) {\scriptsize $Z_{\a}$};
        }
        \draw[->, >=stealth'] (1)--(2);
        \draw[->, >=stealth'] (2)--(6);
        \draw[->, >=stealth'] (3)--(4);
        \draw[->, >=stealth'] (3)--(5);
        \draw[->, >=stealth'] (3)--(9);
        \draw[->, >=stealth'] (4)--(9);
        \draw[->, >=stealth'] (5)--(4);
        \draw[->, >=stealth'] (5)--(7);
        \draw[->, >=stealth'] (6)--(7);
        \draw[->, >=stealth'] (8)--(7);
        \draw[->, >=stealth'] (8)--(6);
        \draw[->, >=stealth'] (8)--(10);
        \draw[->, >=stealth'] (8)--(11);
        \draw[->, >=stealth'] (8)--(1);
        \draw[->, >=stealth'] (8)--(2);
        \draw[->, >=stealth'] (9)--(10);
        \draw[->, >=stealth'] (9)--(11);
        \draw[->, >=stealth'] (9)--(1);
        \draw[->, >=stealth'] (9)--(2);
    \end{tikzpicture}
    \caption{\rev{Directed acyclic graph identified in~\cite{sachs2005causal} from cell-signaling data. To simplify the presentation, we assign a numeric label to each protein as follows: 1=Raf, 2=Mek, 3=Plc$\gamma$, 4=PIP2, 5=PIP3, 6=Erk, 7=Akt, 8=PKA, 9=PKC, 10=P38, and 11=Jnk.}}
    \label{fig:sachs_true}
\end{figure}

\section{Discussion} \label{sec:con}
This paper develops a framework for learning the Markov structure of continuous non-Gaussian probability distributions from data. The framework is built on two key elements. First, we introduce a computationally tractable score for conditional independence, based on averaged Hessian information from the target log-density. %
Though this score is immediately useful in its own right, we also show that it bounds the conditional mutual information under appropriate assumptions. Second, we use a transport-based density estimation method, based on parametric approximations of the triangular Knothe--Rosenblatt rearrangement, that explicitly and iteratively exploits sparsity. In particular, the algorithm uses sparsity bounds for triangular maps, which follow directly from the Markov structure; variance-based thresholding for the score estimator; %
and variable ordering schemes designed to preserve sparsity. Putting these elements together yields the \textsc{sing} algorithm for structure learning. %
Our analysis shows consistent graph recovery for a single iteration of the algorithm in an asymptotic regime, while numerical results demonstrate the benefits of an iterative algorithm with finite samples. %

We demonstrate in multiple examples that this framework can recover the correct conditional independence structure in cases where a Gaussian approximation would yield entirely incorrect results. At the same time, our results show that the map parameterization need not be chosen rich enough to exactly capture the target density. The latter property points to an ``information gap,'' which reflects the idea that learning the graph structure should require capturing less information than the target distribution itself. Indeed, any undirected graph encodes the Markov properties of an infinite number of distributions. %

Put more specifically: there are infinitely many different distributions that will have the same conditional independence score matrix, and even more that will have score matrices with the same sparsity pattern. Moreover, as shown in the analytical example of Section~\ref{ssec:x3}, entries of the score matrix resulting from a biased approximation to the map---i.e., entries that would be zero in expectation if the density approximation were unbiased---can be an order of magnitude smaller than the non-spurious entries. It would be very useful to understand why this is the case, and under what conditions one can expect a favorable interaction between finite sample sizes, threshold estimators, and small spurious entries in the score matrix. These insights could guide the choice of transport map parameterization for any given data set.

Indeed, the parameterization of the triangular transport is a useful degree of freedom of our
overall framework. Polynomials (or the closely related Hermite functions) are convenient as they
include the Gaussian setting as a special case, but future work can certainly explore other options,
e.g., the nonlinear separable representations in \citet{spantini2019coupling}, or single layers of
autoregressive flows. Further analysis of the information gap described above might make it possible
to develop transport map estimators with the \emph{minimal} representation that is needed to learn
the Markov structure of the target distribution. Relatedly, it would be valuable to develop
information theoretic (\emph{representation-independent}) lower bounds on the number of samples
needed to identify the graph in the non-Gaussian setting. To our knowledge, these bounds only exist
for Gaussian and discrete Markov random fields~\citep{wang2010information,
santhanam2012information}. %

We outline a few additional avenues for future work as follows:

\noindent \textbf{Different data sources.} In practical settings, it is of interest to learn the graph from heterogeneous data that may not be independent and identically distributed. Examples include data collected from several related populations that have common structure, and data collected over time. There has been some work to address these problems in the Gaussian setting: \citet{guo2011joint, danaher2014joint} propose to combine optimization problems for separate
precision matrices with shared $\ell_{1}$-penalties to identify common sparsity, while~\citet{zhou2010time} exploits smooth changes in the %
precision matrix to estimate the evolution of the graph over time.

\noindent \textbf{Latent structure.} We also envision extending the \textsc{sing} algorithm to learn the
structure of graphical models with latent variables and partial observations. For Gaussian
distributions, \citet{chandrasekaran2012} proposes a penalized maximum likelihood approach to
identify a precision matrix with sparse and low-rank structure. A similar approach may explore other
properties of the conditional independence score matrix $\Omega$, besides sparsity, to reveal hidden variables and multiscale structure in the target density. %

\noindent \textbf{High-dimensional models.} 
To learn the graph of a $d$-dimensional distribution, the \textsc{sing} algorithm computes transport
maps with $d$ components and stores $\mathcal{O}(d^2)$ entries for the estimated conditional
independence score in memory. For high-dimensional datasets, it may not be feasible to jointly
estimate the associated graph.  Instead, neighborhood selection methods identify local dependencies by
independently estimating the neighborhood of each node in $\mathcal{G}$, i.e.,
$\text{Nb}(k,\mathcal{G})$ for $k=1,\dots,d$ such that $\pi(z_{k}|\bm{z}_{-k}) =
\pi(z_{k}|\bm{z}_{\text{Nb}(k,\mathcal{G})})$. Methods for finding the neighborhood include greedy
selection strategies~\citep{bresler2015efficiently} and penalized maximum likelihood
estimators~\citep{meinshausen2006high}.  Future work will consider a neighborhood selection version
of the \textsc{sing} algorithm that avoids estimating the global transport map for the joint
density; this would reduce the run time and memory required to learn the graph.

\acks{We would like to acknowledge support for this project
from the National Science Foundation (NSF grant IIS-9988642), the
Multidisciplinary University Research Program of the Department
of Defense (MURI N00014-00-1-0637), and the INRIA associate team
Unquestionable. We also thank Daniele Bigoni and Teo Price-Broncucia for programming support. RB and YM acknowledge support from the Department
of Energy, Office of Advanced Scientific Computing Research, AEOLUS (Advances in Experimental design, Optimal control, and Learning for Uncertain complex Systems) center. RB acknowledges support from an NSERC PGSD fellowship. RM %
acknowledges Dr.~Dusa McDuff and the Johnson\&Johnson Foundation for support---financial and otherwise---through the Women in STEM2D Scholars Program. }

\appendix

\section{Proof of Theorem~\ref{th:OmegaBoundCMI}} \label{app:CMI_bound} \begin{proof}%
 The proof contains two steps: we first show the result when $d=2$, then extend it to $d>2$.
 
 Assume $d=2$. The conditional mutual information $I(Z_i ; Z_j|\bm{Z}_{-ij}) $ becomes the mutual information $ I(Z_i ; Z_j)$, given by
 \begin{align*}
  I(Z_i ; Z_j) 
  &= \int \pi(z_i,z_j) \log\left(\frac{\pi(z_i,z_j)}{\pi(z_i) \pi(z_j)}  \right) \textrm{d}z_i\textrm{d}z_j \\
  &= \int h(\bm{z}) \log\left(\frac{h(\bm{z})}{\int h(\bm{z}')\pi(z_i') \pi(z_j')\textrm{d}z_i'\textrm{d}z_j'}\right) \pi(z_i) \pi(z_j)\textrm{d}z_i\textrm{d}z_j ,
 \end{align*}
 where $h(\bm{z}) = h(z_i,z_j) = \pi(z_i,z_j) / (\pi(z_i) \pi(z_j))$.
 Next we apply the logarithmic Sobolev inequality to bound $I(Z_i ; Z_j)$. 
 To do that, we need to show that the product density $(z_i,z_j)\mapsto\pi(z_i)\pi(z_j)$ satisfies the logarithmic Sobolev inequality. By Theorem 4.4. in \cite{guionnet2003lectures}, $(z_i,z_j)\mapsto\pi(z_i) \pi(z_j)$ has a logarithmic Sobolev constant $C(\pi(z_i)\pi(z_j))$ bounded by $\max \{C(\pi(z_i)) ; C(\pi(z_j))\}$. By Assumption \eqref{eq:AssumptionOmegaBoundCMI} the joint density $\pi(z_i,z_j)$ has a logarithmic Sobolev inequality bounded by $C_0$ and therefore the marginals $\pi(z_i)$ and $\pi(z_j)$ satisfy $C(\pi(z_i))\leq C_0$ and $C(\pi(z_j))\leq C_0$ (this can be easily shown by restricting \eqref{eq:AssumptionOmegaBoundCMI} to univariate functions $h:(z_i,z_j)\mapsto h(z_i)$ or $h:(z_i,z_j)\mapsto h(z_j)$). We deduce that the product of marginals satisfies the logarithmic Sobolev inequality with constant $C(\pi(z_i) \pi(z_j)) \leq C_0$. We can write
 \begin{align}
  I(Z_i ; Z_j) 
 &\leq \frac{C_0}{2} \int 
  \left\| 
  \begin{matrix}
   \partial_i \log h(z_i,z_j) \\ \partial_j \log h(z_i,z_j) 
  \end{matrix}
  \right\|_2^2 h(z_i,z_j) \pi(z_i) \pi(z_j)\textrm{d}z_i\textrm{d}z_j \nonumber\\
  &= \frac{C_0}{2} \int 
  \left( \int \big(\partial_i \log \pi(z_i,z_j) - \partial_i \log \pi(z_i)  \big)^2 \pi(z_j|z_i)\textrm{d}z_j \right) \pi(z_i)\textrm{d}z_i \nonumber\\
  & + \frac{C_0}{2} \int 
  \left( \int \big(\partial_j \log \pi(z_i,z_j) - \partial_j \log \pi(z_j)  \big)^2 \pi(z_i|z_j)\textrm{d}z_i \right) \pi(z_j)\textrm{d}z_j . \label{eq:tmp833}
 \end{align}
 
 Next we apply the Poincar\'e inequality to bound the two integrands in the above expression. 
 By assumption \eqref{eq:AssumptionOmegaBoundCMI}, the density $z_i\mapsto\pi(z_i|z_j)$ has a logarithmic Sobolev constant bounded by $C_0$ so that it satisfies the Poincar\'e inequality 
 \begin{equation}\label{eq:Poincare}
  \int \left(f(z_i) - \int f(z_i') \pi(z_i'|z_j)\mathrm{d}z_i'\right)^2 \pi(z_i|z_j)\mathrm{d}z_i \leq C_0 \int f'(z_i)^2 \pi(z_i|z_j) \mathrm{d}z_i ,
 \end{equation}
 for any continuously differentiable function $f:\mathbb{R}\rightarrow\mathbb{R}$. The Poincar\'e inequality \eqref{eq:Poincare} is classically obtained from a logarithmic Sobolev inequality \eqref{eq:logSob} by letting $h=1+\varepsilon f$ and by taking a Taylor expansion as $\varepsilon\rightarrow0$. Notice that
 $$
  \int \big(\partial_j \log \pi(z_i',z_j)\big) \pi(z_i'|z_j)\mathrm{d}z_i' 
  = \int \frac{\partial_j \pi(z_i',z_j)}{\pi(z_i',z_j)} \frac{\pi(z_i',z_j)}{\pi(z_j)}\mathrm{d}z_i' 
  = \frac{\partial_j \int \pi(z_i',z_j) \mathrm{d}z_i' }{\pi(z_j)}
  = \partial_j \log \pi(z_j),
 $$
 so that the Poincar\'e inequality \eqref{eq:Poincare} with $f(z_i)=\partial_j \log \pi(z_i,z_j)$ yields
 \begin{align*}
  \int \big(\partial_j \log \pi(z_i,z_j) - \partial_j \log \pi(z_j)  \big)^2 \pi(z_i|z_j)\textrm{d}z_i 
  &\leq C_0 \int \big(\partial_i\partial_j \log \pi(z_i,z_j)\big)^2 \pi(z_i|z_j) \mathrm{d}z_i .
 \end{align*}
 In the same way, by permuting $i$ and $j$, we obtain
 $$
  \int \big(\partial_i \log \pi(z_i,z_j) - \partial_i \log \pi(z_i)  \big)^2 \pi(z_j|z_i)\textrm{d}z_j 
  \leq C_0 \int \big(\partial_i\partial_j \log \pi(z_i,z_j)\big)^2 \pi(z_j|z_i) \mathrm{d}z_j.
 $$
 Using the above two inequalities in \eqref{eq:tmp833} yields
 \begin{align*}
  I(Z_i ; Z_j) &\leq \frac{C_0}{2}\int \left(C_0\int \big(\partial_i\partial_j \log \pi(z_i,z_j)\big)^2 \pi(z_j|z_i) \mathrm{d}z_j\right)\pi(z_i)\mathrm{d}z_i \\
  &+ \frac{C_0}{2}\int \left(C_0\int \big(\partial_i\partial_j \log \pi(z_i,z_j)\big)^2 \pi(z_i|z_j) \mathrm{d}z_i\right)\pi(z_j)\mathrm{d}z_j \\
  &= C_0^2 \int \big(\partial_i\partial_j \log \pi(z_i,z_j)\big)^2 \pi(z_i,z_j) \mathrm{d}z_i\mathrm{d}z_j.
 \end{align*}
 This shows that $I(Z_i ; Z_j)\leq C_0^2 \Omega_{i,j}$ when $d=2$.
 
 Assume now that $d>2$. For any $\bm{z}_{-ij}\in\mathbb{R}^{d-2}$, replacing $\pi(z_i,z_j)$ by $\pi(z_i,z_j|\bm{z}_{-ij})$ in the previous analysis allows us to write
 \begin{align*}
  \int & \pi(z_i,z_j|\bm{z}_{-ij})  \log\left(\frac{\pi(z_i,z_j|\bm{z}_{-ij})}{\pi(z_i|\bm{z}_{-ij}) \pi(z_j|\bm{z}_{-ij})}  \right) \textrm{d}z_i\textrm{d}z_j \\
  &\leq C_0^2\int \big(\partial_i\partial_j \log \pi(z_i,z_j|\bm{z}_{-ij})\big)^2 \pi(z_i,z_j|\bm{z}_{-ij}) \mathrm{d}z_i\mathrm{d}z_j \\
  &= C_0^2\int \big(\partial_i\partial_j \log \pi(\bm{z})\big)^2 \pi(z_i,z_j|\bm{z}_{-ij}) \mathrm{d}z_i\mathrm{d}z_j,
 \end{align*}
 where the last equality is obtained by $\log \pi(z_i,z_j|\bm{z}_{-ij}) = \log \pi(\bm{z}) - \log \pi(\bm{z}_{-ij})$.
 Multiplying by the marginal $\pi(\bm{z}_{-ij})$ and integrating over $\bm{z}_{-ij}\in\mathbb{R}^{d-2}$ we obtain \eqref{eq:OmegaBoundCMI}, which concludes the proof.
\end{proof}

\section{Proofs of transport map results}\label{app:other_proofs} 
\begin{proof}{\hspace{-.1em}\textbf{of Proposition~\ref{prop:diag}}}
    Let $\bm{\nu}_{\rho}$ be a \rev{measure} on $\mathbb{R}^{d}$ that is Markov with respect to a graph $\mathcal{G}$ and has a strictly positive density $\rho$. %
    By the Hammersley-Clifford theorem, the density $\rho$ factorizes as
    \begin{equation} \label{eq:density_potentials}
        \rho(\bm{x}) = \frac{1}{\mathcal{Z}} \prod_{c \in \mathcal{C}} \varphi_c(\bm{x}_c),
    \end{equation}
    where $\varphi_{c}$ are nonnegative potential functions, $\mathcal{Z}$ is a normalizing constant and %
    $\mathcal{C}$ is the set of maximal cliques of $\mathcal{G}$~\citep{lauritzen1996graphical}. A clique is a fully connected subset of nodes and a maximal clique is a clique that is not a strict subset of another clique. %

    The pullback density of $\rho$ through a differentiable diagonal transport map $D$ is given by
    \begin{align}
        D^\sharp \rho(\bm{z}) &= \rho \circ D(\bm{z}) |\det \nabla D(\bm{z})| \nonumber \\
        &= \rho \circ D(\bm{z}) \prod_{i=1}^d \partial_{i} D^i(z_i) \nonumber \\
        &= \frac{1}{\mathcal{Z}} \prod_{c \in \mathcal{C}} \varphi_{c}(D^{c}(\bm{z}_{c})) \prod_{i=1}^d \partial_{i} D^i(z_i) \nonumber ,
    \end{align}
    where $D^{c}$ represents the subset of components of $D$ corresponding to the nodes in clique $c$. Collecting the derivatives of the map components for the nodes in each clique, we have
    \begin{align}
        D^\sharp \rho(\bm{z}) &= \frac{1}{\mathcal{Z}} \left( \varphi_{c_1} (D^{c_1}(\bm{z}_{c_1})) \prod_{i \in c_1} \partial_{i} D^i(z_i) \right) \left( \varphi_{c_2} (D^{c_2}(\bm{z}_{c_2})) \prod_{i \in c_2; i \notin c_1} \partial_{i} D^i(z_i) \right) \dots \nonumber\\
        & \quad\quad\quad \left( \varphi_{c_M} (D^{c_M}(\bm{z}_{c_M})) \prod_{i \in c_M;
        i \notin c_1,\dots,c_{M-1}} \partial_{i} D^i(z_i) \right) \nonumber \\
        &\equiv \frac{1}{\mathcal{Z}} \psi_{c_1}(\bm{z}_{c_1})\psi_{c_2}(\bm{z}_{c_2}) \cdot \dots \cdot \psi_{c_M}(\bm{z}_{c_M}), \label{eq:pullback_diagonal_potentials}
    \end{align}
    where $\psi_{c_j}(z_{c_j})$ defines new potential functions of the variables $\bm{Z}_{c_j}$ in the maximal clique $c_{j}$, and $M = |\mathcal{C}|$ represents the cardinality of $\mathcal{C}$. %
    From~\eqref{eq:pullback_diagonal_potentials}, the density of $D^\sharp \rho$ also factorizes according to $\mathcal{G}$. Thus, by Proposition 3.8 in~\citet{lauritzen1996graphical}, $D^\sharp \bm{\nu}_{\rho}$ is Markov with respect to $\mathcal{G}$. %
    \end{proof}
We note that the contrapositive of Proposition~\ref{prop:diag} immediately follows: If the minimal I-map of $S^\sharp \bm{\nu}_{\rho}$ is not
equivalent to that of $\bm{\nu}_{\rho}$, then the map $S$ is not diagonal.

\vspace{2em}
\begin{proof}{\hspace{-.1em}\textbf{of Proposition~\ref{prop:gpnp}}}
        Let $\rho$ be a strictly positive %
        density on $\mathbb{R}^{d}$ and let $D$ be a differentiable diagonal transport map. The log of the pullback density $\pi = D^{\sharp}\rho$ is given by
            \begin{align} 
                \log \pi(\bm{z}) = \log D^\sharp \rho(\bm{z}) &= \log \rho \circ D(\bm{z}) + \log|\det \nabla D(\bm{z})| \nonumber \\
                &= \log \rho \circ D(\bm{z}) + \sum_{k=1}^{d} \log \partial_{k} D^k(z^k). \label{eq:log_density_diag}
            \end{align}
    The partial derivatives of the log-density with respect to $z_{i},z_{j}$ are given by
        \begin{align} %
            \partial_i \partial_j \log \pi(\bm{z}) = \partial_{i} \partial_{j} \log \rho \circ D(\bm{z}) \, \partial_{i} D^{i}(z_{i}) \partial_{j} D^{j}(z_{j}), \nonumber %
        \end{align} 
    by using that each term in %
    the log-determinant of the map's Jacobian %
    only depends on a single variable. Thus, entry $(i,j)$ of the conditional independence score for $\pi$ is given by %
    \begin{equation} \label{eq:CMI_diag_pi}
        \Omega_{ij} = %
        \int \left|\partial_{i} \partial_{j} \log \rho \circ D(\bm{z}) \, \partial_{i} D^{i}(z_{i}) \partial_{j} D^{j}(z_{j}) \right|^2 \pi(\bm{z}) d\bm{z}.
    \end{equation}
    Using the measure transformation $\bm{\nu}_{\pi} = D^{\sharp}\bm{\nu}_{\rho}$, we rewrite %
    the conditional independence score of $\pi$ in~\eqref{eq:CMI_diag_pi} as the following expectation over $\rho$ %
    \begin{equation}
        \Omega_{ij} = \int \left|\partial_{i} \partial_{j} \log \rho(\bm{x}) \, \partial_{i} D^{i}((D^i)^{-1}(x_i)) \partial_{j} D^{j}((D^j)^{-1}(x_j)) \right|^2 \rho(\bm{x}) d\bm{x}.
    \end{equation}
    Finally, by applying the inverse function theorem to each continuously differentiable map component to get $\partial_{i} D^{i}((D^{i})^{-1}(x_{i})) = \partial_{i} (D^{i})^{-1}(x_{i})$, we arrive at the score in~\eqref{eq:CIS_npn}.
\end{proof}

\begin{proof}{\hspace{-.1em}\textbf{of Proposition~\ref{prop:mo}}}
    For the polynomial degree $\beta = 1$, the  transport map $S$ is an affine function $S(\bm{z}) = L(\bm{z} - \mathbf{c})$ with an invertible lower triangular matrix $L \in \mathbb{R}^{d \times d}$ and vector $\mathbf{c} \in \mathbb{R}^{d}$. For this class of transport maps, the minimization problem in~\eqref{eq:mle} becomes
    \begin{align}
        \argmax_{S} \frac{1}{n}\sum_{l=1}^{n} \log S^\sharp \eta(\textbf{z}^{l}) &= \argmax_{S} \frac{1}{n}\sum_{l=1}^{n} \log \eta \circ S(\textbf{z}^{l}) + \log |\det \nabla S(\textbf{z}^{l})| \nonumber \\
        &= \argmax_{L,\mathbf{c}} \frac{1}{n}\sum_{l=1}^{n} \left(-\frac{1}{2}(\textbf{z}^l - \mathbf{c})^{T}L^{T}L(\textbf{z}^l - \mathbf{c}) + \log \det L \right). \label{eq:linear_map_objective}
    \end{align}
    For any invertible matrix $L^{T}L$, the optimal $\mathbf{c}$ is the empirical mean $\widehat{\mathbf{m}} := \frac{1}{n}\sum_{l=1}^{n} \textbf{z}^{l}$. Substituting this value for $\mathbf{c}$ in~\eqref{eq:linear_map_objective}, the objective for $L$ is given by %
    \begin{align} \label{eq:objective_cholesky}
        \argmin_{L} \left\{ \text{Tr}\left(L^{T}L \widehat{\Sigma} \right) - \log \det \left(L^{T}L \right) \right\},
    \end{align}
    where $\widehat{\Sigma} := \frac{1}{n} \sum_{i=1}^{n}(\textbf{z}^{l} - \mathbf{c})(\textbf{z}^{l} - \mathbf{c})^{T}$ denotes the empirical covariance matrix and $\text{Tr}$ is the matrix trace operator.
            Setting the gradient of~\eqref{eq:objective_cholesky} with respect to $L^{T}L$ to zero yields the optimal $L$ to be the inverse of the Cholesky factor of $\widehat{\Sigma}$ (which exists for $n \geq d$). %
            Thus, the pullback density $S^{\sharp}\eta$ %
            for a standard Gaussian density $\eta$ yields a multivariate Gaussian approximation to $\pi$ with mean $\widehat{\mathbf{m}}$ and covariance matrix $\widehat{\Sigma}$.
\end{proof}

\section{Proof of Proposition~\ref{prop:sample_complexity}}\label{app:consistency} \begin{proof}%
The proof considers two types of errors: a false positive occurs when the true score $\Omega_{ij}$
    is zero but the threshold estimate $\overline{\Omega}_{ij} =
    \widehat{\Omega}_{ij}\mathbbm{1}(\widehat{\Omega}_{ij} > \tau_{ij})$ is nonzero; and a false
    negative occurs when the true score ${\Omega}_{ij}$ is nonzero but the threshold estimate
    $\overline{\Omega}_{ij}$ is zero.

For each pair of variables $(i,j)$, let $g\colon \bm{\alpha} \mapsto
    \mathbb{E}_{\pi}|\partial_i\partial_j \log S_{\bm{\alpha}}^{\sharp}\eta(\bm{z})|^2$ be a
    continuous function of the coefficients $\bm{\alpha}$. Then, we have $\widehat{\Omega}_{ij} =
    g(\widehat{\bm{\alpha}})$ (as defined in \eqref{eq:exp_ci}) and $\widehat{\upsilon}_{ij} =
    (\nabla_{\bm{\alpha}} g(\widehat{\bm{\alpha}})^T \Gamma(\widehat{\bm{\alpha}})^{-1}
    \nabla_{\bm{\alpha}} g(\widehat{\bm{\alpha}}))^{1/2}$ (defined below \eqref{eq:threshsimple}). 
Assuming that $g$ is twice differentiable, a Taylor expansion of the score estimator around
    $\bm{\alpha}^*$ yields
\begin{equation} \label{eq:taylor_series_Omega}
g(\widehat{\bm{\alpha}}) = g(\bm{\alpha}^*) + \nabla_{\bm{\alpha}}
    g(\bm{\alpha}^*)^{T}(\widehat{\bm{\alpha}} - \bm{\alpha}^*) + \frac{1}{2} (\widehat{\bm{\alpha}}
    - \bm{\alpha}^*)^T \nabla_{\bm{\alpha}}^2 g(\bm{\alpha}^*)(\widehat{\bm{\alpha}} -
    \bm{\alpha}^*) + o_{p}(\|\widehat{\bm{\alpha}} - \bm{\alpha}^* \|^2),
\end{equation}
where the remainder is a term that tends to zero in probability. 

For conditionally dependent variables, we have $g(\bm{\alpha}^*) = \Omega_{ij} \neq 0$ and, by
    assumption, $\nabla_{\bm{\alpha}} g(\bm{\alpha}^*) \neq 0$. Therefore, we can truncate the
    expansion in~\eqref{eq:taylor_series_Omega} to first-order terms and from the delta method we
    have $\sqrt{n}(\widehat{\Omega}_{ij} - \Omega_{ij}) = \sqrt{n}(g(\widehat{\bm{\alpha}}) -
    g(\bm{\alpha}^*)) \overset{d}{\longrightarrow} \mathcal{N}(0,\nabla_{\bm{\alpha}}
    g(\bm{\alpha}^*)^T\Gamma(\bm{\alpha}^*)^{-1}\nabla_{\bm{\alpha}} g(\bm{\alpha}^*))$ as $n
    \rightarrow \infty$.
From the continuous mapping theorem, $\widehat{\upsilon}_{ij}$ tends in probability to the constant
    $\upsilon_{ij} = (\nabla_{\bm{\alpha}} g(\bm{\alpha}^*)^T
    \Gamma(\bm{\alpha}^*)^{-1}\nabla_{\bm{\alpha}} g(\bm{\alpha}^*))^{1/2} \neq 0$ (nonzero for
    positive-definite Fisher information matrix $\Gamma(\bm{\alpha}^\ast)$ by the assumption
    $\nabla_{\bm{\alpha}} g(\bm{\alpha}^*) \neq 0$) as $\widehat{\bm{\alpha}} \rightarrow \bm{\alpha}^*$. %
Then from Slutsky's theorem (Corollary 2.3.2 in~\cite{lehmann2004elements}), the ratio
    $\sqrt{n}(\widehat{\Omega}_{ij}/\widehat{\upsilon}_{ij} - \Omega_{ij}/\upsilon_{ij})
    \overset{d}{\longrightarrow} R$ as $n \rightarrow \infty$ where $R \sim \mathcal{N}
    \left(0,\nabla_{\bm{\alpha}} g(\bm{\alpha}^*)^T \, \Gamma(\bm{\alpha}^*)^{-1} \,
    \nabla_{\bm{\alpha}} g(\bm{\alpha}^*) / \upsilon_{ij}^2 \right) = \mathcal{N}(0,1)$.
Thus, the probability of false negatives given by 
\begin{align}
\mathbb{P}\left(\overline{\Omega}_{ij} = 0; \Omega_{ij} \neq 0 \right) &= \mathbb{P} \left( \widehat{\Omega}_{ij} < \frac{f(n)\widehat{\upsilon}_{ij}}{\sqrt{n}}; \, \Omega_{ij} \neq 0 \right) \nonumber \\
&= \mathbb{P} \left(\sqrt{n} \left(\frac{\widehat{\Omega}_{ij}}{\widehat{\upsilon}_{ij}} - \frac{\Omega_{ij}}{\upsilon_{ij}} \right)< f(n) - \frac{\sqrt{n}\Omega_{ij}}{\upsilon_{ij}}; \, \Omega_{ij} \neq 0 \right), \nonumber
\end{align}
asymptotically converges to %
\begin{equation} \label{eq:tail_bound_fp}
\mathbb{P} \left(R < f(n) - \frac{\sqrt{n}\Omega_{ij}}{\upsilon_{ij}} \right) \leq e^{-\frac{1}{2} \left ( f(n) - \frac{\sqrt{n}\Omega_{ij}}{\upsilon_{ij}} \right )^2},
\end{equation}
where the last inequality provides an explicit bound and follows from the Gaussian tail bound $\mathbb{P}(R < r) \leq e^{-r^2/2}$
for $r \leq -1$. For $f(n)/\sqrt{n} \rightarrow 0$, the term $f(n) - \frac{\sqrt{n}\Omega_{ij}}{\upsilon_{ij}}$ 
tends to negative infinity as $n \rightarrow \infty$, %
and therefore, the probability of a false negative in~\eqref{eq:tail_bound_fp} converges to zero for any $f$ that grows more slowly than $\sqrt{n}$. Conversely, if $f(n) = c\sqrt{n}$ for some $c \geq \Omega_{ij}/\upsilon_{ij}$, the left-hand side in~\eqref{eq:tail_bound_fp} does not go to zero. %

For conditionally independent variables, we have $\partial_{i}\partial_{j}\log S^\sharp_{\bm{\alpha}^*} \eta(\bm{z}) = 0$ for all $\bm{z} \in \mathbb{R}^{d}$; thus $g(\bm{\alpha}^*) = \mathbb{E}|\partial_{i}\partial_{j}\log S^\sharp_{\bm{\alpha}^*} \eta(\bm{Z})|^2 = 0$ and $\nabla_{\bm{\alpha}} g(\bm{\alpha}^*) = 2\mathbb{E}[\partial_i\partial_j \log S^{\sharp}_{\bm{\alpha}^*}\eta(\bm{Z})\nabla_{\bm{\alpha}}\partial_i\partial_j \log S^{\sharp}_{\bm{\alpha}^*}\eta(\bm{Z})] = 0$. %
Under the assumption that $\nabla_{\bm{\alpha}}^2 g(\bm{\alpha}^*) \neq 0$, we can truncate the expansion in~\eqref{eq:taylor_series_Omega} to second-order terms and approximate $\widehat{\Omega}_{ij}$ asymptotically using a quadratic form (i.e., a degree two polynomial). %
By Proposition 2.1 in~\cite{drton2016wald}, 
the Wald statistic $n\widehat{\Omega}_{ij}^2/\widehat{\upsilon}_{ij}^2 = \frac{ng(\widehat{\bm{\alpha}})^2}{\nabla_{\bm{\alpha}} g(\widehat{\bm{\alpha}})^T \Gamma(\widehat{\bm{\alpha}})^{-1} \nabla_{\bm{\alpha}} g(\widehat{\bm{\alpha}})} \overset{d}{\longrightarrow} W$, where the random variable $W$ satisfies, by Proposition 3.4 in~\cite{drton2016wald}, $\mathbb{P}(W \geq x) \leq \mathbb{P}(\frac{1}{4}X \geq x)$ for $X \sim \chi_p^2$ and all $x > 0$. %
Here, $\chi_{p}^2$ denotes a chi-squared variable with $p \geq 1$ degrees of freedom where $p$ corresponds to the dimension of $\bm{\alpha}$.
Then, the probability of false positives, which can be written as %
$$
\mathbb{P}(\overline{\Omega}_{ij} \neq 0; \Omega_{ij} = 0) %
= \mathbb{P} \left(\widehat{\Omega}_{ij} > \frac{f(n)\widehat{\upsilon}_{ij}}{\sqrt{n}} ; \Omega_{ij} = 0 \right) = \mathbb{P} \left(\frac{n\widehat{\Omega}_{ij}^2}{\widehat{\upsilon}_{ij}^2} > f(n)^2 ; \Omega_{ij} = 0 \right)$$ 
since $\widehat{\Omega}_{ij}$ and $\widehat{\upsilon}_{ij}$ are non-negative, 
is asymptotically given by
\begin{equation} \label{eq:tail_bound_fn}
 \mathbb{P} \left( W > f(n)^2 \right) \leq \mathbb{P} \left( X - p > 4p\left(\frac{f(n)^2}{p} - \frac{1}{4} \right) \right) \leq e^{-f(n)^2/p + 1/4},
\end{equation}
where the last inequality uses the $\chi_p^2$ squared tail bound $\mathbb{P}(X - p \geq 4px) \leq \mathbb{P}(X - p \geq 2\sqrt{px} + 2x) \leq e^{-x}$ for $x \geq 1$ (Lemma 1 in~\cite{laurent2000adaptive}). For $f(n) \rightarrow \infty$, the right-hand side in~\eqref{eq:tail_bound_fn} and hence the left-hand side converge to zero as $n \rightarrow \infty$.

To complete the proof we use a union bound over all pairs $(i,j)$ to bound the probability of failing to recover the true edge set
\begin{equation}
\mathbb{P}(\widehat{E}_n \neq E) \leq \sum_{(i,j) \in E} \mathbb{P}(\overline{\Omega}_{ij} = 0; \Omega_{ij} \neq 0) + \sum_{(i,j) \not\in E} \mathbb{P}(\overline{\Omega}_{ij} \neq 0 ; \Omega_{ij} = 0). \label{eq:consistency_bound} %
\end{equation}
Using the asymptotic results in~\eqref{eq:tail_bound_fp} and~\eqref{eq:tail_bound_fn} to bound each term in~\eqref{eq:consistency_bound} we have $\mathbb{P}(\widehat{E}_n \neq E) \rightarrow 0$ as $n\rightarrow \infty$.
\end{proof}

\section{Data on memory usage}\label{app:memory}
\rev{Following the discussion in Section~\ref{sec:costsmain}, we report the memory usage of the first two iterations of \textsc{sing} for different data set dimensions $d$ and polynomial degrees $\beta$, applied to the butterfly dataset of Section~\ref{ssec:but}.}

\begin{table}[h!]
\centering
\begin{tabular}{ |c|c|c|c|c| } 
 \hline
Polynomial degree $\beta$ & Dimension $d$ & Map estimation & Estimating $\widetilde{\Omega}$ & Estimating $\mathbb{V}\text{ar}[\widetilde{\Omega}]$\\
 \hline
1  & 4  & $51.8 \times 10^3$  & $19.3 \times 10^6$  & $10.2 \times 10^7$\\
2  & 4  & $96.2 \times 10^5$  & $55.3 \times 10^6$ & $29.4 \times 10^7$\\
1  & 6  & $18.3 \times 10^6$  & $44.9 \times 10^6$ & $33.3 \times 10^7$\\
2  & 6  & $21.9 \times 10^6$  & $18.0 \times 10^7$ & $12.2 \times 10^8$\\
1  & 8  & $57.6 \times 10^6$  & $13.2 \times 10^7$ & $84.5 \times 10^7$\\
2  & 8  & $73.1 \times 10^6$  & $49.6 \times 10^7$ & $35.9 \times 10^8$\\
1  & 10 & $36.0 \times 10^6$  & $13.4 \times 10^7$ & $15.9 \times 10^8$\\
2  & 10  & $21.6 \times 10^6$  & $99.6 \times 10^7$ & $94.4 \times 10^8$\\
1  & 12  & $49.4 \times 10^6$  & $13.3 \times 10^7$ & $27.8 \times 10^8$ \\
2  & 12 & $35.7 \times 10^6$  & $19.0 \times 10^8$  & $22.0 \times 10^9$ \\
 \hline
\end{tabular}
\caption{Breakdown of memory usage (in bytes) for the first iteration of \text{sing} applied to the butterfly data set with $n = 1000$ samples.}
\end{table}

\begin{table}[h!]
\centering
\begin{tabular}{ |c|c|c|c|c| } 
 \hline
Polynomial degree $\beta$ & Dimension $d$ & Map estimation & Estimating $\widetilde{\Omega}$ & Estimating $\mathbb{V}\text{ar}[\widetilde{\Omega}]$\\
 \hline
1  & 4  & $36.8 \times 10^5$  & $13.4 \times 10^6$  & $33.0 \times 10^6$ \\
2  & 4  & $21.5 \times 10^6$  & $34.4 \times 10^6$  & $63.7 \times 10^6$ \\
1  & 6  & $13.1 \times 10^6$  & $28.2 \times 10^6$  & $97.4 \times 10^6$ \\
2  & 6  & $51.9 \times 10^6$  & $34.3 \times 10^6$  & $13.8 \times 10^7$ \\
1  & 8  & $78.1 \times 10^6$  & $84.1 \times 10^6$  & $24.8 \times 10^7$ \\
2  & 8  & $12.9 \times 10^7$  & $96.0 \times 10^6$  & $34.2 \times 10^7$ \\
1  & 10 & $52.8 \times 10^6$  & $38.5 \times 10^6$  & $35.9 \times 10^7$ \\
2  & 10  & $13.9 \times 10^7$  & $48.1 \times 10^6$  & $52.8 \times 10^7$ \\
1  & 12  & $18.1 \times 10^6$  & $59.7 \times 10^6$  & $61.3 \times 10^7$ \\
2  & 12  & $16.2 \times 10^7$  & $66.4 \times 10^6$  & $89.6 \times 10^7$ \\
 \hline
\end{tabular}\caption{Breadown of memory usage (in bytes) for the second iteration of \text{sing} applied to the butterfly data set with $n = 1000$ samples}
\end{table}

\vskip 0.2in
\bibliography{references.bib}

\end{document}